%% file: nature-from-scratch-clean.tex
\documentclass{ifm-tr}

\input{math_commands.tex}

\usepackage{amssymb}
\usepackage{mathtools}
\usepackage{enumitem}
\usepackage{siunitx}
\usepackage{todonotes}
\usepackage{xcolor}
\usepackage[dvipsnames]{xcolor}
\usepackage[most]{tcolorbox}
\usepackage{pifont}
\usepackage{graphicx}
\usepackage{booktabs}
\usepackage{amsmath}

\usepackage{tikz}
\usetikzlibrary{arrows.meta,positioning,fit,backgrounds}

\usepackage{xcolor}
\providecommand{\cmark}{\ding{51}}
\providecommand{\xmark}{\ding{55}}
\providecommand{\pmark}{$\circ$}
\title{{
PAN 2: Joint Conceptual and Perceptual Modeling Enables World Simulation and Simulative Reasoning
}
}

\author[$*$]{Jiuhai Chen}\authorsep{, }
\author[$*$]{Mingkai Deng}\authorsep{, }
\author[$*$]{Zeyu Feng}\authorsep{, }
\author[$*$]{Yi Gu}\authorsep{, }
\author[$*$]{Matt Gwilliam}\authorsep{, }
\author[$*$]{Guangyi Liu}\authorsep{, }
\author[$*$]{Hongyi Xu}\authorsep{,\\}
\author[$*$]{Zhiyang Xu}\authorsep{,}
\author[]{Ben Caffee}\authorsep{, }
\author[]{Xin Gao}\authorsep{, }
\author[]{Yingqing Guo}\authorsep{, }
\author[]{Ligong Han}\authorsep{, }
\author[]{Guowei He}\authorsep{, }
\author[]{Yiyan Hu}\authorsep{, }
\author[]{Mingjia Huo}\authorsep{, }
\author[]{Jeff Lai}\authorsep{,}
\author[]{Xuezhi Liang}\authorsep{, }
\author[]{Tingting Liao}\authorsep{, }
\author[]{William Lin}\authorsep{, }
\author[]{Zihan Liu}\authorsep{, }
\author[]{Kartik Patwari}\authorsep{, }
\author[]{Phu Pham}\authorsep{,\\}
\author[]{Mohammad Areeb Qazi}\authorsep{,}
\author[]{Junda Su}\authorsep{, }
\author[]{Prateksha Udhayanan}\authorsep{, }
\author[]{Dequan Yang}\authorsep{, }
\author[]{Tianle Yu}\authorsep{, }
\author[]{Yuheng Zha}\authorsep{,\\}
\author[]{Zhengzhong Liu}\authorsep{, }
\author[$\dag$]{Eric P.\ Xing}\authorsep{}
\affiliation[]{Institute of Foundation Models, Mohamed bin Zayed University of Artificial Intelligence}
\affiliationsep{\\{$^*$\,Co-first authors (alphabetical); $^\dag$\,Corresponding author}}

\reportnumber{2026-11-27}

\abstract{
Physical intelligence, the understanding of the physical world necessary for acting within it,
requires a model grounded in an ``Aristotelian'' world of observable consequences, rather than only a ``Platonic'' world of abstract representations.
Such a world model
must simulate how actions change the physical world by reasoning about both conceptual (e.g., what happens) and perceptual information (e.g., what it looks like) to produce predictions that can be inspected and checked against observations, so that an agent can reason about the outcomes of its choices before acting.
We build PAN~2, an open-source world model for these requirements, by combining an existing pretrained large language model (LLM), which predicts the next conceptual state as a structured, time-grounded description, with a latent diffusion model (LDM) we pretrain from scratch on large-scale visual data to predict the corresponding perceptual state, within a generative latent prediction (GLP) architecture.
We examine six dimensions of physical intelligence (6D-PI)
that faithful simulation requires, from physical consistency to
viewpoint invariance,
and the simulation of multi-step plans in which every step is checked.
PAN~2 scores 84.5\% across the six dimensions, exceeding or competitive with leading open-weight and proprietary baselines, and achieves the highest full-plan completion (34.0\%, with 75.1\% step completion).
In particular, full-plan completion follows a compounding curve that predicts rapid growth as per-step reliability continues to improve.
Both LLM-based conceptual and LDM-based perceptual reasoning prove essential, as without either, full-plan completion drops to 0\% with corresponding collapse in per-step completion.
The model also simulates conditions absent from its training data, enabling counterfactual reasoning, and supports agents that learn from simulated interaction.
These findings suggest a promising path towards general-purpose world simulators that learn faithful and actionable physical reasoning from observed data and interaction.
}

\begin{document}
\maketitle

Much of lingual intelligence, the capabilities that make up understanding of and competence with language, is captured by the ability to predict the next word~\citep{radford2019language,brown2020language}.
Physical intelligence, the analogous capabilities for acting in the physical world, rests in the same way on the ability to simulate world possibilities: namely, how the world will change in response to actions.
We call a model with this ability a {\it World Model}~\citep{ha2018world,lecun2022path,xing2025critique}.
Simulation in turn enables simulative reasoning, in which an agent imagines the outcomes of its candidate actions before choosing among them, whether during training or at decision time~\citep{xing2025critique}.
As the simulation becomes more faithful, decisions reached this way approach optimal decisions in the real world~\citep{xing2026cam}, and an agent can learn and plan without costly, slow, or irreversible physical trials~\citep{hafner2019dream}.

What does it mean to simulate the physical world well?
Concretely, the predicted next world should obey the laws of physics, actions should take effect in the order given, entities should persist out of sight, physical constraints should hold, the camera should move without changing the world it shows, and the world should keep moving on its own even without action.
We condense these properties into six dimensions of physical intelligence (6D-PI), each corresponding to a capability of world simulation:
\emph{physical consistency}, \emph{causal coherence}, \emph{entity persistence}, \emph{constraint satisfaction}, \emph{viewpoint invariance}, and \emph{objective dynamism}.

For simulative reasoning, a world model must simulate not just a single action but a sequence of actions, each starting from the state the previous one produced.
Simulating merely a successful outcome is not enough, as a model can reach the right end state through an impossible sequence of events~\citep{ha2018world}, so each step must be checked individually.
We measure this with \emph{step completion}, the fraction of steps carried out correctly, and \emph{full-plan completion}, the fraction of plans in which every step succeeds.

These capabilities place requirements on the model itself.
Because a world model is meant to serve many agents and tasks, it must be general, covering as much of the physical world as possible, and so must be a foundation model pretrained on broad data~\citep{bommasani2021opportunities}.
It must take in the multimodal signals through which the world is observed, combining discrete representations for symbolic information (e.g., objects, language) with continuous ones for perceptual detail (e.g., texture, contact), and it must be interactable, conditioned on the actions of the agents it serves.
It must reason over both the conceptual and the perceptual aspects of the world,
suggesting its backbone can build on large language models (LLMs), which reason over concepts through text-based chains of thought~\citep{wei2022chain,brown2020language}, and on latent diffusion models (LDMs), which predict perceptual detail through iterative denoising~\citep{rombach2022high,ho2020denoising}.
Its predictions must be auditable through content that can be seen rather than an abstract representation, and grounded, trained and verified against the next observation itself.
Together, these requirements converge on next-world prediction, formalized by the generative latent prediction (GLP) architecture of an encoder, a backbone and a decoder~\citep{xing2025critique}.

Existing world models share parts of this design and differ in others.
Many are generative, predicting future observations as video, and are often obtained by fine-tuning pretrained video generators for applications such as games, spatial navigation, or robots~\citep{genie3,worldlabs2026atlas,bi2026motus2};
Pandora and PAN, our earlier models, took this route to realize general-purpose simulation~\citep{xiang2024pandora,panteam2026panworldmodelgeneral}.
Others predict the future in a learned latent space without reconstructing observations~\citep{lecun2022path,mur2026v}, or learn world prediction jointly with an action policy~\citep{lu2026world,ye2026world}.
LLMs have also been used to simulate the world, either in text~\citep{hao2023reasoning,zuo2026qwen} or by writing programs that drive renderers and physics engines~\citep{chen2026code,tang2026video2world}.

In this work,
we study how such a world model can be built by coupling conceptual reasoning by an LLM with perceptual reasoning by an LDM that we pretrain from scratch, in a new open-source model we call PAN~2.
The two backbones are connected by a text-based state that we design so that the LLM's predictions at inference match the annotations the LDM learned from.

The LLM predicts the next state of the world as a structured, time-grounded description of events, actions and camera motion, and the LDM, trained on videos annotated in the same format, predicts the perceptual detail that the description leaves open.
Because most of a world model's knowledge arguably derives from its base model, we build PAN~2 from its data onward,
so that the contribution of each component, including the conceptual and the perceptual backbone, can be studied directly.
We examine what simulative reasoning requires of such a model: the
six dimensions (6D-PI), the simulation of each action in multi-step plans, and the contribution of the conceptual backbone
toward problem solving (i.e., reaching the goal).
Model capabilities in 6D-PI are shown to be acquired at different stages of training; across them,
PAN~2 passes 84.5\% of the 6D-PI checks pooled over all test cases, within the range of the other open-weight or proprietary models we study (61.3--87.2\%).

Simulating a plan in full compounds on simulating each step. Leading open-weight and proprietary models fall between 4.0 and 34.0\% for full-plan completion, versus 38.5--75.1\% for step completion. PAN~2 achieves the highest full-plan completion at 34.0\%, with 75.1\% step completion.
In particular, full-plan completion rises steeply along a compounding curve as each step becomes more reliable ($R^2 = 0.986$ across models), showing a path towards support for simulative reasoning.

Both the conceptual and perceptual backbones are essential: without the LLM, 6D-PI scores drop by about half and full-plan completion to zero; without adequate computation for the LDM, 6D-PI scores fall to about a quarter and full-plan completion likewise reaches zero.

The model also simulates combinations of conditions absent from its training data, and supports agents that discover hidden properties as they explore an expanding world through simulated interaction, although keeping a simulated world consistent over long interactions remains a challenge.

The remainder of the paper is organized as follows.
We describe PAN~2's architecture and training
(Section~\ref{sec:foundation-model-physical-intelligence}), evaluate it on the six
dimensions and multi-step plans (Sections~\ref{sec:simulation}--\ref{sec:simulative-reasoning}), examine its generalization beyond
the training data (Section~\ref{sec:generalization}) and its support for agent learning (Section~\ref{sec:agent-learning}), and
ablate its conceptual and perceptual backbones (Section~\ref{sec:ablations}). We close with a discussion of what the results reveal about building world models for physical intelligence.
To support research and development in world models, we open-source the model weights at \url{https://huggingface.co/IFM/PAN2} and code for training and inference at \url{https://github.com/ifm-ai/PAN2}. To our knowledge, existing world models are either open-weight only or fully proprietary, adding difficulties for thorough analysis and comparison attempted in this paper.

\section{A Foundation Model for Physical Intelligence}
\label{sec:foundation-model-physical-intelligence}

\begin{figure*}[!t]
	\centering
	\includegraphics[width=\textwidth,page=3]{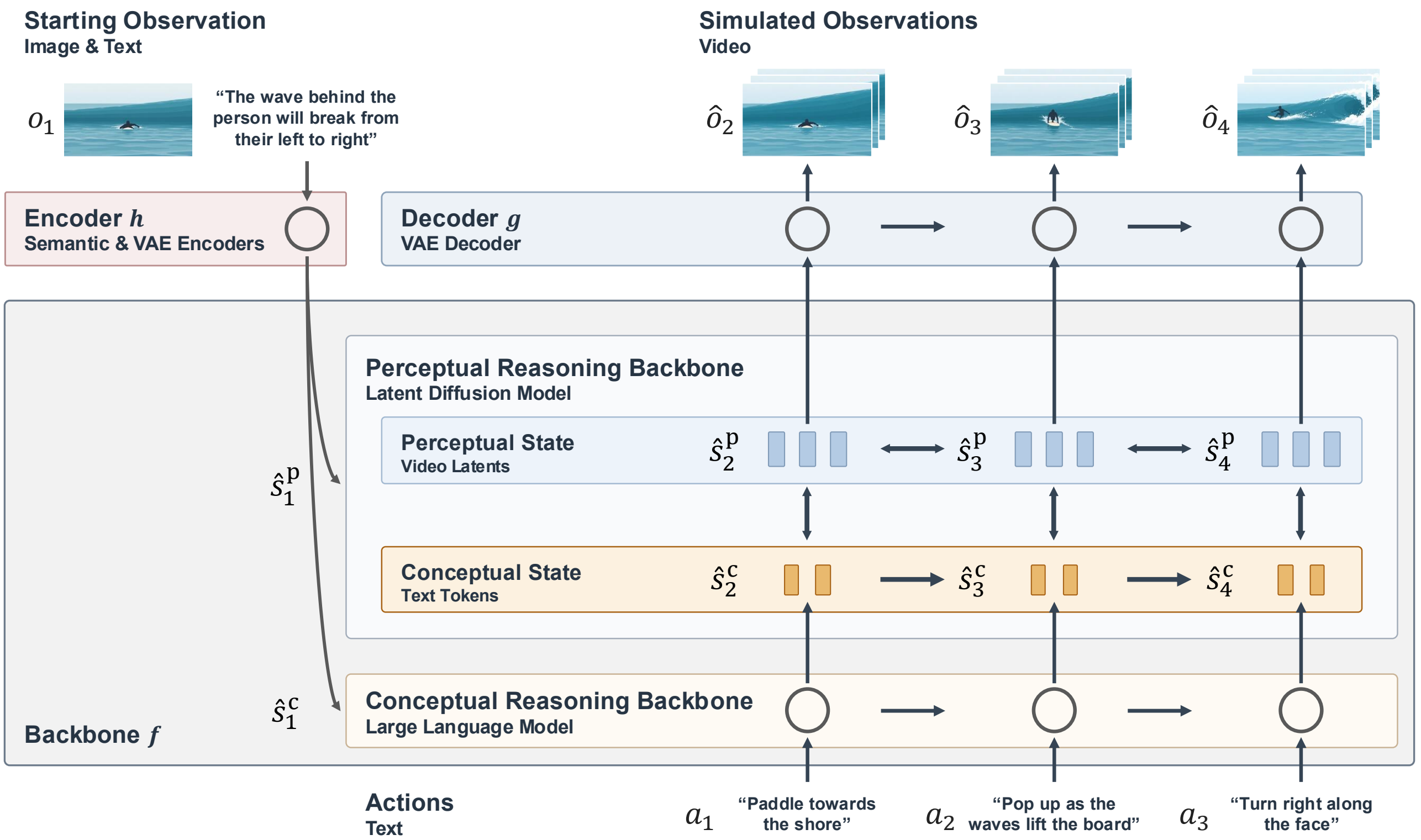}
    \caption{
    \textbf{A world model that couples conceptual and perceptual reasoning.}
    The encoder represents the current observation with a semantic latent, from the vision encoder of a vision-language model, and a generation latent, from a VAE encoder. The LLM (conceptual backbone) predicts a structured, time-grounded text description of the next state. The LDM (perceptual backbone) predicts the corresponding video latents, conditioned on the text-based state and the generation latent. The VAE decoder maps the predicted latents to the next observation.
    }
	\label{fig:overview}
\end{figure*}

PAN~2 employs the Generative Latent Prediction (GLP) architecture to build an action-conditioned world model from imagery (video and still images) and text data,
trained by learning to reconstruct the next observation.
It supports both symbolic (conceptual) and continuous (perceptual) reasoning, and simulates the next world through visual rollouts (Fig.~\ref{fig:overview}).
Formally,
at step $t$, it models the next observation $o_{t+1}$, a video, given the current observation $o_t$, an optional image with text, and an action $a_t$, an instruction in text with the duration to simulate, as below:
\begin{equation}
p(o_{t+1}\mid o_t,a_t)=\sum_{\hat s_t,\hat s_{t+1}}\underbrace{p_h(\hat s_t\mid o_t)}_{\text{encoder}}\;\underbrace{p_f(\hat s_{t+1}\mid \hat s_t,a_t)}_{\text{backbone}}\;\underbrace{p_g(o_{t+1}\mid \hat s_{t+1})}_{\text{decoder}},
\label{eq:pan2}
\end{equation}
where each estimated state $\hat s$ has a conceptual and a perceptual part.
The encoder $h$ represents the conceptual part through the vision encoder and text embedding of a multimodal LLM~\citep[Kimi-K3;][]{kimiteam2026kimik3openfrontier}, and the perceptual part through the encoder of a variational autoencoder~\citep[Wan2.2 VAE;][]{wan2025}.
The backbone $f$ predicts the conceptual part of the next state using a pretrained LLM (Kimi-K3) and the perceptual part through a 29B LDM that we design and train from scratch.
The Wan2.2 VAE decoder $g$ maps the predicted latents to video, which makes each prediction auditable and lets it be trained and checked against the next observation.

The LLM conceptual reasoning backbone, given an action, reasons about how the world will evolve and writes the next conceptual state as a structured, time-grounded description (Methods; Eq.~\ref{eq:conceptual-state}).
In PAN~2, we design this state as structured text which explicitly models the subjects, the timing of events and actions, and the camera motion separated from the motion of subjects, and leaves geometry, appearance, and the details of motion to the perceptual state.
Fig.~\ref{fig:overview} illustrates the division with a surfing example: given a wave building behind a surfer, the LLM writes when the surfer pops up and which way they turn, while the LDM predicts how the wave and the rider move concretely.
For the LDM to condition on such states, we annotate every training video in this format with a three-stage annotation pipeline that labels when things happen, estimates camera motion geometrically, and describe subject actions conditioned on both (Methods).

Training the LDM backbone for perceptual reasoning requires a progressive curriculum,
because high-resolution, long, motion-rich videos are substantially harder training examples than text corpora.
We therefore train the LDM in four stages
with increasing perceptual and dynamical complexity (Methods; Extended Data Table~\ref{tab:training-stages}).
Stage~I trains on about 700 million 480p broad-coverage video clips of up to 20 seconds, interleaved with images.
Stage~II continues at 720p on about 93 million clips, under stricter quality filters.
Within Stages~I and II, training clips are further ordered by increasing duration, quality and topic balance.
Stage~III trains on about 3 million clips under the strictest filters, balanced across topics and durations and favoring high motion and longer duration.
Stage~IV refines the model by reward learning with VisionReward~\citep{xu2024visionrewardfinegrainedmultidimensionalhuman}, a video reward model trained on human judgment, using 13 of its questions on object structure and persistence, motion, composition and perceptual detail.
Because a learned reward covers only part of what faithful simulation requires, optimizing it further can trade the dimensions it covers for those it does not.
We therefore release a weighted average of the Stage~III and IV checkpoints, a common remedy for such over-optimization~\citep{wortsman2022model,rame2023rewarded}.

\section{Simulating World Possibilities}
\label{sec:simulation}

\begin{figure*}[!t]
\centering
\includegraphics[width=0.95\textwidth]{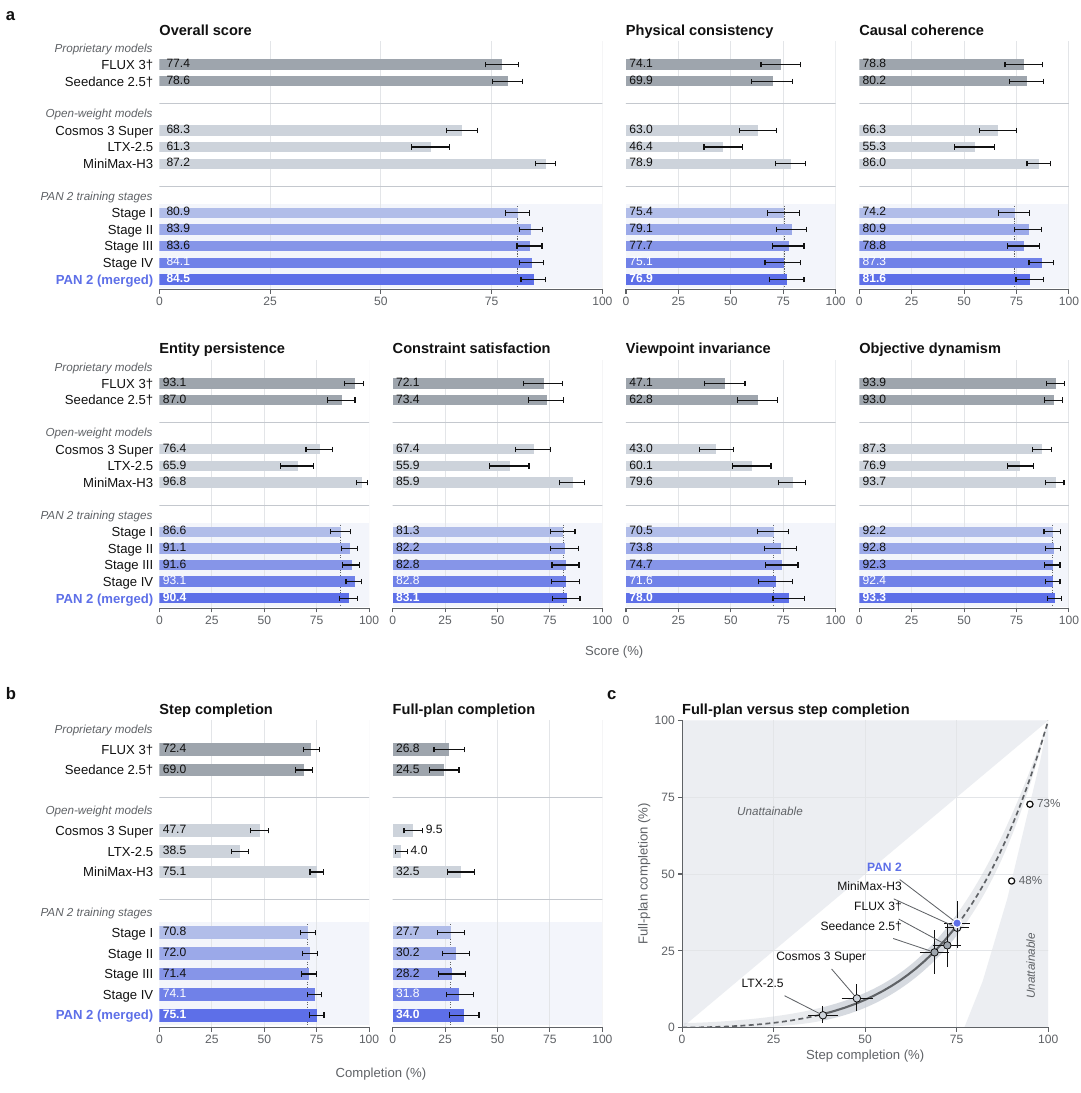}
\caption{\small \textbf{Six dimensions of physical intelligence and multi-step simulation.}
\textbf{a,} Overall score and scores on the six dimensions for five other models, the four PAN~2 training
checkpoints (Stages~I--IV) and the released PAN~2, which merges the checkpoints before and after
reward-based learning (Stages~III and IV). Bars show mean scores and error bars show 95\% percentile bootstrap confidence
intervals over test cases: 10,000 draws over 204 base cases, retaining paired modalities and all seeds within
each case. Dotted lines mark the Stage~I score.
$^\dagger$, proprietary models accessed through an API, with one generation per test case; their intervals
are conditional on that generation.
\textbf{b,} Step completion and full-plan completion on multi-step plans for the same models and checkpoints.
Intervals resample the 100 plans, retaining both modalities and all seeds. All intervals describe variation
over test cases, not over new seeds, and all axes start at zero.
\textbf{c,} Full-plan completion against step completion for the six models (95\% confidence intervals as
in \textbf{b}). The line is the compounding curve fitted across the six models: solid over their observed
range of step completion, dashed where it is extrapolated. The band is its pointwise 95\% plan-bootstrap
band, lighter where extrapolated. Shaded regions are values that full-plan completion cannot take, however
failures are distributed. Open circles are the lower bounds at 90\% and 95\% step completion (Methods).}
\label{fig:main-results}
\end{figure*}

Aesthetic and realism quality of video generation, as aimed by and measured on by many existing world models, are reflective of some capabilities of world simulation.
We push further to examine world simulation and reasoning capacities closer to what biological agent would employ their mental world model~\citep{ullman2017mind}: obeying the laws of physics, coherent over history, persistence beyond sight, conformity to constraints, invariance to viewpoints, and adherence to objective dynamism.

Each dimension is examined
through test cases that give a starting observation (i.e., an optional image and a text description) together with an action in text, and ask whether the simulated video demonstrates
the capability along that dimension in question.
We built these test cases from existing benchmarks and our own: reviewing the prompts of 11 public video-generation and world-model benchmarks, we reused the kinds of test case they cover and wrote new ones for what they rarely test, such as actions whose order matters and movements of the camera (Methods).
From a pool of 387, we selected 300 that balance the
six dimensions
and avoid near-duplicates.
Most test cases involve
several dimensions at once
and are scored for each of them; 50 of the 300 test cases
center on each dimension.
PAN~2 was not trained on any of the test cases; the training data of the other models are not disclosed, so we cannot tell whether they include these test cases or the public benchmarks that informed them.
Each test case lists yes/no questions with expected answers, and a
dimension's score
is the share of its questions answered as expected.
Viewpoint invariance
is judged from camera trajectories estimated from the generated video;
the other dimensions are scored
by a vision-language model (VLM) judge
whose verdicts correlate with those of human annotators (Methods).
The world model is meant to support simulative reasoning by agents and would ideally be judged by how much it improves the agents' decisions; these automated scores, validated against human annotators rather than the AI agents, are a scalable proxy for that standard, but leave possibilities for human bias.

As a frame of reference, we run the same test cases on five other models: the open-weight Cosmos-3 Super, MiniMax-H3, and LTX-2.5, and the proprietary FLUX-3 and Seedance-2.5.
Each model is run as a holistic system with its recommended prompt rewriter. The exception is MiniMax-H3,
whose own rewriter is a hosted service that does not accept test cases of this length; following its official guide for using one's own LLM, we run H3's prompt rewriting instructions with Kimi-K3, the same LLM that PAN~2 uses.
Latent-only~\citep[e.g.,][]{mur2026v} and code-based~\citep[e.g.,][]{chen2026code} world models that we are aware of do not produce observations from an image-and-text input, which our test cases require; placing them on the same evaluation would require building observation decoders for their latent predictions, which we discuss how builders or users of these models could pursue themselves (Discussion).
Across all decided capability questions, scores range from 61.3\% to 87.2\% across models, with PAN~2 at 84.5\% (Fig.~\ref{fig:main-results}a).
The more informative picture, however, emerges
dimension by dimension
(Fig.~\ref{fig:main-results}a).

\subsection*{Physical Consistency: Obeying the laws of physics}

The simulated world should adhere to the same laws as the physical world.
An egg dropped from a height should fall, accelerate, and crack; honey should flow more slowly than water.
Without this, no simulated outcome can be relied upon to guide decisions in the real world.

Physical consistency is the hardest of the
six dimensions
for
top-performing models,
as scores range from 46.4\% to 78.9\% across models, and it is the
lowest-scoring dimension
for both MiniMax-H3 and PAN~2 (76.9\%).
In PAN~2, physical consistency rises
from 75.4\% after Stage~I to 79.1\% after Stage~II, consistent with the finer detail and longer dynamics that higher resolution and longer clips make visible.
It does not improve further in Stage~III, whose stricter selection favors visual quality and motion, or in Stage~IV, whose reward is less reliable on physical correctness~\citep{xu2024visionrewardfinegrainedmultidimensionalhuman};
the merged model (76.9\%) lies between its Stage~III and IV checkpoints, below the Stage~II peak.

\subsection*{Causal Coherence: Following actions in order}

The simulated world should evolve according to each supplied action, in the order given.
Unlocking a door and then opening it should yield a different outcome from trying to open the door before unlocking it.
Without this, plans that differ only in sequence look the same, and an agent cannot trust that each of its actions has been simulated.
In practice, although the actions are given in order, models often merge them, skip one or render only the final state, so our test cases tie the order to a physical precondition (e.g., unfolding a stool before reaching a jar on a high shelf), under which skipping or reordering a step changes the outcome.

Causal coherence ranges from 55.3\% to 86.0\% across models (81.6\% for PAN~2), and of the
six dimensions
it changes most over training.
In PAN~2, it rises from 74.2\% after Stage~I, which trains on images and lower-resolution video, to 80.9\% after Stage~II, which trains on higher-resolution video, drops slightly after Stage~III to 78.8\%, and reaches 87.3\% with reward learning in Stage~IV.
Merging the parameters of the Stage~III and IV checkpoints (70\%/30\% by weight)
gives back part of that increase (81.6\%).

\subsection*{Entity Persistence: Remembering what is out of sight}

Entities, whether subjects, objects or agents, should keep their identity and state through occlusion or when out of view.
A jar placed in a cupboard should still be there, unchanged, when the cupboard is opened again.
Without this, whatever the agent stops looking at is lost, and long-horizon interactions cannot be simulated.
Over longer interactions, persistence also depends on how the simulated world keeps its state across clips, which we examine in the context of agent interactions later (see `Towards agent learning in world simulation').

Entity persistence is among the two
highest-scoring dimensions
for every model (65.9--96.8\% across models; 90.4\% for PAN~2).
In PAN~2, persistence improves steadily through training, from 86.6\% after Stage~I to 93.1\% after Stage~IV, the largest gain after causal coherence.
The largest step comes in Stage~II (to 91.1\%), consistent with its longer clips, which give objects more time to leave view and return; Stage~IV, whose reward includes VisionReward's questions on
object permanence,
drives further increase to 93.1\%.
Merging the Stage~III and IV checkpoints results in a slight decrease (90.4\%).

\subsection*{Constraint Satisfaction: Respecting what is possible}

Obstacles, support, reach, clearance and prerequisites should limit what can happen.
A lamp should not turn on unless it is plugged in.
Without this, infeasible plans appear feasible.

Constraint satisfaction ranges from 55.9\% to 85.9\% across models (83.1\% for PAN~2).
Unlike causal coherence, it is largely in place in PAN~2 after Stage~I (81.3\%) and changes little thereafter (82.2--83.1\%).
Neither higher resolution, motion-rich selection, nor reward learning moves it appreciably, suggesting that it is learned early, from the broad coverage of Stage~I.

\subsection*{Viewpoint Invariance: Moving the eye without changing the world}

The model should execute the specified camera movement while keeping the scene spatially coherent.
Panning left should reveal what lies to the left, and tilting up should show the ceiling, without anything in the scene moving because the camera did; the viewpoint should also face or show an object when requested.
Without this, an embodied agent cannot even simulate looking around, which is itself an action.

Viewpoint invariance
separates models more
than any other dimension:
scores range from 43.0\% to 79.6\% (78.0\% for PAN~2).
In PAN~2,
viewpoint invariance
rises from 70.5\% to 74.7\% over Stages~I--III, as resolution and clip length increase.
It falls to 71.6\% after reward-based learning, which emphasizes the stability and dynamism of camera motion rather than controllability.
Merging the Stage~III and IV checkpoints gives 78.0\%, above every individual checkpoint, which can be understood as an effect of parameter averaging that can exceed either parent~\citep{wortsman2022model}.

\subsection*{Objective Dynamism: A world that moves on its own}

Agents and moving objects in the simulated world, such as vehicles, crowds, animals and machines, should behave coherently and act independently, whether or not they respond to the input actions.
A person sitting idle in a cafe should still see others come and go, chat and drink coffee.
Without this, the world freezes around the agent whenever it stops acting.

Objective dynamism
separates models the least.
Four of the six models, PAN~2 among them, score between 93.0\% and 93.9\%, and the other two score 76.9\% and 87.3\%.
In our study, the highest-scoring models are therefore close to the ceiling, and harder probes would be needed to separate them.
In PAN~2, dynamism is already at 92.2\% after Stage~I
and stays close to that thereafter; the released model scores 93.3\%.

\subsection*{What the six dimensions reveal}

Three patterns emerge across the
six dimensions.
First, these dimensions are distinct:
a model's standing on one does not reliably predict its standing on another.
For example, FLUX 3, which scores the highest on
objective dynamism,
is among the lowest on
viewpoint invariance.
Second, the capabilities are acquired at different points in training, in ways consistent with the changes at each stage.
Objective dynamism
and constraint satisfaction are stable after the broad coverage of Stage~I.
Physical consistency and causal coherence, on the other hand, increase from the higher resolution and longer clips of Stage~II.
Entity persistence improves throughout, including in Stage~IV, whose reward function evaluates for object preservation.
Physical consistency and
viewpoint invariance,
on which the reward offers little or no signal, do not improve there.
Causal coherence, however, improves from reward learning, even though the reward
does not score it explicitly.
No single checkpoint is
best on every dimension.
Merging model parameters from
the Stage~III and IV checkpoints recovers most of what reward learning lost in viewpoint invariance and physical consistency, and keeps part of its gain in causal coherence but not in entity persistence, giving a slightly higher overall score than any single checkpoint.
Among three candidate merges, reviewers rated the released one highest overall (Supplementary Information~\ref{sec:model-soup-human-preference}).
Third, physical consistency and
viewpoint invariance
are the two
lowest-scoring dimensions
for both MiniMax-H3 and PAN~2, suggesting that they remain the main challenges for models of the physical world.
Each test case for these capabilities involves a single action; whether they hold together when actions are chained into a plan is the subject of the next section.

\section{Towards Simulative Reasoning}
\label{sec:simulative-reasoning}

The eventual goal of simulating world possibilities is to enable simulative reasoning.
Concretely, an agent proposes a sequence of actions $(a_t, a_{t+1}, \dots)$, uses the world model to simulate the states that would result $(\hat{s}_{t+1}, \hat{s}_{t+2}, \dots)$, and decides on its course of action only after weighing the simulated outcomes, whether at decision time or during training.
This requires the world model to simulate not only a single action correctly, but a sequence of actions, each starting from the state the previous one produced.
Beyond the capabilities exercised by single actions, we therefore examine the simulation of action sequences as a whole.

We construct a second set of test cases in the same format as before, a starting observation in optional image and text and an action in text, with two differences.
First, every test case is an everyday chore that fits within 20 seconds, such as taking a dog out, seen through the eyes of a person whose hands and forearms are the only visible parts of the body.
Second, the action text describes a plan of 3 to 6 steps that must be carried out sequentially in order to complete the plan (e.g., turn right, pick up the hammer, turn back, and put the hammer in the box). We assume the actor does not know the stepwise or cumulative outcome of these steps (or even know, would hold to itself and compare with the actual rollout), and leaves them for the world model to answer -- a scenario we stick to for the agent-world relationship.
Because existing multi-step benchmarks rarely require such turns, we wrote the 100 plans ourselves; they span 456 steps, 4.6 per plan on average, and each is provided to the model both with a starting image and with text alone, resulting in 200 simulations of up to 20 seconds per model.
Every plan is designed to succeed when executed faithfully.
This isolates the question we ask here, as a simulator that cannot faithfully simulate a correct plan cannot be relied on to distinguish good plans from bad ones.

Each step is judged separately, so that a simulation that jumps to the final outcome without carrying out the intermediate steps fails them.
Steps involving camera motion are checked against camera trajectories estimated from the video.
We report two measures: \emph{step completion}, the fraction of steps completed per simulation, and \emph{full-plan completion}, the fraction of simulations in which every step is completed.
We study the same models as we did for world simulation, under the same settings.

Simulating every step of a plan is much harder than simulating most of them (Fig.~\ref{fig:main-results}b).
Across models, step completion ranges from 38.5\% to 75.1\%, and full-plan completion from 4.0\% to 34.0\%. PAN~2 achieves the highest point estimate of full-plan completion among the models we studied, 34.0\%, with 75.1\% step completion.
Even for the highest-scoring models, then, about two in three simulated plans contain at least one step that is not carried out.

Step completion changes little across Stages~I--III (70.8--72.0\%).
It rises to 74.1\% with reward learning in Stage~IV, the same stage in which causal coherence and entity persistence, on which multi-step plans depend, reach their highest scores.
The merged model completes 75.1\% of steps, the highest among PAN~2's checkpoints.
Full-plan completion follows the same course, from 27.7\% after Stage~I to 34.0\% for the merged model.

Full-plan completion is lower than step completion because errors compound: a plan succeeds only if every one of its steps does.
Across models, the two measures follow an intuitive compounding curve ($R^2 = 0.986$, Fig.~\ref{fig:main-results}c; Methods), along which full-plan completion rises increasingly steeply as steps become reliable.
On this curve, raising step completion from 75\% to 90\% would raise full-plan completion from about one-third to two-thirds, and however failures are distributed across simulations, its minimum would be about 48\% (Methods).
Error compounding thus explains why current models' full-plan completion is low, but it does not keep it low as each simulated step becomes more reliable.

In this study,
a world model
receives a whole plan at once and simulates it in a single pass, without observing intermediate results or revising the plan.
This open-loop setting tests whether the model can faithfully simulate a plan, instead of the full loop of simulative reasoning.
Evaluating how faithfully a world model simulates failure, as well as success, requires higher simulation fidelity than current models offer, and we leave it to future work.

\section{Generalization Beyond the Training Data}
\label{sec:generalization}

\begin{figure*}[!t]
	\centering
	\includegraphics[width=\textwidth]{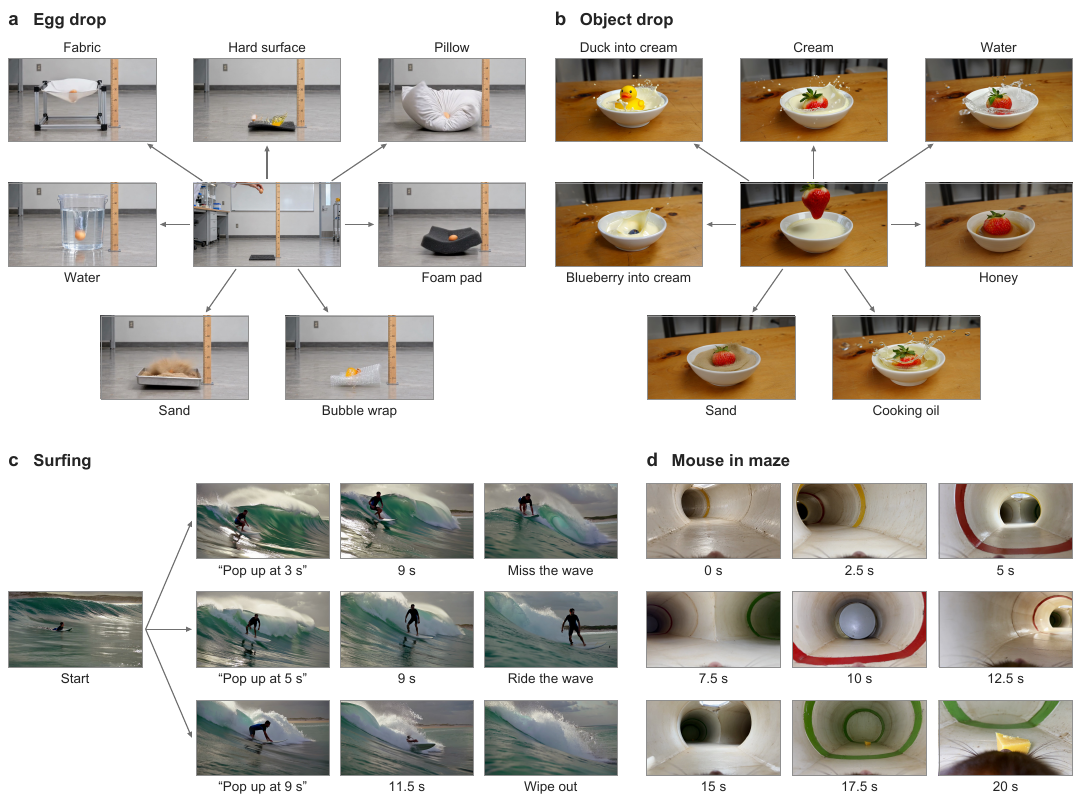}
    \caption{\textbf{Simulating beyond the training data.} Each panel varies one factor of the starting state or action while holding the rest fixed. (a) An egg dropped from different heights onto different surfaces breaks, bounces or stays whole (7 of 30 conditions shown). (b) A rubber duck, a strawberry and a blueberry dropped into different liquids and into sand, generated with an orbiting camera. (c) A surfer popping up at 3, 5 or 9 seconds into the same wave. (d) A mouse searching a maze for cheese, a domain for which we found no close match in the training data. }
	\label{fig:simulation_branches}
\end{figure*}

To support simulative reasoning, a world model should be able to answer the counterfactual question ``what if?''
The alternatives an agent
 compares
  are, almost by definition, situations it has not encountered (e.g., the same egg dropped onto a different surface, the same wave caught a few seconds later).
A world model that could only reproduce the situations in its training data would be of little help in comparing them.
Because a general-purpose world model is trained on data curated for broad coverage of the real world, generalization for it is not simply a matter of a domain held out from training.
We instead consider it in two senses.
First, for phenomena well represented in training, a world model should simulate them under new combinations of conditions, varying physics, space and time in ways the training data do not cover.
Second, for domains the data happen to miss, or for which real footage is hard to obtain, it should still produce coherent simulations with no close match in the training corpus.

We examine both through case studies in which one factor is varied while the rest of the starting state is held fixed, and ask whether the simulated outcome changes in the physically expected direction (Fig.~\ref{fig:simulation_branches}).
To assess novelty, we search for our text prompt in the training corpus based on caption embeddings (Supplementary Information~\ref{app:case-studies}).

\subsection*{Simulating new combinations of known phenomena}

In the first example (Fig.~\ref{fig:simulation_branches}a), an egg is dropped from three heights onto ten surfaces.
The outcomes cluster in ways one would expect:
on hard surfaces the egg breaks, more violently from greater heights;
on soft surfaces such as a pillow, a sponge or fabric, it bounces to different heights;
on moderately cushioning surfaces such as cardboard or water, it stays whole when dropped from lower heights and breaks when dropped from greater ones.
The model also makes mistakes: dropped onto a single layer of bubble wrap on the ground, the egg bounces as if from a much softer surface.
The original query from the human or agent gives the action, not the outcome. The conceptual reasoner (LLM) determines the high-level outcome for most of the egg cases.

In the second example (Fig.~\ref{fig:simulation_branches}b), a rubber duck, a strawberry, and a blueberry are dropped into water, cream, honey, cooking oil and sand.
Outcomes depend on both the object and the medium.
The duck floats, and the blueberry raises a smaller splash than the strawberry.
Honey yields more slowly than cream, and cream more slowly than water, while sand barely splashes at all.
In every case the camera orbits the scene as specified while the drop unfolds.

In the third (Fig.~\ref{fig:simulation_branches}c), the starting state is fixed and only the timing of an action changes: a surfer pops up 3, 5 or 9 seconds into the simulation.
Popping up at 3 seconds, before the wave has given the board enough speed, leaves the surfer behind as the wave passes; at 5 seconds the surfer catches the wave; at 9 seconds the wave overwhelms the surfer, who is wiped out.

For all three examples, the training corpus contains the underlying phenomena: eggs breaking, strawberries and blueberries falling into milk and water, and surfing.
It does not, however, contain this range of landing surfaces, these combinations of objects and media, drops filmed with an orbiting camera, or variations in the timing of a surfer's pop-up.

\subsection*{Simulating unknown domains}

Some domains have no close counterpart in the training data at all.
Prompted with a mouse searching for a piece of cheese in a tunnel-like maze (Fig.~\ref{fig:simulation_branches}d), PAN~2 simulates the search over 20 seconds: from a first-person view, a mouse walks down a tunnel, explores the left branch of a Y-intersection, turns back to explore the right branch upon finding no cheese, and finds the cheese at the end of the tunnel; the interaction is coherent despite minor artifacts in the angle between tunnels and in the color consistency of one tunnel ring.
We find no instance resembling this example in the training corpus, although due to the sheer volume, we cannot rule out completely the possibility that a similar video exists.
For other models, we cannot investigate this phenomenon, since their training corpora are not described.

\subsection*{Implication of the capabilities}

Both capabilities broaden what a world model can be used for.
Simulating familiar phenomena under new conditions lets an agent, or a person, assess how a change to one factor alters an outcome, by running parallel simulations that differ only in that factor.
Simulating unfamiliar domains makes it possible to visualize situations that would be impractical to set up in the real world, and may help an agent prepare for situations it has not encountered.

\section{Towards Agent Learning in World Simulation}
\label{sec:agent-learning}

\begin{figure*}[!t]
	\centering
    \includegraphics[width=\textwidth]{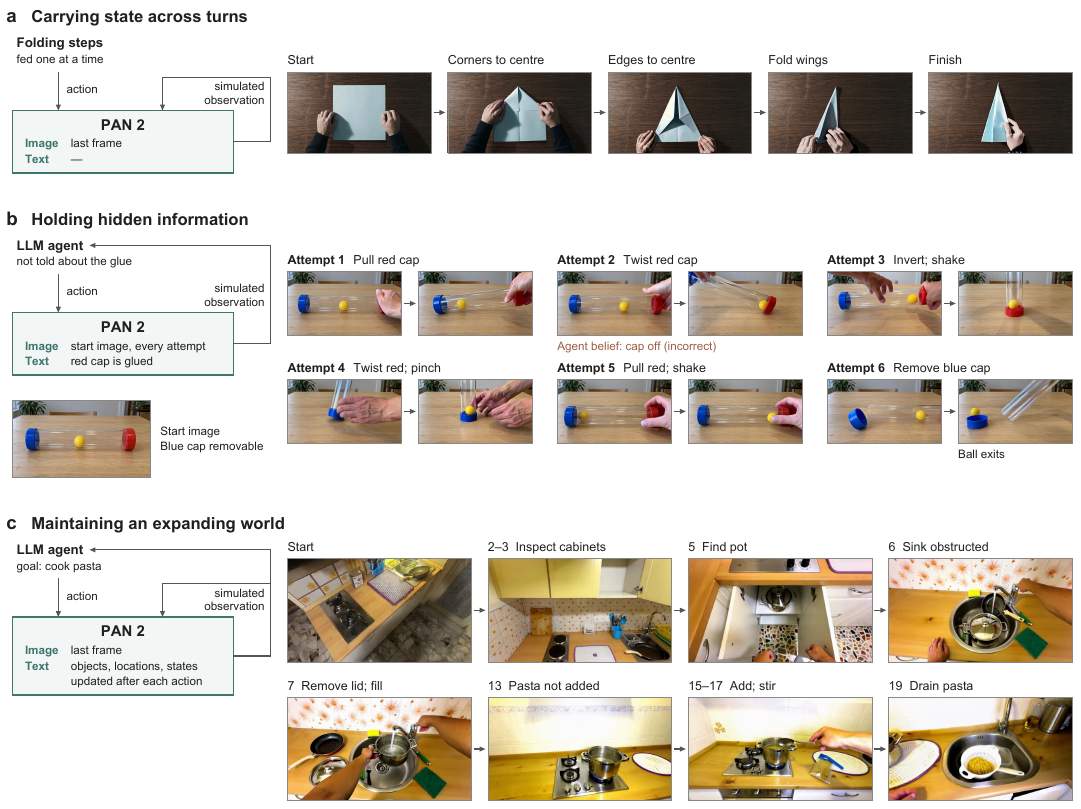}
    \caption{
    \textbf{Agents interacting with simulated worlds.}
    (a) A person folds a paper plane in simulation; the last frame of each clip starts the next step. (b) An agent tries to get a ball out of a tube whose red cap is glued shut, a fact given to PAN~2 in text but not to the agent; it succeeds on its sixth attempt by removing the blue cap. (c) An agent cooks a pot of pasta in a kitchen it has never seen, over 19 actions, while a lightweight text memory stores where objects are and what state they are in. Panels show selected frames.}
	\label{fig:agent-interaction}
\end{figure*}

One of the most powerful forms of simulative reasoning is for an agent to interact with a world model and learn from the experience by trying actions, observing their simulated consequences, and adjusting their behavior, without the cost or risk of acting in the real world.
This places three demands on a world model beyond simulating a given plan.
It must carry the state of the world across turns, so that each action starts where the last one ended.
It must hold information the agent does not have, and reveal it when triggered by the agent, just like the real world.
And it must extend and maintain the world as the agent explores it, keeping what lies behind each door once it has been opened.
A simulator that lacks any of these offers limited value to a real agent, which faces all three.

We examine these demands through three case studies, where the first two each isolate a single demand and the last case study combines all three (Fig.~\ref{fig:agent-interaction}).
Across them, the main difference is what the world model
must maintain: the visual state in image, hidden information expressed in text, or a growing world with both.
Specifically, the first case involves feeding PAN 2 a sequence of actions; the second and third include an autonomous agent that acts without human intervention (implementation details in Supplementary Information~\ref{app:case-studies}).

The first case isolates the simplest demand of carrying state across turns when nothing is hidden.
Starting from an image of a sheet of paper on a table and a list of folding steps (each an action instruction), we give PAN~2 the image and the first step, take the last frame of the simulated clip as the starting image for the next step, and continue until a paper plane is folded (Fig.~\ref{fig:agent-interaction}a).
The full sequence of folds carries forward correctly, demonstrating that the last frame suffices to maintain state when nothing is hidden.
Passing the last frame from one clip to the next, however, cannot carry any information the image does not show.

The second case adds hidden information.
A ball sits in a tube closed by two caps, a blue one that is removable and a red one that is glued shut.
The glue is not visible in the initial image, but instead stated only in the text given to the world model, never to the agent.
An AI agent built on Kimi-K3 is asked to get the ball out (Fig.~\ref{fig:agent-interaction}b).
In each attempt, it proposes a short plan, PAN~2 simulates the plan from the starting image, and the agent judges from the clip whether it has succeeded.
If not, it tries again, with all its previous plans, clips and judgements available, in a setup similar to Reflexion~\citep{shinn2023reflexionlanguageagentsverbal}.
The agent first tries the red cap, pulling and then twisting it.
Misreading the second simulation, it concludes that the cap has come off and keeps trying to force the ball out of that end, at one point shaking the tube, before turning to the blue cap on its sixth attempt, which succeeds.
Through simulated interaction alone, the agent discovers a property of the world it could not see. The image, the visual-based state, carries the majority of the information about the puzzle, specifically that a ball is trapped in a tube with two different-colored caps.
The text-based state carries what the image cannot, which is one role of the conceptual half of PAN~2's representation.

The third case combines all three demands.
An agent built on Kimi-K3 is asked to cook a pot of pasta in a kitchen it has never seen, starting from a single image of a counter with a sink and a two-burner stove (Fig.~\ref{fig:agent-interaction}c).
We keep a text memory of the objects, their places and states, and use the last frame of each clip as the starting image for the next.
For each action the agent requests, PAN~2's conceptual reasoner references this memory while writing text-based next states, and updates it after each simulation turn.
The acting agent sees only the goal and the frames,
not this memory (Supplementary Information~\ref{app:case-studies}).

Over 19 actions and 294 seconds of simulated video, the agent searches the kitchen and cooks the pasta.
It searches many cabinets to find the pot under the stove.
When it fills the pot in the crowded sink and the water runs into a frying pan instead, it moves the pan aside, removes the lid and fills the pot again.
After lighting the burner, it searches again to find the spaghetti.
It then stirs the pasta, lets it cook, returns to the sink for the colander it saw earlier, and drains the pasta.
Across visits, objects stay where the agent first found them.
PAN~2 provides an environment where the agent can explore, such as to find the pot and the spaghetti, and solve physical problems, such as filling the pot in a crowded sink.
The simulated world is nonetheless far from fully consistent.
The agent's own review of the episode judged the goal achieved but noted that the kitchen's tiles, stove and sink change between clips: the two-burner stove becomes a four-burner one, and the sink changes shape.
Pasta also appears in the colander before draining is shown, and the simulation does not represent realistic cooking time or doneness.

The level of detail preserved by the model, however, is sufficient in this case to allow the agent to complete its task.

Together, the three case studies show that a world model can already simulate closed-loop interaction, hidden information, and an expanding world.
Each depends on how the state of the world is represented: the last frame suffices when nothing is hidden, whereas hidden properties and an expanding world require state in text that the conceptual reasoning backbone can read and update.

This division of responsibilities is integral to the GLP architecture, where the LLM manages the conceptual state with discrete, possibly hidden facts, the LDM handles the perceptual aspects outside the LLM's capabilities, and both components are trained via generative supervision.
On the other hand, models that represent the world solely through the perceptual latents of the current observation and are supervised by next-latent prediction~\citep{lecun2022path} would face structural difficulties, as their representation cannot recover information that is hidden or no longer in view (e.g., the glued cap, the objects behind closed cabinets), and the lossy objective provides no guarantee that the latents retain all action-relevant distinctions, especially for actions outside the training distribution~\citep{xing2025critique}.
To our knowledge, no world model architecture other than GLP has yet been shown to support all the three cases.



The perceptual inconsistencies in long-horizon simulation also point at a limitation of the current state representation.
The text-based state keeps track of which objects exist, where they are and even how they look, yet the kitchen's appearance drifts from clip to clip, because the perceptual backbone sees no more of the past than the last frame.

From the standpoint of task completion, not all drift is equally costly, as the agent completed the cooking task despite the changing tiles and stove, because those changes did not affect the objects or states the task depended on.
When long-term perceptual consistency is desired, however, a stateful representation for long-horizon simulation will need its conceptual and perceptual parts to be maintained in coordination.

The perceptual backbone, too, must take part, for instance through a short-term cache of prior video context that it attends to alongside the text-based state, which can be achieved by post-training the LDM~\citep[e.g.,][]{chen2024diffusion,panteam2026panworldmodelgeneral}.
Building this coordination into the model and its training, and training agents on the experience a world model provides rather than only letting them learn within a single episode, are the next steps towards agent learning in world simulation.

\section{How the Parts of a World Model Work Together}
\label{sec:ablations}

Coupling conceptual and perceptual reasoning divides the work of simulation between an LLM, an LDM, and the text-based state that connects them.
To understand how these parts work together,
we ablate each backbone in turn while holding the other fixed.

\subsection*{Ablation of the Conceptual Backbone}

\begin{figure*}[!t]
\centering
\includegraphics[width=\textwidth]{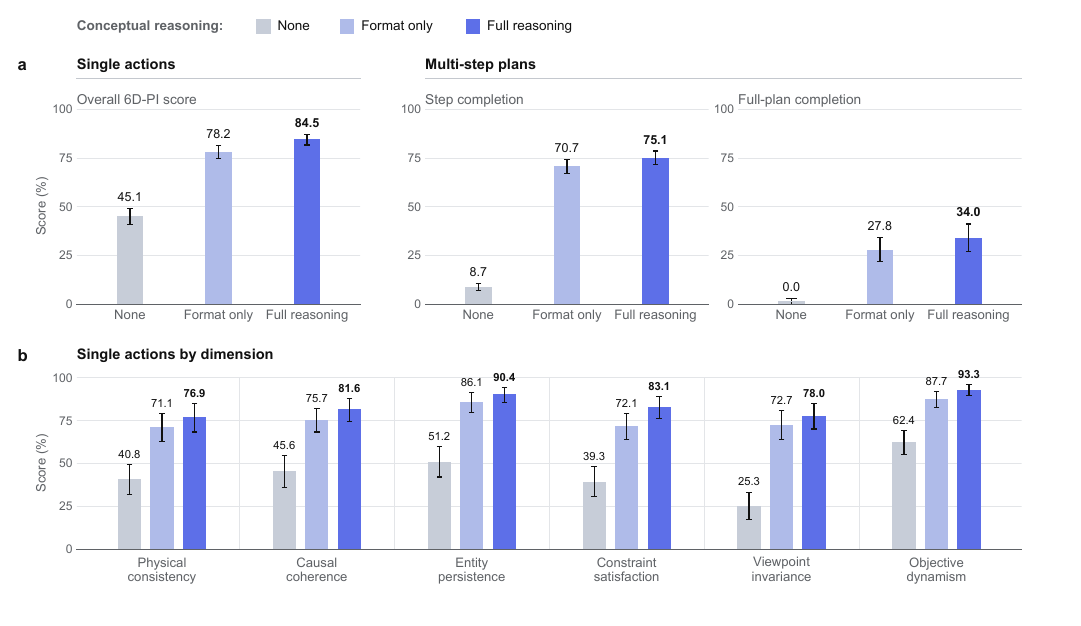}
\par\vspace{2mm}
\includegraphics[width=\textwidth]{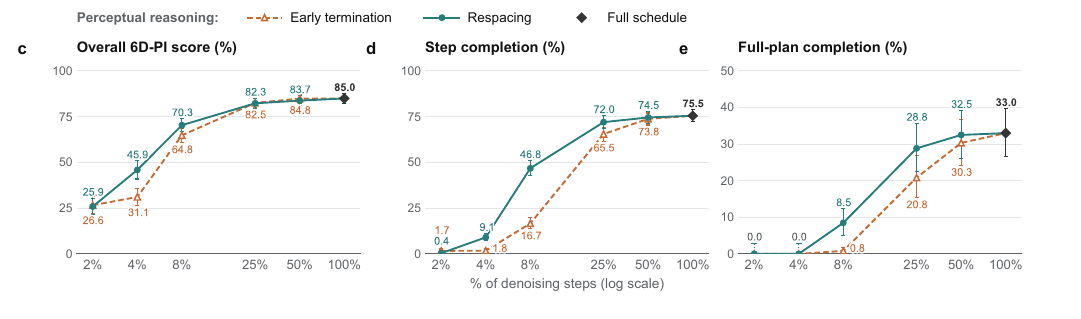}
\caption{\textbf{Ablation of the conceptual and perceptual backbones.} \textbf{a,b,} Ablation of the conceptual backbone. A fixed PAN~2 LDM receives the action instruction without any LLM processing (none), the input converted by PAN~2's LLM into its structured format without
instructions to reason (format only), or the next conceptual state predicted by the LLM (full reasoning, the released PAN~2 configuration of Fig.~\ref{fig:main-results}).
\textbf{a,} Overall score on single actions, and step completion and full-plan completion on multi-step plans.
\textbf{b,} Single-action scores on the six dimensions.
\textbf{c-e,} Ablation of the perceptual backbone.
With PAN~2's LLM text-based states fixed, we reduce the LDM's denoising steps by early termination (stopping partway through the original schedule) or by respacing (covering the full denoising interval in fewer steps). Diamonds mark the full schedule.
\textbf{c,} 6D-PI score.
\textbf{d,} Step completion.
\textbf{e,} Full-plan completion.
Bars (a,b) and points (c--e) show mean scores over three seeds, and
error bars show 95\% percentile bootstrap confidence intervals over
the 204 base cases (single actions) or the 100 plans (multi-step),
with paired modalities and all seeds retained within each case
($N{=}10{,}000$ shared resamples).
In~a, the dashed whisker on zero full-plan completion under none
shows the one-sided 95\% upper bound over 100 plans, 3.0\%.
The same bound applies to zero full-plan completion in~e.
Supplementary Information~\ref{app:perceptual-compute} gives the
numerical results and protocol.}
\label{fig:reasoner-caption}
\end{figure*}

PAN~2's conceptual reasoning backbone, an LLM, predicts the next world state as a structured, time-grounded description in the format the LDM was trained on.
To isolate its contribution, we test three levels of conceptual reasoning with the same fixed LDM:
\emph{none}, in which the action instruction is passed directly to the LDM without any LLM processing;
\emph{format only}, in which the LLM converts the input into the structured format without instructions to reason about the world; and
\emph{full reasoning}, in which the LLM reasons about how the world will evolve before writing the next state.
Without any conceptual reasoning (\emph{none}), the LDM alone scores 45.1\% across 6D-PI capabilities, completes 8.7\% of multi-step plan steps, and completes no plan in full (Fig.~\ref{fig:reasoner-caption}a).
It thus retains about half of its single-action capability but almost none of its ability to simulate plans.
In other words, the model's dependence on conceptual prediction is greatest when simulation requires coordinated state changes across an entire plan.

Format-only processing by the LLM recovers most of the gap in 6D-PI score (78.2\%) and much of step completion (70.7\%), with full-plan completion rising to 27.8\%.
Constructing the text-based state is thus already a substantial contribution of the conceptual backbone, even without an explicit instruction to reason.

Full reasoning improves further from 78.2\% to 84.5\% on 6D-PI, from 70.7\% to 75.1\% step completion, and from 27.8\% to 34.0\% full-plan completion.
Viewpoint invariance shows the largest gain from no conceptual reasoning to full reasoning, rising from 25.3\% with none to 72.7\% with format-only processing and 78.0\% with full reasoning (Fig.~\ref{fig:reasoner-caption}b), consistent with camera motion being explicitly planned in the text-based state.
Together, these results identify two contributions of the conceptual backbone: constructing and organizing the text-based state, and reasoning about its evolution to improve the accuracy of a fixed perceptual backbone.

\subsection*{Ablation of the Perceptual Backbone}

The previous ablation shows that the LDM alone is insufficient for world simulation; the symmetric question is whether the conceptual backbone alone suffices.
Because PAN~2's decoder operates on the LDM's latent space, the perceptual backbone cannot be removed entirely; instead, we ablate it by progressively reducing its denoising steps while keeping the LLM's text-based states fixed (Fig.~\ref{fig:reasoner-caption}c--e).
Just as the previous ablation tested three levels of conceptual reasoning, we test different levels of perceptual reasoning through two reduction strategies: \emph{early termination}, which follows the original denoising schedule but stops early for decoding, and \emph{respacing}, which covers the full denoising interval in fewer steps (Supplementary Information~\ref{app:perceptual-compute}).

\emph{Early termination} reveals what the perceptual backbone accomplishes at each point in its original denoising schedule.
At about 25\% of the schedule, overall 6D-PI score is 82.5\% and step completion 65.5\%,

down from 84.5\% and 75.1\% at the full schedule;
at about 8\%, step completion falls to 16.7\% and full-plan completion to zero.
This suggests that the perceptual backbone builds up the detail for faithful simulation progressively across denoising, and interrupting that process degrades simulation even though the text-based states are unchanged.

\emph{Respacing} tests whether the same perceptual reasoning can be compressed into fewer steps by spreading them across the full denoising interval.
At 25\% of the original denoising steps, respacing retains 82.3\% 6D-PI score, 72.0\% action step completion and 28.8\% full-plan completion;
at the same denoising budget, early termination scores a similar 82.5\% on 6D-PI but completes only 65.5\% of steps and 20.8\% of full plans.

At about 8\%, step completion is 46.8\%, higher than 16.7\% for early termination but already clearly declining; at 4\%, even respacing falls to 9.1\% step completion and zero full-plan completion.
This suggests the perceptual backbone can compress its reasoning to a certain degree, but not arbitrarily.

As with the conceptual ablation, multi-step simulation is more sensitive to the perturbations. Specifically, perceptual reasoning can be substantially weakened before single-action scores decline, but plans, which compound errors across steps, degrade earlier and more sharply.
Together, the ablations show that while the conceptual backbone specifies how the world should evolve, that specification alone does not produce faithful simulation when the perceptual backbone cannot fill in the details it leaves open.

\section{Discussion}

The purpose of a world model is to support simulative reasoning.
We have therefore studied world models by the standards of faithful simulation rather than by the beauty of their videos, firstly by evaluating along the
six dimensions of physical intelligence (6D-PI): physical consistency, causal coherence, entity persistence, constraint satisfaction, viewpoint invariance, and objective dynamism; then by measuring
the completion of multi-step plans.
Because we built PAN~2 from its data onward, we could ask how the capabilities emerge across stages of training, how the conceptual backbone contributes, and what is lost from ablating the perceptual reasoning.
The analyses so far have not only led to insights on how a world model of the physical world should be built, but also left behind open technical problems that point to future directions.

Simulating the physical world requires both conceptual and perceptual reasoning.
Without an LLM, PAN~2's LDM scores about half as high on 6D-PI and completes no plan in full.

Conversely, given the same LLM output, ablating the LDM's denoising compute leaves only about a quarter of 6D-PI and none of full-plan completion.

The LLM that contemporary video generation models use as a prompt rewriter~\citep{nvidia2026cosmos3omnimodalworld} is thus better understood as a conceptual reasoning backbone, and a world model can build on LLMs~\citep{xing2025critique} and benefit from their progress; training the LLM backbone for better temporal, physical, and spatial reasoning is a natural next step.

On the perceptual side, the controllability, stability, and efficiency of LDMs can be further improved, although fundamentally, a new type of model for perceptual inference, or one that merges conceptual and perceptual reasoning into a single model, could represent more sustainable solutions to all three problems.

The text-based state is a design choice with measurable consequences, as it decides what the LLM explicitly models and what the LDM implicitly fills; indeed, richer, structured states help the LDM learn (Supplementary Fig.~\ref{fig:caption-recipes}).
The same division applies to actions, since text specifies dexterous actions, such as where each finger meets an object, only coarsely, whereas embodied uses such as robotics need both.
Learning where the boundary should lie, and adding action interfaces at the perceptual level alongside text, are closely related next steps.

Long interactions need stateful representations kept in both parts of the world model. In the kitchen simulation example, a text memory kept objects in place while the room's appearance drifted, which a perceptual memory built into the model, such as a cache of prior video context or an explicit three-dimensional scene~\citep{chen2024diffusion,panteam2026panworldmodelgeneral}, would address.
A world model should also remain separate from the agents it serves, holding facts an agent lacks and revealing them only through simulated consequences, whereas a model trained jointly with a policy~\citep{lu2026world,ye2026world} risks replacing faithful simulation with wishful simulation~\citep{eysenbach2022mismatched,xing2026cam}.

Several questions lie beyond this study: whether specialized world models inherit more from a general-purpose base model than they would learn on domain-specific data~\citep{bommasani2021opportunities}; how embodied reasoning should incorporate touch, sound and proprioception; and how next-world prediction scales to more modalities and longer simulation horizons.
World models that predict only in a latent space, such as JEPA-style models~\citep{lecun2022path,assran2025v,mur2026v}, are typically evaluated on visual understanding or by deploying their latents in downstream control; placing them on the same simulation tests as done in this study would require decoders that render their predicted states, as some recent work already trains~\citep{zhou2024dino}, and would allow a direct comparison only after that.
Just as much of lingual intelligence is captured by predicting the next word, the path towards physical intelligence runs through predicting the next world reliably, with conceptual reasoning that draws on what language records about the world, perceptual reasoning learned from how the world looks and moves, and training signals and tests that supervise both against the world itself.

\bibliographystyle{plainnat}
\bibliography{references-fixed,cwm_ref-fixed}

\section{Methods}

\subsection{Model Architecture}
\label{subsec:model_architecture}

The GLP architecture (Eq.~\ref{eq:pan2}) decomposes each state into conceptual and perceptual parts. We describe each component below.

\subsubsection{Conceptual reasoning backbone}

The conceptual encoder $h_{\mathrm{sem}}$ and backbone $f_{\mathrm{LLM}}$ are two parts of one pretrained multimodal LLM, used without further training.
It predicts the next conceptual state as
\begin{equation}
\hat s^{\mathrm c}_{t+1} = \bigl(c,\ (e_1,\dots,e_E)\bigr) \sim p_{f_{\mathrm{LLM}}}\bigl(\cdot \mid \hat s^{\mathrm c}_t,\ a_t\bigr),
\qquad e_k=\bigl(\iota_k,\ \delta_k,\ \alpha_k,\ \kappa_k\bigr).
\label{eq:conceptual-state}
\end{equation}
The global context $c$ describes the initial scene, the key subjects and their attributes, and the point of view of the observer.
Each event $e_k$ occupies a time interval $\iota_k$ within the requested duration and carries a description $\delta_k$ of the changes in scene and object state it brings about, the actions $\alpha_k$ within it, each timed and bound to an actor, and the camera motion $\kappa_k$, which describes only the movement of the viewpoint.
We use Kimi-K3~\citep{kimiteam2026kimik3openfrontier} for all studies unless stated otherwise; prompts and settings are given in Supplementary Information~\ref{app:world-reasoner-settings}.

\subsubsection{Perceptual backbone}
\label{subsec:ldm_implementation}

The perceptual backbone is a latent diffusion model implemented as a multimodal diffusion transformer~\citep{esser2024scaling} with 29.2 billion trainable parameters, which we train from scratch
(Extended Data Fig.~\ref{fig:ldm-architecture}; Supplementary Table~\ref{tab:ldm-model-card}).
We use the frozen Wan2.2 VAE~\citep{wan2025}.
The VAE encoder thus plays the role of $h_{\mathrm{VAE}}$: the latents of the starting frame, when one is given, are the current perceptual state $\hat s^{\mathrm p}_t$, which we also write $z^{\mathrm{ref}}$.
At each denoising step, the transformer concatenates the noisy video latents, $z^{\mathrm{ref}}$ and a binary mask marking the starting frame along the channel axis and embeds them as video tokens.

The conceptual state reaches the LDM through a text pathway.
A frozen Qwen3.5-9B language model~\citep{qwen3.5} encodes $\hat s^{\mathrm c}_{t+1}$, and a trainable text projector maps its features into the transformer's token space.
The transformer stacks $L=44$ joint-attention blocks.
Writing $x^{\mathrm v}_\ell$ and $x^{\mathrm c}_\ell$ for the video and text tokens entering block $\ell$ at diffusion time $\tau$, each block updates both,
\begin{equation}
\bigl(x^{\mathrm v}_{\ell+1},\,x^{\mathrm c}_{\ell+1}\bigr)=B_{\theta,\ell}\bigl(x^{\mathrm v}_{\ell},\,x^{\mathrm c}_{\ell};\,\tau\bigr),\qquad \ell=0,\dots,L-1,
\label{eq:mmdit-block}
\end{equation}
with modality-specific query, key, value and output projections and feed-forward networks.
Queries from either modality attend jointly to the keys and values of both, and gated residual connections update each modality separately.
Video tokens are modulated by the diffusion time and carry three-axis rotary position embeddings; text tokens are modulated by a learned, time-independent vector.
We adopt joint attention for its uniformity and extensibility, since with $M$ modalities, connecting every directed pair of modalities explicitly would require $M(M-1)$ cross-attention paths, whereas joint attention applies one rule, concatenating the tokens of all modalities, so that adding a modality only requires defining its tokens and their positional information.
After the final block, the text tokens are discarded, and an output head maps each video token to its share of the flow velocity in latent space.
Further implementation details, including normalization, activations and feed-forward widths, are given in Supplementary Information~\ref{app:ldm-supp}.

\subsubsection{Learning objectives}
\label{subsec:ldm_training}

The LDM is trained by flow matching on video latents~\citep{lipman2022flow}.
For clean target latents $z$, Gaussian noise $\epsilon\sim\mathcal N(0,I)$ and diffusion time $\tau\in(0,1)$, we form $z_\tau=(1-\tau)z+\tau\epsilon$ and train the velocity prediction $u_\theta$ with
\begin{equation}
\mathcal L_{\mathrm{FM}}=\mathbb E_{z,\epsilon,\tau}\Bigl[\bigl\lVert u_\theta\bigl(z_\tau,\,z^{\mathrm{ref}},\,\hat s^{\mathrm c}_{t+1},\,\tau\bigr)-(\epsilon-z)\bigr\rVert_2^2\Bigr],
\label{eq:flow-matching}
\end{equation}
where $z^{\mathrm{ref}}=\hat s^{\mathrm p}_t$ denotes the latents of the starting frame when one is given, and the denoised latents form the predicted perceptual state $\hat s^{\mathrm p}_{t+1}$.
Every frame of the target clip is noised and contributes to the loss, including the starting frame, whose clean latents are supplied separately through the conditioning channels.
The loss is computed only on video latents, but it also trains the text pathway through joint attention; no loss is imposed on the conceptual state itself.
We sample $\tau$ from a shifted logit-normal distribution in Stages~I and~II, which emphasizes intermediate noise levels, and uniformly in Stage~III, which covers the full range used at inference.

In Stage~IV, we refine the LDM by reward-based learning, using feedback on its own generated videos instead of the flow-matching loss.
Following the generative scoring principle of RewardDance~\citep{wu2025rewarddancerewardscalingvisual}, we use the probability that a video reward model answers ``yes'' to a question as a continuous, differentiable score, taken over the full vocabulary rather than as a hard answer; unlike RewardDance's comparison with a reference video, our scoring is pointwise.
We use VisionReward~\citep{xu2024visionrewardfinegrainedmultidimensionalhuman} with 13 questions selected from its checklist, covering object structure and persistence, motion, composition and perceptual detail, and score a video $V$ as
\begin{equation}
R(V)=\frac{\sum_{q\in\mathcal Q}w_q\,\bar p_\phi(\text{yes}\mid V,q)}{\sum_{q\in\mathcal Q}w_q},
\label{eq:video-reward}
\end{equation}
where $\mathcal Q$ is the set of questions, $w_q$ the weight of question $q$ and
$\bar p_\phi$ the probability of the affirmative token, averaged over the views scored for that question; $w_q$ are VisionReward's published weights, renormalized over the 13 questions, which are listed in Supplementary Information~\ref{app:training-details}.
The score does not depend on the text-based state.
For each update, we generate most of the denoising trajectory without gradients and make one gradient-enabled prediction of the clean latents; the frozen VAE decodes this prediction, and the reward's gradient flows through the decoder and the reward model into the LDM.

\subsection{Text-based state annotation}
\label{sec:tgc-overview}

We annotate every training video in the structured format of Eq.~\ref{eq:conceptual-state}, the format the LLM writes at inference.
Separating camera motion from the motion of subjects and objects is the main difficulty: when the camera pans left, stationary objects move right in the image, and the two kinds of motion often occur together.
A single pass of a vision-language model (VLM) tends to miss camera motion or attribute it wrongly (Supplementary Information~\ref{subsubsec:camera_ablation}), so we annotate each clip in three stages
(Extended Data Fig.~\ref{fig:discrete_representation}a).

First, a VLM proposes a small number of chronological events, with boundaries at changes in activity, interaction or visible scene state, and finer, possibly overlapping action slots within each event; this temporal scaffold fixes only when things happen.
Second, we estimate the camera pose of sampled frames with VGGT-$\Omega$~\citep{wang2025vggt,wang2026vggtomega} and compute, for each pair of adjacent frames, the rates of yaw, pitch and roll, the translation of the camera center and the change in focal length; each rate above an empirically calibrated threshold yields a motion label, and the labels are condensed into a camera plan for each event.
Third, a second VLM pass writes the full structured caption: it keeps the given events, checks each action against the frames, attributes it to a visible actor and drops actions it cannot confirm; the camera field of each event is then taken from the geometric camera plan rather than from the VLM.

All captioning requests use Qwen3.6-27B. A separate request describes only the initial frame as a still image, and the resulting captions contain up to 1,536 tokens.
Sparse frame sampling, occlusion, deformation and errors in camera estimation can still affect the annotations, so we treat them as structured supervision derived from observation, not as calibrated physical ground truth.
Sampling settings, thresholds and output fields are given in Supplementary Information~\ref{app:world-caption-details}.

\subsection{Data Curation and Concept Balancing}
\label{sec:data}

We curate video and image data that encode coherent physical dynamics while covering a broad range of scenes, subjects and interactions.
To provide the continuous perspective needed for stable simulation, we split videos at scene cuts.
We remove clips of low visual quality, clips with accelerated playback, and clips with little temporal information, such as a person speaking to the camera without gesturing, or a camera panning over a still image.

Because common content with little movement, such as people talking, dominates the filtered pool, we balance the training data over the concepts a world model needs to learn.
Within each concept, we select clips for perceptual quality, motion and semantic diversity.
From Stage~II onward, we exclude gameplay, animation, computer-generated renders and simulation.

\subsection{Training}
\label{sec:training}

We train the LDM in four stages that progressively increase the perceptual and dynamical complexity of the data
(Extended Data Table~\ref{tab:training-stages}; Extended Data Fig.~\ref{fig:training-stages-curves}a).
Training losses and rewards are shown in Extended Data Fig.~\ref{fig:training-stages-curves}b,c.

We optimize the LDM with a hybrid of Muon~\citep{jordan2024muon} and AdamW~\citep{loshchilov2019decoupled}.
Muon builds matrix-valued updates by approximately orthogonalizing the momentum of the gradient with five Newton--Schulz iterations, $\mathcal N_5$.
We apply it per attention head to the query, key and value projections, matrix-wise to the other large linear layers, and use AdamW for the remaining parameters; in the 29.2-billion-parameter LDM, 24.7 billion parameters are updated by Muon and 4.5 billion by AdamW.
For a projection $p\in\{\mathrm Q,\mathrm K,\mathrm V\}$ whose rows stack the matrices of its $H$ heads, $W^p=[W^{p,1};\dots;W^{p,H}]$, with momentum update $S^p=[S^{p,1};\dots;S^{p,H}]$, per-head Muon orthogonalizes each head's block separately:
\begin{equation}
\mathcal N^{\mathrm{head}}_5(S^p)=\bigl[\mathcal N_5(S^{p,1});\dots;\mathcal N_5(S^{p,H})\bigr].
\label{eq:muon-per-head}
\end{equation}
This differs from orthogonalizing the whole matrix, which couples heads through their cross-head Gram terms; aligning the update with the structure of multi-head attention lets each head's singular spectrum be normalized on its own.
The same partition applies to the modality-specific projections of the joint-attention blocks and to the text projector.

Attention uses FlashAttention-4~\citep{zadouri2026flashattention}; the modality-specific projections and feed-forward networks use grouped matrix-multiplication kernels from QuACK~\citep{guo2025quack}, and other element-wise and normalization operations are fused with Triton kernels.
We use hybrid-sharded data parallelism~\citep{zhao2023pytorch} with dynamic batching and sequence packing to balance computation across videos of different durations.
The text encoder is replicated on each device, and video latents are precomputed with the VAE encoder before training.

\subsection{Studying simulation capabilities}
\label{sec:world-model-eval}

\subsubsection{Single-action test cases}
\label{sec:tier1-capabilities}

We built the single-action test cases around the
six dimensions
described in the main text.
We collected 5,198 unique prompt strings from 11 public video-generation and world-model benchmarks, with Physics-IQ and Physics-IQ Verified sharing prompts, and separately reviewed capability coverage in the papers of 15 benchmarks, six overlapping with the prompt sources. Coverage concentrates on physical properties and rarely tests causal coherence or
viewpoint invariance
directly; we reused the types of test case they cover and wrote new ones where they do not
(Extended Data Table~\ref{tab:coverage}).
The resulting 300 test cases each consist of a starting state, given as an image with text (201 test cases) or as text alone (99), an action describing what happens over the clip, including how the camera moves where relevant, and yes/no questions with expected answers.
Each dimension is the primary focus
of 50 test cases, but most test cases
exercise several dimensions,
so every test case is tagged with
every dimension it tests
(2.4 on average) and scored on each;
each dimension is thus scored
on 100 to 164 test cases.
Models generate a video of 10 to 20 seconds, as each test case specifies.

For each test case, one question checks whether the prompted action occurred, and one question
per tagged dimension
checks whether the capability holds (Extended Data Fig.~\ref{fig:cap-method}); some capability questions are paired with additional questions that rule out trivial passes (e.g., an object remaining on a table only because the camera never left it).
A capability question passes only if it and its additional questions are answered as expected; the action question is otherwise reported separately and does not override the capability questions.
Questions about camera motion are decided from geometry: we label the camera trajectory estimated by VGGT-$\Omega$ with the pans, tilts and translations it contains, and the question passes if these match the motions, order and timing the action requires; when a move must end facing a named target, the judge must also confirm the target.
A camera question also passes if the trajectory matches any alternative reading listed with the test case, and fails if the judge reports a hard scene cut. Only an affirmative checklist answer records a hallucination or cut; a checklist entry affects a
dimension score
only when it is itself
that dimension's question.
The judge answers yes, no or unclear, using unclear when the frames cannot settle a question. An unclear capability answer is excluded from
that dimension's denominator;
an additional question answered wrongly fails it, whereas an unclear additional answer excludes an otherwise passing question. Persistence questions on the hallucination checklist are scored only when the action question is answered yes. A
dimension's score
is the fraction of its remaining questions that pass; the overall score pools all decided capability questions. Across the main-model and reasoner comparisons, 328 of 38,661 capability-question evaluations (0.85\%) are excluded: 275 unclear answers and 53 checklist questions whose action answer was no; none involve camera questions. Per-model exclusion rates range from 0.37\% to 1.83\%.

\subsubsection{Multi-step plans}
\label{sec:tsr}

The 100 multi-step test cases are everyday activities seen through the eyes of a person, whose hands and forearms are the only visible parts of the body. We provide each to the model with a starting image and with text alone, for 200 simulations of 20 seconds each.
Each plan comprises three to six ordered steps (456 steps in total) and draws on two to four of the
six dimensions
(2.6 on average).
Actions say where objects are but not how to move to them, so that the model must work out the movement itself; the hardest steps require turning toward an object out of view.
Because a video model can render the goal without a viable path to it, every step is checked separately, and a simulation that jumps to the goal without the intermediate steps is marked as failed.

For each step, the judge answers a question about whether the step was completed, counting unclear as not completed, and gives the time at which it occurs (Extended Data Fig.~\ref{fig:tsr-method}).
A completed step must have a timestamp and cannot precede an earlier completed step by more than 0.5~s, except for 14 steps marked order-free. Earlier failed steps do not constrain subsequent steps.

Because judges tend to accept a step too readily, each ``yes'' is checked five more times with a prompt asking the judge to look for evidence that the step was not completed, and any ``no'' overturns it.
Camera steps require the estimated trajectory to show the movement; every camera step except pure movement also requires the judge to confirm the step.
Hallucination and invariant checklists are recorded as diagnostics and do not independently override the step verdicts.
We report step completion, the fraction of a plan's steps completed per simulation, and full-plan completion, the fraction of simulations in which every step is completed.

Because a simulation counts towards full-plan completion only if every one of its steps is completed, the measure compounds on step completion.
We fit the compounding curve $F=S^{k}e^{-c(1-S)}$, which gives $F=1$ at $S=1$, to the six models by least squares on $\ln F$ ($R^2=0.986$), and bound $F$ at a given $S$ from the step counts of the plans, since a simulation that does not complete its full plan misses at least one of its steps.
The derivation, validation checks and caveats are given in Supplementary Information~\ref{app:compounding}.

\subsubsection{Judges, settings and further studies}

The judge for both sets of test cases is Qwen3.8-27B, reading videos at 8 frames per second and 480p, with VGGT-$\Omega$ for camera motion.
To validate it, three annotators labelled 160 single-action and 120 multi-step simulations.
We assess agreement between annotators and between the judge and annotators
(Extended Data Table~\ref{tab:judge-validation}).
The judge tracks the annotators' pooled judgements to a moderate degree on single actions and more closely on multi-step plans, although individual verdicts remain noisy.

We run each open-weight model with three seeds and each proprietary model, which offers no seed control, once.
Each model uses its recommended prompt rewriter, except MiniMax-H3, whose rewriter is available only through an API and
for which we use Kimi-K3, the same LLM that PAN~2 uses.
We report means over seeds with 95\% bootstrap confidence intervals over test cases, retaining all seeds and paired starting-state modalities within each resampled case.

We also examine physical simulation on three public benchmarks,
Physics-IQ Verified~\citep{radsch2026verified}, PhyGround~\citep{phyground}
and RBench~\citep{rbench-deng26r} (Extended Data Table~\ref{tab:public-benchmarks}; Supplementary Information~\ref{sec:public-benchmarks}), and in blind pairwise comparisons by human reviewers (Supplementary Information~\ref{sec:human-eval}).
Construction details, inference settings, geometric checks and further validation are given in Supplementary Information~\ref{app:two-tier-evaluation-details}.

\section*{Author Contributions}

J.C., M.D., Z.F., Yi.G., M.G., G.L., H.X., Z.X. and Zi.L. led the development and evaluation of the system. B.C. developed the evaluation framework. B.C., Yin.G., G.H., Y.H., L.H., X.L., K.P., P.P., M.A.Q., P.U., T.Y. and D.Y. contributed to the data pipeline. L.H. contributed to model development. J.L., K.P. and L.H. contributed to evaluation. G.H., J.S. and T.Y. contributed to infrastructure. W.L. contributed to agent simulation experiments. T.L., M.H. and Y.Z. contributed to discussions and testing. X.G. managed the program. Zh.L. supervised the project. E.P.X. conceived, designed and led the project. M.D., M.G., Zh.L. and E.P.X. wrote the paper. All authors reviewed and approved the final manuscript.

\section*{Corresponding Author}
Correspondence and requests for materials should be addressed to Eric P.\ Xing (\href{mailto:eric.xing@mbzuai.ac.ae}{eric.xing@mbzuai.ac.ae}).

\clearpage

\section*{Extended Data}
\setcounter{figure}{0}
\setcounter{table}{0}
\renewcommand{\figurename}{Extended Data Fig.}
\renewcommand{\tablename}{Extended Data Table}
\renewcommand{\thefigure}{\arabic{figure}}
\renewcommand{\thetable}{\arabic{table}}

\begin{figure}[h]
\centering
\includegraphics[width=0.9\textwidth]{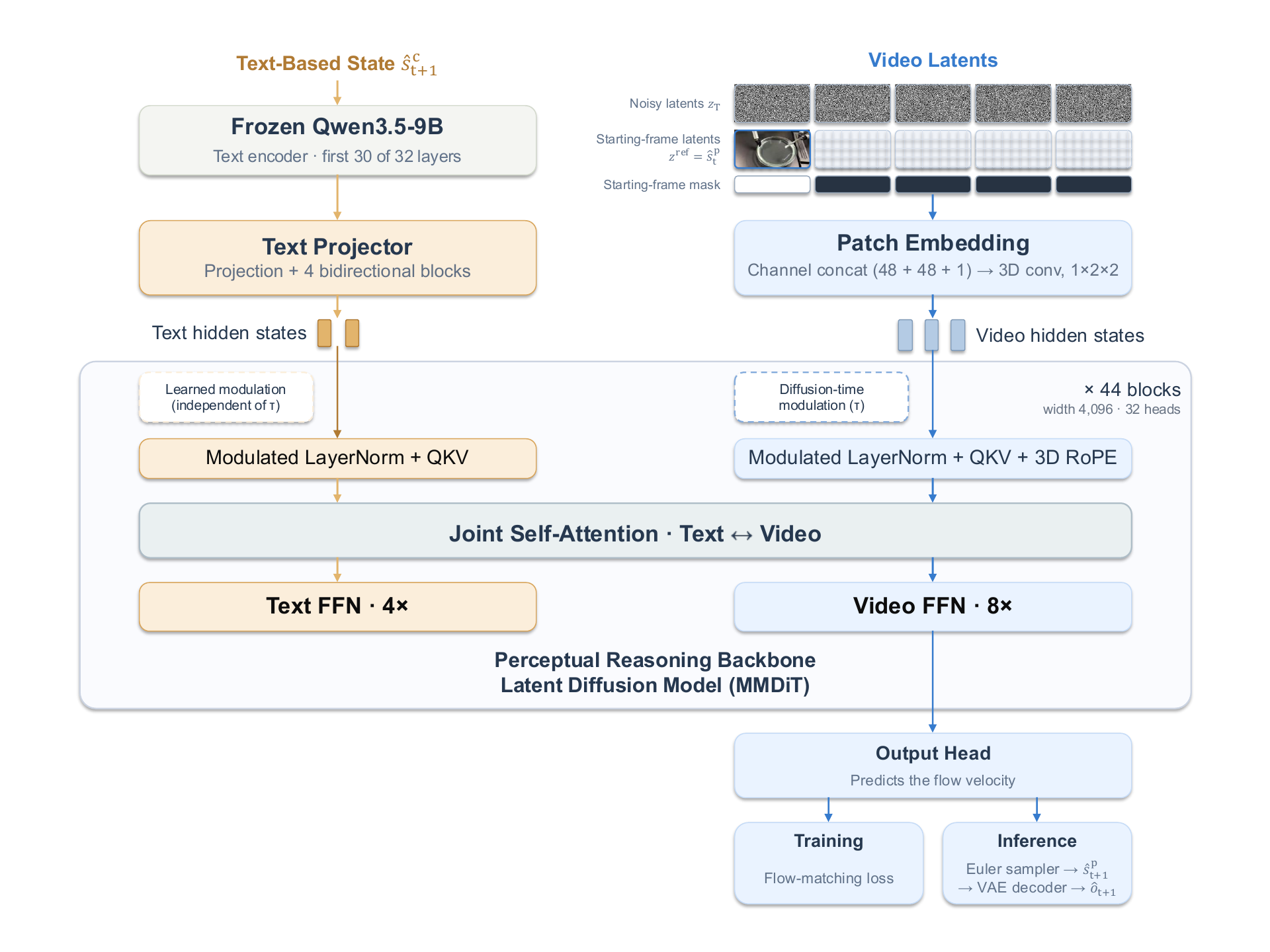}
\caption{\textbf{Perceptual backbone.}
Video and text tokens have modality-specific weights and are processed together with joint self-attention.
The final output head predicts the flow velocity for video tokens only.
At inference, an Euler sampler integrates the velocity and the VAE decoder maps the denoised latents to video.
Full settings are given in Supplementary Table~\ref{tab:ldm-model-card}.
}
\label{fig:ldm-architecture}
\end{figure}

\clearpage

\begin{figure*}[t]
\centering
\begin{minipage}{\textwidth}
\raggedright\textbf{a}\par
\centering
\includegraphics[width=0.95\textwidth]{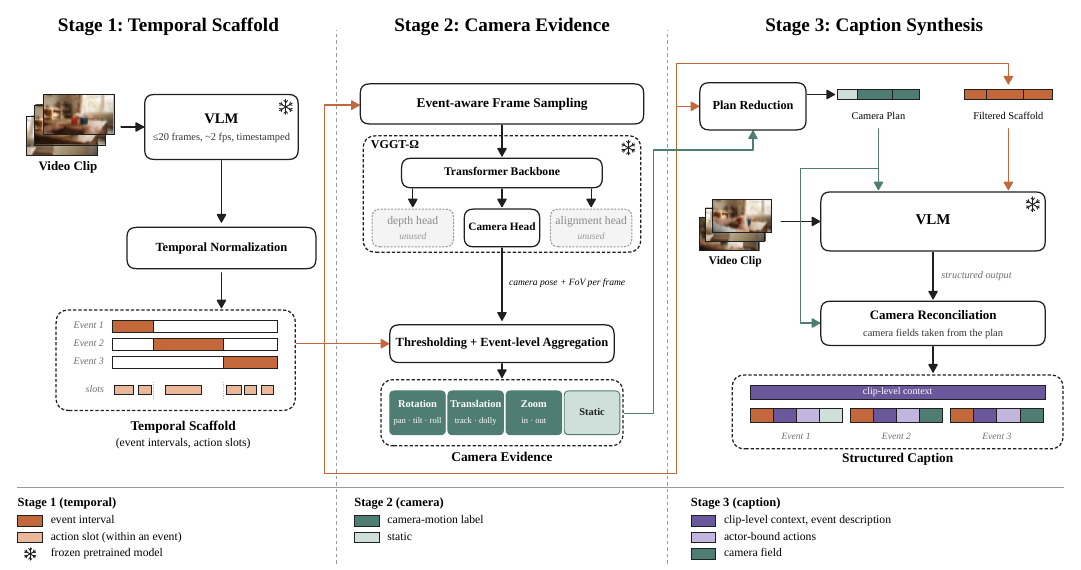}
\end{minipage}
\par\medskip
\begin{minipage}{\textwidth}
\raggedright\textbf{b}\par
\centering
\includegraphics[width=0.85\textwidth]{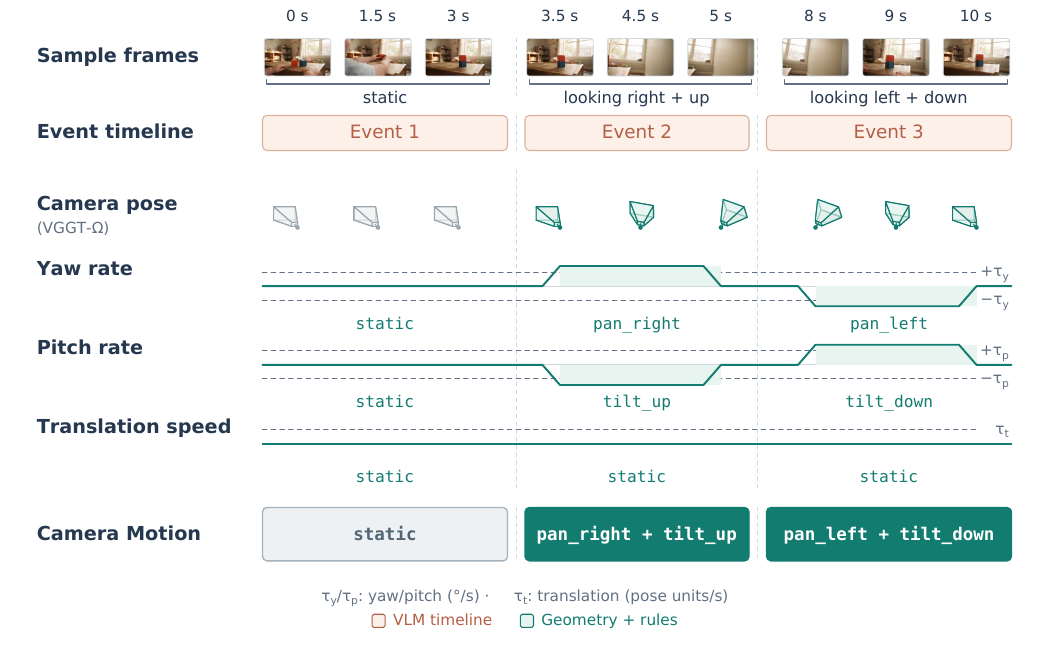}
\end{minipage}
\caption{\textbf{Annotating videos with structured text-based states.}
\textbf{a,} Full pipeline.
An initial VLM pass proposes chronological events and action slots; VGGT-$\Omega$ computes time-stamped camera plans within each event; a second VLM pass writes the final structured caption, with camera plans re-used directly.
\textbf{b,} Camera annotation details.
Camera orientation, position, and focal length are computed for sample frames.
Changes in these parameters between frames are converted to plain text labels (pans, tilts, rolls, tracks, dollies, and zooms) by thresholding their rates.
For more details, refer to Supplementary Information~\ref{app:world-caption-details}.}
\label{fig:discrete_representation}
\label{fig:camera_annotation}
\end{figure*}

\clearpage

\begin{table}[t]
\centering
\small
\setlength{\tabcolsep}{3pt}
\caption{\textbf{Training stages.}
Settings shared by all stages (e.g. weight decay, gradient clipping) are given in Supplementary Information~\ref{app:training-details}.
The released model merges the Stage~III checkpoint (70\%) with the Stage~IV checkpoint after 575 steps (30\%).
I2V, image-to-video; --, not applicable.}
\label{tab:training-stages}
\resizebox{\linewidth}{!}{%
\begin{tabular}{lcccc}
\toprule
Setting & Stage~I & Stage~II & Stage~III & Stage~IV \\
\midrule
Objective & Flow matching & Flow matching & Flow matching & Reward \\
Data & 707.7M clips; 1.07B images & 92.6M clips & 3.1M clips & 896K prompts \\
Resolution & 480p & 720p & 720p & 720p \\
Frame rate & 16 fps & 24 fps & 24 fps & 24 fps \\
Clip duration & 2--20 s & 4--20 s & 7--20 s & 5--9.33 s \\
Image sampling probability per pack item & 0.6 & 0 & 0 & 0 \\
Initialization & Random & Stage~I & Stage~II & Stage~III \\
Diffusion-time sampling & Logit-normal & Logit-normal & Uniform & Uniform \\
Time shift & 5 & 9 & 9 & 9 \\
Token capacity per step & 61.1M & 212M & 212M & -- \\
Batch & -- & -- & -- & 896 videos \\
Peak learning rate & $1\times10^{-4}$ & $2\times10^{-5}$ & $2\times10^{-5}$ & $6\times10^{-6}$ \\
Warm-up & 1,000 steps & 1,000 steps & 500 steps & 10 steps \\
Schedule & Cosine, 5\% floor & Cosine, 5\% floor & Cosine, 5\% floor & Constant \\
Steps & 189k & 30k & 10k & 575 \\
I2V probability / caption drop & 0.3 / 0.05 & 0.3 / 0.05 & 0.3 / 0.05 & 0.3 / -- \\
Parameters updated & All & All & All & Muon partition \\
\bottomrule
\end{tabular}
}
\par\smallskip
\end{table}

\clearpage

\begin{figure*}[t]
\centering
\begin{minipage}{\textwidth}
\raggedright\textbf{a}\par
\centering
\includegraphics[width=\linewidth]{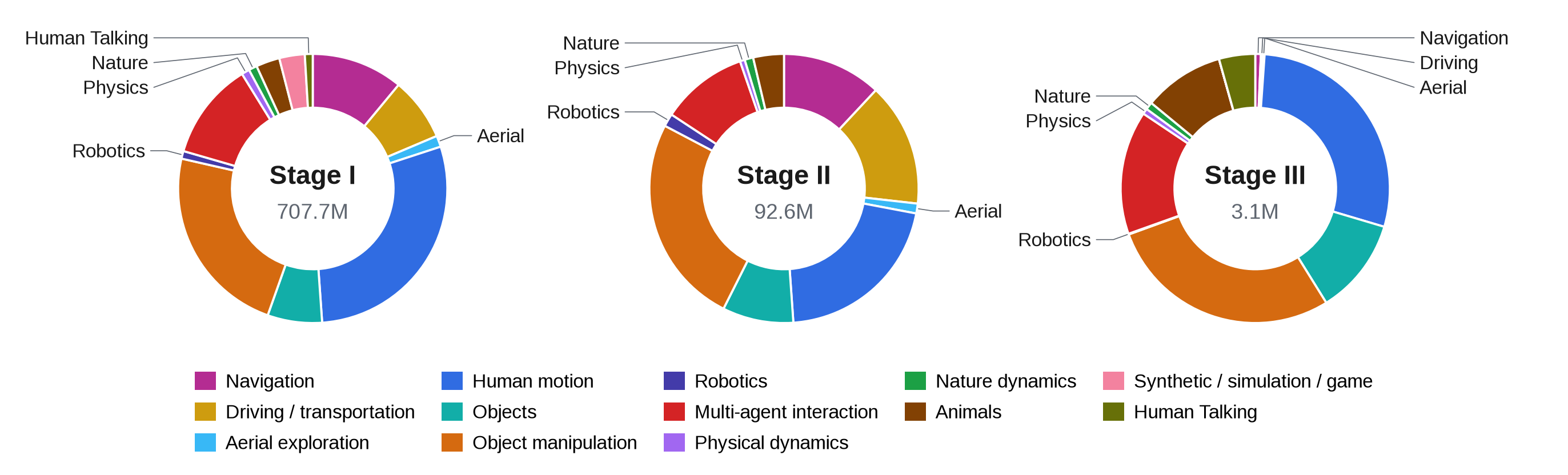}
\end{minipage}
\par\medskip
\begin{minipage}{0.49\textwidth}
\raggedright\textbf{b}\par
\centering
\includegraphics[width=\linewidth]{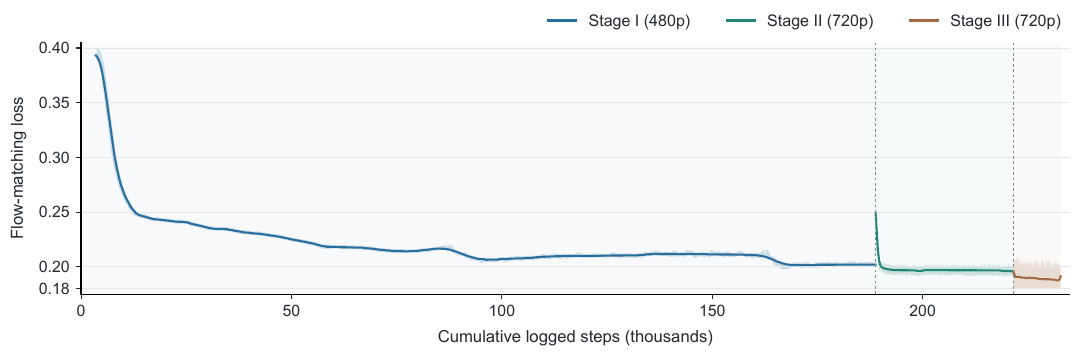}
\end{minipage}\hfill
\begin{minipage}{0.49\textwidth}
\raggedright\textbf{c}\par
\centering
\includegraphics[width=\linewidth]{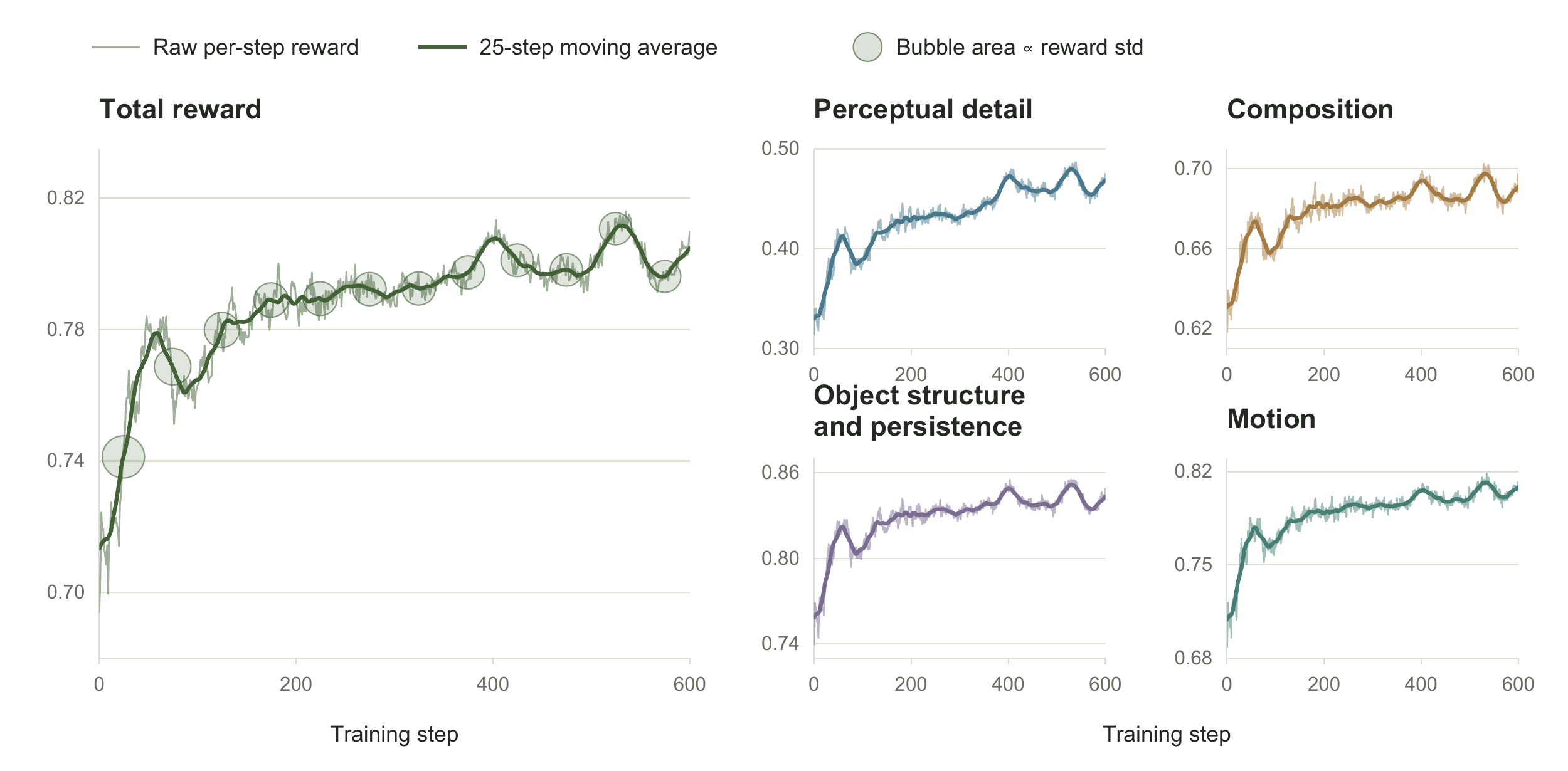}
\end{minipage}
\caption{\textbf{Training stages.}
\textbf{a,} Data distributions in Stages~I--III, by clip count. Stage~IV samples prompts from the Stage~III distribution.
\textbf{b,} Flow-matching loss in Stages~I--III, with dashed lines between each stage, plotted as Gaussian-weighted averages with $\sigma=400$ training steps. Losses are not fully comparable across stages due to primarily to differences in resolution, duration, and data distribution.
\textbf{c,} Total reward during Stage~IV (left) and reward for the four groups of VisionReward questions (right).
}
\label{fig:training-stages-curves}
\label{fig:training-stage-mixtures}
\label{fig:training_loss_curve}
\label{fig:reward_curves}
\end{figure*}

\clearpage

\begin{table}[t]
\centering
\scriptsize
\setlength{\tabcolsep}{3pt}
\caption{\textbf{Dimensions tested by public
video-generation and world-model benchmarks.}
\cmark{}, tested directly; \pmark{}, tested partially, indirectly or for a subset of prompts; \xmark{}, not tested.
Each cell was checked against the benchmark's own paper; the Physics-IQ cells come from our tagging of the 66 scenarios of Physics-IQ Verified, of which 61 test physical consistency, 14 entity persistence, 7 constraint satisfaction, 1 causal coherence, and none
objective dynamism
or
viewpoint invariance,
all being filmed with a fixed camera.
Coverage concentrates on physical consistency; causal coherence and
viewpoint invariance
are rarely tested directly.
Our single-action test cases cover the gaps and are tagged with
every dimension they exercise
(bottom row;
2.4 dimensions per test case
on average).
The coverage review contains 15 benchmarks, six of which also supplied prompts among the 11 sources. Visual quality is not scored in our study.
}
\label{tab:coverage}
\resizebox{\linewidth}{!}{%
\begin{tabular}{lccccccc}
\toprule
Benchmark & \shortstack{Physical\\consistency} & \shortstack{Causal\\coherence} & \shortstack{Entity\\persistence} & \shortstack{Constraint\\satisfaction} & \shortstack{Viewpoint\\invariance} & \shortstack{Objective\\dynamism} & \shortstack{Visual\\quality} \\
\midrule
VBench~\citep{vbench}                     & \pmark & \xmark & \pmark & \xmark & \pmark & \pmark & \cmark \\
EvalCrafter~\citep{evalcrafter}           & \xmark & \xmark & \pmark & \xmark & \xmark & \pmark & \cmark \\
ETVA~\citep{guan2025etva}                 & \cmark & \xmark & \pmark & \xmark & \xmark & \pmark & \cmark \\
VBench-2.0~\citep{vbench2}                & \cmark & \cmark & \cmark & \pmark & \cmark & \cmark & \xmark \\
Physics-IQ~\citep{physicsiq}              & \cmark & \pmark & \pmark & \pmark & \xmark & \xmark & \cmark \\
VideoPhy~\citep{videophy}                 & \cmark & \xmark & \pmark & \pmark & \xmark & \pmark & \xmark \\
PhyGenBench~\citep{phygenbench}           & \cmark & \pmark & \xmark & \pmark & \xmark & \pmark & \xmark \\
PhyGround~\citep{phyground}               & \cmark & \pmark & \pmark & \cmark & \xmark & \pmark & \xmark \\
T2V-CompBench~\citep{t2vcompbench}        & \pmark & \pmark & \pmark & \xmark & \xmark & \cmark & \xmark \\
StoryEval~\citep{temporal}                & \xmark & \cmark & \pmark & \xmark & \xmark & \pmark & \xmark \\
WorldModelBench~\citep{worldmodelbench}   & \cmark & \xmark & \pmark & \cmark & \xmark & \cmark & \cmark \\
WorldScore~\citep{worldscore}             & \pmark & \xmark & \pmark & \xmark & \cmark & \cmark & \cmark \\
CameraCtrl~\citep{camctrl}                & \xmark & \xmark & \pmark & \xmark & \cmark & \pmark & \cmark \\
PAI-Bench-G~\citep{paibench}              & \cmark & \pmark & \pmark & \pmark & \pmark & \cmark & \cmark \\
WorldArena~\citep{worldarena}             & \cmark & \pmark & \pmark & \pmark & \xmark & \cmark & \cmark \\
\midrule
Single-action test cases (this study)     & \cmark & \cmark & \cmark & \cmark & \cmark & \cmark & \xmark \\
\quad Test cases tagged                   & 117 & 100 & 117 & 111 & 121 & 164 & -- \\
\bottomrule
\end{tabular}
}
\end{table}

\clearpage

\begin{figure}[t]
\centering
\includegraphics[width=\linewidth]{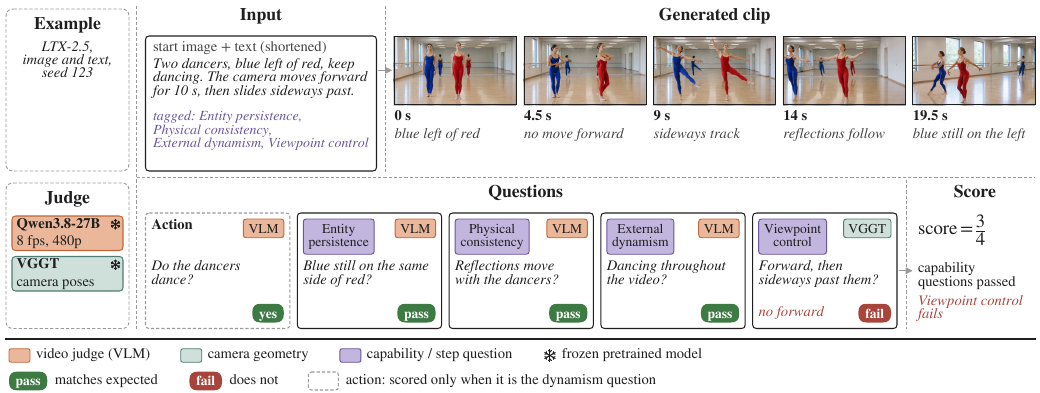}
\caption{\textbf{Building and scoring single-action test cases.}
\textbf{Building.}
We gather 5,198 unique prompts from 11 existing public benchmarks.
We identify which of the 6D-PI they adequately test.
We create new samples to fill in the gaps, primarily related to casual coherence and viewpoint invariance.
We condense the pool to 300 test cases, each tagged with all of the dimensions it tests.
\textbf{Scoring.}
An auto-grader answers questions.
One question always checks whether the prompted actions occur, and each tagged dimension introduces a yes/no question for a grader to answer, which confirms whether the capability holds in the generated video.
A VLM serves as the auto-grader in all cases except for viewpoint invariance, which we score using VGGT-$\Omega$.
In the example shown above, the dancers stay in the correct positions and keep dancing, and their reflections follow them, but the camera never moves forward and only tracks sideways, so the viewpoint-control question fails and the simulation passes 3 of 4 capability questions.
Each of the 6D-PI are scored as the share of its questions that pass the checks.
}
\label{fig:cap-method}
\end{figure}

\clearpage

\begin{figure}[t]
\centering
\includegraphics[width=\linewidth]{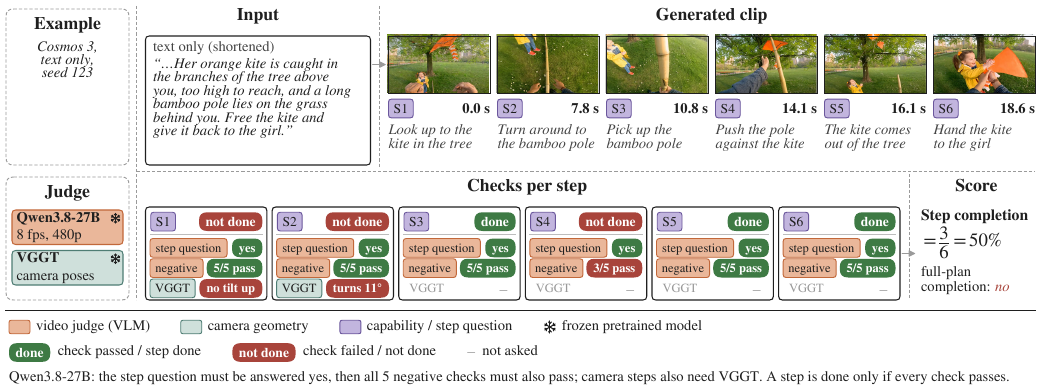}
\caption{\textbf{Scoring a multi-step plan.}
Qwen3.8-27B grades by answering whether each step occurs, and the time at which it occurs.
The Qwen3.8-27B grader is made to iterate on its answer up to 5 times, each time looking for evidence to potentially overturn its decision; a pass requires all 5 to return ``yes.''
For steps involve camera motion, we estimate the trajectory with VGGT-$\Omega$ before grading with the VLM.
In the example simulation above (generated by Cosmos~3, with text input only), steps 1 and 2 pass the judge and all five checks, but the camera trajectory shows no upward tilt for step~1 (the kite is already in view) and only an $11^\circ$ turn for step~2 (the pole is found by looking down rather than by turning around), so both fail.
Step~4 fails since only 3 of the checks return ``yes.''
Since only 3 out of 6 steps are completed successfully, this does not count towards full-plan completion.
}
\label{fig:tsr-method}
\end{figure}

\clearpage

\begin{table}[t]
\centering
\small
\setlength{\tabcolsep}{4pt}
\caption{\textbf{Agreement between annotators, and between the judge and annotators.}
Three annotators labeled simulations by four models from both image-and-text and text-only starting states: 160 single-action simulations (20 test cases), answering the same yes/no questions as the judge, and 120 multi-step simulations (15 plans; 528 steps), marking each step as completed or not.
In each set, 48 simulations were labelled by all three annotators and the rest by one annotator each.
Agreement between annotators is computed on the questions or steps of the 48 shared simulations: raw agreement is pooled over annotator pairs; Light's $\kappa$~\citep{light1971measures} averages Cohen's $\kappa$~\citep{cohen1960coefficient} over the three pairs; Fleiss' $\kappa$~\citep{fleiss1971measuring} extends it to more than two raters; and Krippendorff's $\alpha$~\citep{krippendorff2004content} (nominal) compares observed with chance-expected disagreement. For all three coefficients, 1 is perfect agreement and 0 is chance level.
Agreement between the judge and annotators is Spearman's $\rho$ between the judge's score for each simulation (the share of questions passed, or step completion) and the pooled annotator score, taken as the mean of the simulation's annotators or as a per-question majority vote, over all labelled simulations; brackets give 95\% bootstrap confidence intervals over simulations.
The two kinds of agreement use different measures and are not directly comparable. Inter-annotator intervals are 95\% percentile intervals from 10,000 simulation-level bootstrap draws, retaining all questions or steps and all three annotators within each sampled simulation; they condition on these annotators and seed 123.
}
\label{tab:judge-validation}
\label{tab:cap-agreement}
\label{tab:pan-tsr-agreement}
\begin{tabular}{lcc}
\toprule
 & Single-action test cases & Multi-step plans \\
\midrule
Simulations labelled (shared by all three) & 160 (48) & 120 (48) \\
\midrule
\multicolumn{3}{l}{\emph{Between annotators}} \\
Questions or steps compared & 157 & 212 \\
Raw agreement (\%) & 80.47 [74.95, 85.71] & 79.25 [74.53, 83.41] \\
Light's $\kappa$ & 0.58 [0.47, 0.68] & 0.55 [0.45, 0.64] \\
Fleiss' $\kappa$ & 0.58 [0.47, 0.68] & 0.55 [0.45, 0.63] \\
Krippendorff's $\alpha$ & 0.58 [0.47, 0.68] & 0.55 [0.45, 0.63] \\
\midrule
\multicolumn{3}{l}{\emph{Between the judge and annotators (Spearman's $\rho$)}} \\
Pooled annotator mean & 0.46 [0.31, 0.58] & 0.62 [0.48, 0.72] \\
Pooled annotator majority & 0.48 [0.34, 0.61] & 0.61 [0.47, 0.72] \\
\bottomrule
\end{tabular}
\end{table}

\clearpage

\begin{table}[t]
\centering
\small
\setlength{\tabcolsep}{4pt}
\caption{\textbf{Public benchmarks of physical simulation.}
Higher is better for all scores, which follow each benchmark's convention except that RBench is multiplied by 100.
Physics-IQ Verified~\citep{radsch2026verified} (image and text) is normalized so that the physical variation between two real recordings of the same scenario scores 100; PhyGround~\citep{phyground} is scored from 1 to 5 by Qwen3.8-27B; RBench~\citep{rbench-deng26r} is the unweighted mean over five task categories and four embodiment categories.
Physics-IQ Verified is reported in two settings.
In the public-leaderboard setting, scores for Cosmos~3~Super, MiniMax-H3, Seedance~2.5 and Omni Flash~1.1 are their leaderboard entries, and FLUX~3's is its entry with Claude Opus 5.5 prompts, without best-of-$N$ selection; PAN~2 uses prompts rewritten by Claude Opus 5.5 with an instruction adapted from FLUX~3's, together with scene-specific negative prompts (seeds 123--126); --, no leaderboard entry.
In the system-default setting, we run each model's own default pipeline (Cosmos~3~Super, its Reasoner; MiniMax-H3, its official prompt skill with Kimi-K3; LTX-2.5, its Gemma-4 E2B prompt enhancer; PAN~2, Kimi-K3 as conceptual reasoning backbone), with seeds 123--125.
PhyGround and RBench are run by us with the same rewriters and seeds.
We report means, with case-based 95\% bootstrap confidence intervals in brackets (seeds averaged within each test case; 10,000 draws for Physics-IQ Verified and PhyGround, 20,000 for RBench).
$^{\S}$We use the scores from public leaderboards, so no interval is given.
Bold marks only the highest mean in each column, not considering potential overlap in the intervals.
$^\dagger$Except for the public leaderboard scores, videos from proprietary models are generated for one seed only through the fal API and scored by us; some prompts or videos are blocked by fal, so these are excluded (PhyGround: Seedance~2.5 241 of 250, Omni Flash~1.1 248 of 250; RBench: FLUX~3 644 of 650, Seedance~2.5 638 of 650, Omni Flash~1.1 649 of 650).
More settings are described in the Supplementary Information~\ref{sec:public-benchmarks}.
}
\label{tab:public-benchmarks}
\resizebox{\linewidth}{!}{
\begin{tabular}{lcccc}
\toprule
 & \multicolumn{2}{c}{Physics-IQ Verified} & & \\
\cmidrule(lr){2-3}
Model & \shortstack{Public-leaderboard\\setting} & \shortstack{As run in\\this study} & PhyGround & RBench \\
\midrule
Cosmos~3~Super           & 48.62$^{\S}$ & 36.11\,{\scriptsize[31.46, 40.84]} & 4.63\,{\scriptsize[4.58, 4.68]} & 50.64\,{\scriptsize[48.31, 52.93]} \\
MiniMax-H3               & 39.80$^{\S}$ & \textbf{42.91}\,{\scriptsize[38.26, 47.80]} & 4.73\,{\scriptsize[4.68, 4.77]} & 56.16\,{\scriptsize[53.90, 58.43]} \\
LTX-2.5                  & -- & 26.85\,{\scriptsize[23.04, 30.64]} & 4.57\,{\scriptsize[4.51, 4.63]} & 51.55\,{\scriptsize[49.19, 53.88]} \\
PAN~2                    & 49.38\,{\scriptsize[44.58, 54.24]} & 37.48\,{\scriptsize[32.97, 42.08]} & 4.73\,{\scriptsize[4.69, 4.78]} & 54.54\,{\scriptsize[52.30, 56.74]} \\
\midrule
FLUX~3$^\dagger$         & \textbf{51.11}$^{\S}$ & 38.48\,{\scriptsize[33.70, 43.43]} & 4.73\,{\scriptsize[4.69, 4.78]} & 57.96\,{\scriptsize[55.40, 60.45]} \\
Seedance~2.5$^\dagger$   & 42.43$^{\S}$ & 40.08\,{\scriptsize[35.24, 44.96]} & 4.73\,{\scriptsize[4.69, 4.78]} & 56.38\,{\scriptsize[53.73, 59.00]} \\
Omni Flash~1.1$^\dagger$ & 35.34$^{\S}$ & 33.86\,{\scriptsize[29.39, 38.59]} & \textbf{4.78}\,{\scriptsize[4.73, 4.83]} & \textbf{60.79}\,{\scriptsize[58.35, 63.19]} \\
\bottomrule
\end{tabular}
}
\end{table}

\clearpage
\section*{Supplementary Information}
\setcounter{section}{0}
\setcounter{figure}{0}
\setcounter{table}{0}
\setcounter{equation}{0}
\renewcommand{\figurename}{Supplementary Fig.}
\renewcommand{\tablename}{Supplementary Table}
\renewcommand{\thesection}{\arabic{section}}
\renewcommand{\thefigure}{\arabic{figure}}
\renewcommand{\thetable}{\arabic{table}}
\renewcommand{\theequation}{S\arabic{equation}}

\section{Notation}
\label{app:notation}
\label{app:pan-comparison}



Supplementary Table~\ref{tab:notation} defines symbols used throughout the paper.
We use three kinds of time, and keep them distinct as follows:
the simulation step index is $t$;
physical time elapsed within a simulation step, in seconds, is defined by event intervals $\iota_k$ and frame times $u_n$;
diffusion time steps are given as $\tau\in(0,1)$.
Transformer depth is denoted by block index $\ell$.

\begin{table}[h]
\centering
\small
\setlength{\tabcolsep}{4pt}
\caption{\textbf{Notation.}}
\label{tab:notation}
\begin{tabular}{@{}p{0.30\linewidth}p{0.64\linewidth}@{}}
\toprule
Symbol & Meaning \\
\midrule
\multicolumn{2}{@{}l}{\emph{World model}} \\
$t$ & Simulation step \\
$o_t$ & Observation: an optional image with text \\
$a_t$ & Action: an instruction in text and the duration $d_t$ to simulate \\
$\hat s_t=(\hat s^{\mathrm c}_t,\hat s^{\mathrm p}_t)$ & Estimated world state: conceptual and perceptual parts \\
$h_{\mathrm{sem}}$, $h_{\mathrm{VAE}}$ & Semantic encoder (vision encoder and text embedding of the multimodal LLM); VAE encoder \\
$f_{\mathrm{LLM}}$, $f_{\mathrm{LDM}}$ & Conceptual reasoning backbone; perceptual backbone \\
$g$, $\hat o_{t+1}$ & VAE decoder; predicted observation (video) \\
\midrule
\multicolumn{2}{@{}l}{\emph{Text-based state}} \\
$c$ & Global context: scene, subjects and their attributes, point of view \\
$e_k=(\iota_k,\delta_k,\alpha_k,\kappa_k)$ & Event $k=1,\dots,E$: time interval, description of state changes, timed actor-bound actions, camera motion \\
\midrule
\multicolumn{2}{@{}l}{\emph{Perceptual backbone}} \\
$z$, $z^{\mathrm{ref}}$ & Clean latents of the target clip; latents of the starting frame ($=\hat s^{\mathrm p}_t$) \\
$\tau$, $\epsilon$, $z_\tau$ & Diffusion time; Gaussian noise; noised latents $(1-\tau)z+\tau\epsilon$ \\
$u_\theta$ & Velocity predicted by the LDM with parameters $\theta$ \\
$x^{\mathrm v}_\ell$, $x^{\mathrm c}_\ell$, $B_{\theta,\ell}$, $L$ & Video and text tokens entering block $\ell$; block $\ell$; number of blocks (44) \\
$d$, $H$, $d_h$ & Transformer width (4,096); attention heads (32); head dimension (128) \\
$\chi$, $\phi_\chi$ & Time shift; shifted time grid \\
$w$ & Classifier-free guidance scale \\
\midrule
\multicolumn{2}{@{}l}{\emph{reward-based learning and optimization}} \\
$R(V)$, $\mathcal Q$, $w_q$, $\bar p_\phi$ & Reward of video $V$; question set; question weight; probability of ``yes'' \\
$\mathcal N_K$, $\mathcal N^{\mathrm{head}}_K$ & $K$-step Newton--Schulz orthogonalization; its per-head version \\
$B_t$, $S_t$, $\mu$ & Momentum buffer; Nesterov-style update; momentum coefficient \\
$\eta_t$, $\lambda_{\mathrm{wd}}$ & Learning rate; decoupled weight decay \\
\midrule
\multicolumn{2}{@{}l}{\emph{Section-local}} \\
$T_V$, $u_n$, $\Lambda_n$, $\mathcal Z$, $\mathcal K_k$ & Annotation (Supplementary Information~\ref{app:world-caption-details}): clip duration; frame time; active camera labels; temporal scaffold; camera evidence \\
\bottomrule
\end{tabular}
\end{table}

\section{Conceptual reasoning backbone}
\label{app:world-reasoner-settings}
\label{app:kimi-reasoner}

We give encoding settings for the conceptual reasoning backbone in Supplementary Table~\ref{tab:kimi-reasoner-settings}.
Its final output is the predicted conceptual state $\hat s^{\mathrm c}_{t+1}$, a single JSON object, \texttt{\{"refined\_prompt": "<serialized caption JSON>"\}}.
The conceptual state has seven fields: \texttt{video\_style}, \texttt{lighting}, \texttt{perspective\_view}, \texttt{subjects\_lifecycle}, \texttt{init\_frame\_visual\_description}, \texttt{events} and \texttt{quality}.
Following the schema in which the training videos are annotated (Supplementary Table~\ref{tab:tgc-caption-fields}), the \texttt{events} field contains events $e_k$, and the remaining fields comprise the global context $c$.

The LDM accepts up to 1,536 tokens, measured with the Qwen3.5-9B tokenizer, for the state and for the negative prompt separately.
Over-length states are rejected.
A response that is rejected or fails validation, including for invalid JSON or violations of our conceptual state format, is resampled for up to four total attempts.
Only validated states are used.
The negative prompt that the LDM uses for classifier-free guidance is universal and fixed, and requires no modifications from the conceptual reasoning backbone.




\begin{table}[h]
\centering
\small
\caption{\textbf{Settings of the conceptual reasoning backbone.} Exceptions in individual studies are reported with those studies.}
\label{tab:kimi-reasoner-settings}
\begin{tabular}{@{}p{0.28\linewidth}p{0.67\linewidth}@{}}
\toprule
Setting & Value \\
\midrule
Model & \texttt{moonshotai/Kimi-K3} \\
Instruction & Fixed world-reasoning instruction, supplied verbatim \\
Reasoning & \texttt{thinking: true}; \texttt{thinking\_effort: "max"} \\
Temperature; top-$p$ & 1.0; 0.95 \\
Completion budget & 32,768 tokens, including reasoning tokens \\
Inputs & \texttt{original\_prompt}; \texttt{target\_duration\_sec}; starting image, if any \\
Image handling & No resizing or cropping by the caller \\
Attempts & Up to four per call; no fallback model \\
Output limit & 1,536 tokens of the LDM's text encoder \\
\bottomrule
\end{tabular}
\end{table}


\section{Perceptual backbone}
\label{app:ldm-supp}

We provide the configuration of the perceptual backbone in Extended Data Fig.~\ref{fig:ldm-architecture} and Supplementary Table~\ref{tab:ldm-model-card}.
Our LDM operates in the latent space of the frozen Wan2.2 VAE~\citep{wan2025}.
For training, latents are normalized per channel by the VAE's latent mean and standard deviation.
When the starting state includes an image, its VAE latents occupy the first latent frame of the reference tensor, with zeros at later frames.
In this image-to-video (I2V) setting, the reference frame latents are marked by a mask.
In text-to-video (T2V), the mask is zero everywhere.
Video tokens carry three-axis rotary position embeddings~\citep{su2024roformer} over latent time, height and width; temporal positions are scaled by $16/f$ at frame rate $f$, so that the same interval of physical time gives the same rotary difference at any frame rate.
Text tokens carry no rotary position embedding.

\begin{table}[h]
\centering
\small
\setlength{\tabcolsep}{3pt}
\caption{\textbf{Perceptual backbone configuration.}
}
\label{tab:ldm-model-card}
\begin{tabular}{@{}p{0.25\linewidth}p{0.70\linewidth}@{}}
\toprule
Component & Configuration \\
\midrule
Trainable parameters & 29.20B, of which 24.71B updated by Muon and 4.49B by AdamW \\
Transformer & 44 joint-attention blocks with modality-specific weights; width 4,096; 32 heads of dimension 128 \\
Attention & Multi-head; joint over video and text tokens; Q/K/V biases; per-head Q/K RMS normalization; FlashAttention-4 with packing mask \\
Normalization & Layer normalization without affine parameters, followed by modality-specific adaptive scale, shift and gate \\
Modulation & Video: six vectors per block from the diffusion-time embedding. Text: six learned vectors per block, independent of diffusion time \\
Feed-forward & GELU (tanh); video 4,096$\to$32,768$\to$4,096; text 4,096$\to$16,384$\to$4,096 \\
Video positions & 3D RoPE over (time, height, width); dimensions [16, 56, 56]; base 256 \\
Input & 97 channels (48 noisy, 48 reference, 1 mask); 3D convolution with $1\times2\times2$ patch \\
Output head & Adaptive layer normalization from the diffusion time; linear 4,096$\to$192; rearranged to 48 channels \\
Text encoder & Frozen \texttt{Qwen/Qwen3.5-9B}, first 30 of 32 layers, final normalization removed (7.51B parameters) \\
Text projector & RMSNorm; linear 4,096$\to$4,096; four bidirectional blocks with 32 heads, SwiGLU (expansion 4) and Q/K RMSNorm ($\approx$1.09B parameters) \\
Text length & Up to 1,536 tokens per sample \\
VAE & \texttt{Wan-AI/Wan2.2-TI2V-5B-Diffusers}, frozen; 48 channels; compression 4 in time, 16 in space \\
Video shape & $704\times1248$ at 24 fps; latent grid $44\times78$ (858 tokens per latent frame); up to 481 frames (121 latent frames, 20 s) \\
\bottomrule
\end{tabular}
\end{table}

To predict clean latents for the VAE to decode, we integrate the flow from seeded Gaussian noise with a first-order Euler solver over a shifted time grid,
\begin{equation}
\phi_\chi(\tau)=\frac{\chi\,\tau}{1+(\chi-1)\,\tau},
\label{eq:time-shift}
\end{equation}
which concentrates steps at high noise levels for $\chi>1$.
The model predicts a velocity, $u_{\mathrm{pos}}$, for the conceptual state $\hat s^{\mathrm c}_{t+1}$, as well as a velocity $u_{\mathrm{neg}}$ for the negative prompt, separately.
Using classifier-free guidance with guidance scale $w$, the predicted velocity is
\begin{equation}
u_{\mathrm{CFG}}=u_{\mathrm{neg}}+w\,\bigl(u_{\mathrm{pos}}-u_{\mathrm{neg}}\bigr),
\label{eq:cfg}
\end{equation}

Guidance starts at the third iteration; sampling for the first two iterations uses $u_{\mathrm{pos}}$ alone.
The first iteration of text-to-video sampling applies zero velocity.

For image-to-video simulation, the first latent frame is updated with $u_{\mathrm{pos}}$ alone, that is, with $w=1$, and later latent frames with the guidance scale $w$.
The first latent frame is sampled like the others and is not overwritten with $z^{\mathrm{ref}}$.

\begin{table}[h]
\centering
\small
\caption{\textbf{Perceptual backbone sampling settings.} Iterations are zero-indexed.}
\label{tab:ldm-sampling}
\begin{tabular}{lcc}
\toprule
Setting & Text only & Image and text \\
\midrule
Euler steps & 75 & 50 \\
Time shift $\chi$ & 7 & 9 \\
Guidance scale $w$ & 3 & 3 \\
Iterations with zero velocity & 1 & 0 \\
First guided iteration & 2 & 2 \\
\bottomrule
\end{tabular}
\end{table}

\section{Text-based state annotation}
\label{app:world-caption-details}
\label{app:additional-captioner-details}
\label{sec:tgc-stage3}



We use a three-stage pipeline to annotate each training clip $V$ of duration $T_V$ with a dense structured JSON text caption (Extended Data Fig.~\ref{fig:discrete_representation}, Supplementary Table~\ref{tab:tgc-caption-fields}).
The first stage proposes events with a VLM, the second uses a 3D vision model to determine camera motion within each event, and the third finalizes the caption by adding time-stamped actions and fine details.

For the first stage, event proposal, the VLM processes the clip $V$ at two frames per second, capped at 20 frames for any given clip.
We explicitly provide the clip duration and time stamps for each frame in plain text.
The VLM provides time-stamped endpoints, in seconds, that bound potential events and actions.
Changes of viewpoint (camera movement) may coexist within the same event, and action slots can overlap when multiple subjects act simultaneously.
We post-process the VLM output such that predicted times are parsed into seconds, event intervals are clipped to not exceed $T_V$, action slots are clipped to their bounding event, and reversed endpoints are swapped.
We refer to the resulting collection of event intervals $\iota_k$ and their action slots as a temporal scaffold $\mathcal Z$.
For VLM inference, we use temperature 0 with a limit of 1,024 tokens per video.

In the second stage, we estimate camera motion geometrically.
Frames are sampled separately for this purpose: for each event, a two-frames-per-second grid starts at the beginning of $\iota_k$ and both endpoints of the event are included, as are the endpoints of the clip; the union of these times is mapped to native frames and deduplicated.
The camera head of VGGT-$\Omega$~\citep{wang2025vggt,wang2026vggtomega} (with its balanced-512 preprocessing) estimates the pose and focal length of each sampled frame.
Let $n$ index the sampled frames in temporal order and $u_n$ be the time of frame $n$.

For each pair of adjacent frames, we compute the change from camera for $n$ to the camera for $n+1$ in the coordinate frame corresponding to frame $n$ and divide it by $u_{n+1}-u_n$.
From the change in viewing direction, the velocity of the camera center, and rate of change of the focal length, we derive yaw, pitch, roll, translation, and zoom.
Whenever any of these four reach or exceed their hand-tuned thresholds (Supplementary Table~\ref{tab:tgc-camera-mapping}), it yields a motion label.
$\Lambda_n$ is the set of all labels active within the time interval $[u_n,u_{n+1}]$.

We measure the support for label $m$ in event $k$ as the time for which the label is active within the event,
\begin{equation}
\operatorname{supp}_k(m)=\sum_n\bigl|[u_n,u_{n+1}]\cap\iota_k\bigr|\;\mathbf 1\bigl[m\in\Lambda_n\bigr],
\label{eq:tgc-support}
\end{equation}
and we keep labels that have sufficient absolute and relative support,
\begin{equation}
\mathcal K_k=\Bigl\{m:\ \operatorname{supp}_k(m)\ge\min\bigl(0.6\,\mathrm s,\,|\iota_k|\bigr)\ \text{and}\ \operatorname{supp}_k(m)\ge0.16\,|\iota_k|\Bigr\}.
\label{eq:tgc-evidence}
\end{equation}

If no labels have sufficient support, then the time interval contains only static camera or minor motion.
We also condense the camera plans to avoid excessive short or competing motions.
Instances covering at least 40\% of events of 3 seconds or less, or covering 30\% of longer events, are retained.
Shorter instances are retained if they have a sufficiently high rate-to-threshold ratio.
Furthermore, the plan may only keep one camera group, a sub-interval of $\iota_k$, per event for events of 3 seconds or less, and two camera groups otherwise.
Each camera group consists of up to two dominant compatible labels: if labels conflict, the one that correlates with the event summary is kept if the event summary gives any indication, or the more dominant movement otherwise.
The more dominant retained label is the primary label, and a secondary label is kept only if it is at least 50\% as strong as the primary label.
An event is flagged for unstable camera motion if we have evidence that the camera moves, but no label survives our selection process, or when only brief, high-magnitude instances survive and either at least 20\% of the event supports the opposite direction or at least 45\% supports directions absent from the event summary.

After designing the pipeline and hand-tuning thresholds, we ensure that our pipeline outperforms VLM-based alternatives (Supplementary Table~\ref{tab:camera_ablation}).

At the start of the third stage, Qwen3.6-27B captions only the first frame in a standalone request, without access to later frames.
A subsequent request uses this first-frame caption, the camera plan, and the temporal scaffold, along with the video, to write the final caption.
It processes the clip at 2 frames per second, with a minimum of 8 frames and a maximum of 20 frames, with temperature 0.2, top-$p$ 0.9 and a limit of 4,096 tokens.
The wording of the first stage is mostly hidden, so that this pass describes the clip instead of copying the first stage.
The output follows our fixed JSON schema and includes a provisional camera field $\kappa^{\mathrm{vlm}}_k$ which is used in cases where the geometric plan $\kappa_k$ does not exist.
We instruct the VLM to exclude camera operation from event descriptions, so that viewpoint motion is disentangled from other scene and subject motion.
Actions are very brief descriptions of subjects and what they do, with immediate visible consequences described in $\delta_k$.
Event intervals and action times serve as fine-grained, dense semantic temporal conditioning for the LDM.


We caption images with the video schema, but without temporal fields.
Each image is thus described by only global context $c$, mainly with \texttt{init\_frame\_visual\_description}.

\begin{table}[h]
\centering
\caption{\textbf{Camera measures, thresholds and motion labels.} Thresholds are inclusive and apply to absolute values.}
\label{tab:tgc-camera-mapping}
\begingroup
\small
\setlength{\tabcolsep}{4pt}
\renewcommand{\arraystretch}{1.15}
\begin{tabular}{@{}lll@{}}
\toprule
Measure & Threshold & Label (positive / negative) \\
\midrule
Yaw rate & \(2.2^\circ\,\mathrm{s}^{-1}\) & \texttt{pan\_right} / \texttt{pan\_left} \\
Pitch rate & \(2.2^\circ\,\mathrm{s}^{-1}\) & \texttt{tilt\_down} / \texttt{tilt\_up} \\
Roll rate & \(2.6^\circ\,\mathrm{s}^{-1}\) & \texttt{roll\_clockwise} / \texttt{roll\_counterclockwise} \\
Lateral translation & See note & \texttt{track\_right} / \texttt{track\_left} \\
Vertical translation & See note & \texttt{dolly\_down} / \texttt{dolly\_up} \\
Forward translation & See note & \texttt{dolly\_forward} / \texttt{dolly\_backward} \\
Fractional focal-length change & \(0.025\,\mathrm{s}^{-1}\) & \texttt{zoom\_in} / \texttt{zoom\_out} \\
\bottomrule
\end{tabular}
\par\smallskip
\parbox{0.95\linewidth}{\footnotesize The threshold for translation is set in terms of pose units: 0.02 pose units per second. For diagonal motion, a label is retained if the speed along its axis is at least 45\% of the largest absolute component.
}
\endgroup
\end{table}

\begin{table}[h]
\centering
\caption{\textbf{Fields of the text-based state.} The same schema is used to annotate training videos and is written by the conceptual reasoning backbone at inference.}
\label{tab:tgc-caption-fields}
\begingroup
\small
\setlength{\tabcolsep}{4pt}
\renewcommand{\arraystretch}{1.15}
\begin{tabular}{@{}lp{0.35\linewidth}p{0.45\linewidth}@{}}
\toprule
Component & Output field & Content \\
\midrule
$c$ & \texttt{subjects\_lifecycle} & Clip-local subject identifiers, category and stable appearance \\
$c$ & \texttt{init\_frame\_visual\_description} & Layout and state in the first visible frame \\
$c$ & \texttt{video\_style}, \texttt{lighting}, \texttt{perspective\_view}, \texttt{quality} & Presentation and quality of the clip \\
$\iota_k$ & \texttt{event\_index}, \texttt{timestamp} & Identifier and time interval of the event \\
$\delta_k$ & \texttt{event\_description} & Visible change of state within the event \\
$\alpha_k$ & \texttt{contained\_actions} & Timed actions, each bound to an actor \\
$\kappa_k$ & \texttt{camera\_motion} & Timed groups of camera labels, from the camera plan \\
\bottomrule
\end{tabular}
\endgroup
\end{table}

\section{Training}
\label{app:training-details}
\label{app:training-source-schedules}


We first describe the three flow-matching training stages, then Stage~IV's reward learning.
Extended Data Table~\ref{tab:training-stages} summarizes settings for each stage.
Gradients are clipped to a norm of 1.0, the decoupled weight decay is $\lambda_{\mathrm{wd}}=0.05$ for both optimizers, and the random seed is 42.

\subsection{Stages of flow-matching training}

Stage~I starts at 480p, 16 frames per second for large-scale pretraining.
Stage~II continues with resolution increased to 720p and framerate increased to 24 frames per second.
The final two stages (including reward learning) keep these same settings.

In all three flow-matching stages, we use the image-to-video setting with probability $0.3$.
We replace the caption encoding with an encoding of the empty caption with probability $0.05$, which enables classifier-free guidance for better control at inference.
Starting in Stage~II, we drop all time stamps from the caption of a sampled video with probability $0.4$.
This helps increase robustness to problems with time stamps provided by the conceptual reasoner at inference, while still retaining the core time-grounding capabilities.



\subsection{Flow-matching loss and diffusion time}

We compute the loss given in Eq.~\ref{eq:flow-matching} on video latents only, which excludes text tokens and padding.
For mean squared error, we sum squared error and the number of valid latent elements across all ranks before dividing, so that longer clips contribute proportionally more to the loss.
Diffusion time steps are drawn from a logit-normal distribution, $\tau=\operatorname{sigmoid}(\xi)$ with $\xi\sim\mathcal N(0,1)$, for the first two stages, and a uniform distribution, $\tau\sim\mathcal U(0,1)$, for the final two stages.
In all stages, the diffusion time steps are shifted by $\phi_\chi$ as shown in Eq.~\ref{eq:time-shift}.


\subsection{Reward-based learning}

In Stage~IV, we switch from flow-matching loss to reward-based learning.
Instead of supervising with full real video, we sample only captions and, for image-to-video, real starting frames, from the Stage~III videos.
We then supervise by generating video rollouts from these prompts, and scoring the videos with our reward model using the thirteen reward questions in Supplementary Table~\ref{tab:reward-questions}.


We use classifier-free guidance with guidance scale $w=1$ on a 50-step Euler schedule with time shift $9$ for all rollouts.
We choose a denoising step from steps 40--49 for the gradient-enabled prediction of clean latents.
For reward questions on structure, motion, and composition, we use the standard short-side resize and center crop.
For questions concerning perceptual detail, we instead take three $224{\times}224$ crops at the full 720p resolution, to avoid distorting the fine-grained details or hiding artifacts via resizing.
These crops are taken randomly for each question and each rollout, with one assigned to each third of the frame, and kept fixed across frames so as to preserve local spatio-temporal correspondences.
The probabilities of ``yes'' for all three crops are averaged equally.

We sample prompts from the Stage~III data without replacement.
Thus, our 575-step checkpoint is trained on 515,200 unique prompts from the pool of 896,000 candidate prompts.
We score decoded frames at one-second intervals, starting at a random frame within the first second.
For image-to-video rollouts, we never start with the first frame.


Two regularizers stabilize our reward learning.
For image-to-video rollouts, an anchor on the first-frame latents preserves the starting image.
For all rollouts, a penalty on periodic artifacts aligned with the token grid discourages grid patterns.
The grid penalty uses the reward-scored frames plus the first two decoded frames for text-to-video rollouts, and the reward-scored frames plus the second decoded frame for image-to-video rollouts.

The per-rollout minimization objective is
\begin{equation}
\mathcal L_{\mathrm{IV}}=-\lambda_r\,R(D(\hat z_0))+\mathbf 1[\mathrm{I2V}]\,\operatorname{MSE}(\hat z_0[f=0],z^{\mathrm{ref}})+\lambda_g(j)\,G(D(\hat z_0)),
\label{eq:reward-objective}
\end{equation}

where $\lambda_r \approx 0.025$, $D$ is the frozen Wan VAE decoder, and the first-frame anchor averages squared error over the first-frame latents.
$G$ is the grid penalty, which averages $\max(0,\operatorname{Var}(P_f)-2.198\times10^{-5})$ for decoded frames $f$.
To construct $P_f$, we subtract a $9\times9$ box blur from the frame's luma, average the residual over all complete $32\times32$-pixel patches aligned with diffusion transformer tokens, and mean-center the resulting pattern.
We use the 97th percentile of the grid pattern variance measured during training steps 0--19 as the threshold for the penalty, ensuring that the reward learning stage does not exacerbate edge case quality degradations.
At optimization step $j$, the coefficient $\lambda_g(j)$ is zero before step 20 during calibration, ramps linearly to $3\times10^4$ over steps 20--39, and then stays fixed for the remainder of reward learning.
Only Muon-optimized parameters are learnable; the AdamW-optimized parameters, the text encoder, the VAE, and the reward model are frozen.



\begin{table}[t]
\centering
\footnotesize
\setlength{\tabcolsep}{3pt}
\caption{\textbf{VisionReward questions used in Stage~IV.}
We give the zero-based indices from the original 29-question VisionReward-Video checklist.
We show both published weights $w_q$, unchanged, and effective weights: $w_q/\sum_{q\in\mathcal Q}w_q$ (rounded to 4 decimal places).
For views, ``center'' denotes use of the standard resized center view, while ``three crops'' indicates that we use three native-resolution $224\times224$ windows.}
\label{tab:reward-questions}
\begin{tabular}{@{}rp{0.52\linewidth}rrl@{}}
\toprule
Index & Question & $w_q$ & Effective & Views \\
\midrule
\multicolumn{5}{l}{\textit{Object structure and persistence}} \\
13 & Is the shape of the object at the beginning of the video completely accurate? & 0.129416 & 0.0582 & Center \\
14 & Does the shape of the object at the beginning have no obvious errors? & 0.098917 & 0.0445 & Center \\
15 & Is the shape of the object at the beginning not chaotic? & 0.188393 & 0.0847 & Center \\
16 & Is the shape of the object perfectly maintained throughout the video? & 0.184434 & 0.0830 & Center \\
17 & Is the shape of the object not chaotic throughout the video? & 0.263553 & 0.1185 & Center \\
\multicolumn{5}{l}{\textit{Motion}} \\
20 & Is the smoothness of the object's movement very good? & 0.025628 & 0.0115 & Center \\
21 & Is the object's movement completely realistic? & 0.438989 & 0.1974 & Center \\
\multicolumn{5}{l}{\textit{Composition}} \\
3 & Is the composition aesthetically pleasing? & 0.034957 & 0.0157 & Center \\
4 & Does the composition have no obvious flaws? & 0.025237 & 0.0114 & Center \\
\multicolumn{5}{l}{\textit{Perceptual detail}} \\
22 & Is the image quality very stable? & 0.268577 & 0.1208 & Three crops \\
23 & Are the details very refined? & 0.429252 & 0.1931 & Three crops \\
24 & Are the details not rough? & 0.008462 & 0.0038 & Three crops \\
25 & Are the details not significantly rough? & 0.127573 & 0.0574 & Three crops \\
\bottomrule
\end{tabular}
\end{table}

\section{Optimization}
\label{app:optimizer-algorithm}
\label{sec:per-head-muon}


We optimize the LDM with a hybrid optimizer: Muon~\citep{jordan2024muon,liu2025muonscalable} for most matrix-shaped weights and AdamW~\citep{loshchilov2019decoupled} for the rest.
For query, key and value projections, we apply Muon on a per-head basis (Methods, Eq.~\ref{eq:muon-per-head}).
This section gives the update rule and the division of parameters between the two optimizers.
We provide an empirical comparison with pure AdamW optimization in Supplementary Information~\ref{sec:optimizer-ablation}.

\subsection{Update rule}

For a weight matrix $W$ with gradient $G_t$ at step $t$, after clipping, we keep a momentum buffer $B_t$ and form a Nesterov-style update $S_t$,
\begin{equation}
B_t=\mu B_{t-1}+(1-\mu)G_t,\qquad S_t=(1-\mu)G_t+\mu B_t,
\label{eq:muon-momentum}
\end{equation}
with $B_0=0$ and $\mu=0.95$.
We use $K=5$ quintic Newton--Schulz iterations,
\begin{equation}
X_0=\frac{S}{\max(\lVert S\rVert_F,\,10^{-7})},\qquad
A_i=X_iX_i^\top,\qquad
X_{i+1}=aX_i+bA_iX_i+cA_i^2X_i,
\label{eq:muon-ns}
\end{equation}
for $i=0,\dots,K-1$, with $(a,b,c)=(3.4445,-4.7750,2.0315)$, and write $\mathcal N_K(S)=X_K$.
Tall matrices are transposed before the iteration and back afterwards, so that the Gram matrix has the smaller dimension.

Each matrix block $W^{(j)}\in\mathbb R^{r_j\times c_j}$ processed by Muon is updated as
\begin{equation}
W^{(j)}_t=(1-\eta_t\lambda_{\mathrm{wd}})\,W^{(j)}_{t-1}-\eta_t\,\varrho(r_j,c_j)\,\mathcal N_K\bigl(S^{(j)}_t\bigr),\qquad \varrho(r,c)=0.2\sqrt{\max(r,c)}.
\label{eq:muon-scaled-update}
\end{equation}
The dimensions $r$ and $c$ refer to the block being orthogonalized.
For each query, key, or value head, this block has shape $d_h\times d$, giving a scaling factor $\varrho(d_h,d)=0.2\sqrt d=12.8$.
Weight decay uses the base rate $\eta_t$ and is not multiplied by $\varrho$.
The remaining parameters use AdamW with $(\beta_1,\beta_2)=(0.95,0.95)$ and $\epsilon=10^{-8}$; both optimizers share the schedule $\eta_t$ and the weight decay $\lambda_{\mathrm{wd}}$, which also applies to biases and normalization gains in the AdamW partition.

\subsection{Partition between Muon and AdamW}

Muon updates every trainable tensor with two dimensions except those in an explicitly excluded set $\mathcal E$, and AdamW updates the rest:
\begin{equation}
\Theta_{\mathrm{Muon}}=\{\vartheta\in\Theta_{\mathrm{train}}:\ \operatorname{ndim}(\vartheta)=2,\ \vartheta\notin\mathcal E\},\qquad
\Theta_{\mathrm{AdamW}}=\Theta_{\mathrm{train}}\setminus\Theta_{\mathrm{Muon}}.
\label{eq:muon-parameter-partition}
\end{equation}
Supplementary Table~\ref{tab:muon-partition} lists the families of parameters.
In the joint-attention blocks, the fused query-key-value weight of both modalities has shape $6d\times d$ -- its rows are partitioned into $6H$ blocks of shape $d_h\times d$, indexed by modality, projection, and head, and each block is orthogonalized independently.
The text projector's fused query-key-value weight is partitioned in the same way into $3H$ blocks.
The fused attention output weight, of shape $2d\times d$, is split into two $d\times d$ matrices, one per modality, each orthogonalized as a whole so as to retain its mixing across heads.
Feed-forward weights are stored as separate matrices for the two modalities and receive matrix-wise Muon.

\begin{table}[h]
\centering
\small
\renewcommand{\arraystretch}{1.1}
\caption{\textbf{Partition of trainable parameters between Muon and AdamW.} Families marked $\mathcal E$ are excluded from Muon, including their two-dimensional weights.}
\label{tab:muon-partition}
\begin{tabularx}{\linewidth}{@{}p{0.36\linewidth}p{0.12\linewidth}X@{}}
\toprule
Parameter family & Optimizer & Scope of update \\
\midrule
Video and text Q/K/V weights & Muon & Per projection, modality and head \\
Text-projector Q/K/V weights & Muon & Per projection and head \\
Video and text attention output weights & Muon & One matrix per modality \\
Video and text feed-forward weights & Muon & Each weight matrix \\
Text-projector input projection, attention output and feed-forward weights & Muon & Each weight matrix \\
\midrule
Patch embedding ($\mathcal E$) & AdamW & Convolution kernel and bias \\
Token-source embedding ($\mathcal E$) & AdamW & $3\times d$ lookup table; only the caption row is used \\
Diffusion-time embedding ($\mathcal E$) & AdamW & All weights and biases of the time MLP \\
Modulation and gates ($\mathcal E$) & AdamW & Projections generating shifts, scales and gates; learned text modulation vectors \\
Output head ($\mathcal E$) & AdamW & Adaptive-normalization and output projections, with biases \\
Biases & AdamW & Q/K/V, attention output, feed-forward and text-projector biases \\
Normalization gains & AdamW & Q/K RMSNorm gains; text-projector RMSNorm gains \\
Other tensors not of rank two & AdamW & Any remaining scalars, vectors or higher-rank tensors \\
\bottomrule
\end{tabularx}
\par\smallskip
\end{table}

\section{Studying simulation capabilities}
\label{app:two-tier-evaluation-details}

\subsection{Shared protocol}
\label{app:shared-protocol}

\paragraph{Models, conceptual backbones, and seeds.}

We consider each model as a complete GLP-style system, and run it with its recommended conceptual backbone (prompt rewriter): Cosmos~3 Super uses Cosmos Reasoner, LTX-2.5 uses Gemma-4 E2B, while we send requests for FLUX~3 and Seedance~2.5 through fal API, meaning they both use their own proprietary prompt handling.
PAN~2 uses Kimi-K3, and for MiniMax-H3, whose exact rewriter is proprietary and rejects our test cases due to length, we also use Kimi-K3 with H3's official prompt skill to enable reproduction with confidence intervals for reliable comparisons.
We rewrite a single time only for each test case for each model, and share the rewritten prompt across seeds for the given model.
We use three seeds (123, 124, 125) each for the open weight models, and a single seed for each proprietary model (fal API offers no control over seed).

\paragraph{Inference settings.}

We generate each simulation for the duration each test case specifies, always at 24 frames per second.
PAN~2 generates at $704{\times}1248$ following our settings in Supplementary Table~\ref{tab:ldm-sampling}.
H3 uses 50 steps at 768p, with aspect ratio preservation for image-to-video generation.
Cosmos~3 generates at $1280\times720$ with 35 steps for image-to-video and 50 steps for text-to-video.
LTX-2.5 uses $1280\times704$ with its distilled pipeline.
For Seedance~2.5 and FLUX~3, we use the fal API default settings.

\paragraph{Judge validation.}

We measure and report human annotator agreement in Extended Data Table~\ref{tab:judge-validation}.
We show that agreement between humans and the auto-grader, as well as among humans, are both fairly high.
This enables us to draw conclusions based on our results.


\paragraph{Uncertainty.}

Unless stated otherwise, all scores average results over three seeds.
We compute 95\% bootstrap confidence intervals over test cases, not seeds.
Thus, when we show intervals, they quantify test-case variation, not randomness due to seeds.
By contrast, in the few cases where we show $\pm$, this denotes the standard deviation over seeds.


\subsection{Single-action test cases}
\label{app:tier1-capabilities-details}


Extended Data Fig.~\ref{fig:cap-method} illustrates how the single-action test cases are built and scored.
We provide more details on how and why they are built, and how they are scored, below.

\paragraph{Construction.}

We start construction by considering each of the 6D-PI separately, mining and creating cases that will adequately test a primary capability, with 50 cases per capability.
We then label which capabilities are actually tested by each case.
Most cases test multiple capabilities: 281 test at least 2, 106 cover at least 3, with an average of 2.43 capabilities tagged per sample.
The most common capability pairings are objective dynamism and physical consistency (84 cases), objective dynamism with entity persistence (76), and viewpoint invariance with entity persistence (71).
198 cases require a 20-second simulations and the rest are 10 seconds. Every model's outputs must cover the full duration, since many test cases have events and actions that are time-sensitive.


\paragraph{Filling gaps in public benchmarks.}

In terms of our 6D-PI, existing public world model benchmarks focus mainly on physical consistency, but even those lack adequate coverage for sliding friction, springs, levers, gears, pulleys, refraction, lenses, melting, and freezing.
We wrote 31 new test cases for physical consistency to fill in these gaps.
For example, we have multiple samples covering mirror optics: hinged mirrors, the reversal of mirrored text, and breath fogging a mirror.
Existing benchmarks also consider temporal order~\citep{t2vcompbench,vbench2}, but mainly consider whether prompted events occur in order.
Our test cases instead test causal coherence more directly by tying the order to a physical precondition or an earlier world state, such as by simulating a stool being unfolded before a jar on a high shelf can be reached.


\paragraph{Viewpoint invariance.}

Viewpoint invariance is neglected in existing public benchmarks.
Physics-IQ uses a stationary camera, and VBench-2.0 applies a single move (pan, tilt, zoom, orbit or dolly) to a largely static scene.
Ego-centric viewpoints seems particularly negledted: using keyword searches, we do not find any coverage for in-place $360^\circ$ or $180^\circ$ turns.
We have 121 total cases (from 80 base cases) that test yaw, pitch, and/or translation.
More specificaly, we test in-place $360^\circ$ turns (6 base cases), $180^\circ$ turns (4), full and partial orbits (13), turning away and then returning to the original view (11), tilting up and down (11), drone or crane moves up and down (5), moves towards and away from a subject (13), sideways tracking (11) and following a moving subject (9).
We use a mix of first-person, third-person, and aerial views.
We chain multiple camera moves so some cases count under several moves.
Because VLMs judge camera motion poorly (Supplementary Information~\ref{subsubsec:camera_ablation}), we judge viewpoint invariance from camera geometry with our VGGT-$\Omega$ pipeline.


\paragraph{Diversity and selection.}
The 300 test cases come from 204 base test cases, 96 of which are posed both with an image and as text alone.
The 86 newly written base test cases span 16 domains, with at most 8 per domain (for example farm animals, kitchens, sports, construction, crowds, laboratories, coastlines, and snow and ice); actions range from 6 to 129 words, and 18 are written in the first person.
We condensed a pool of 387 test cases to 300, 50 per
primary dimension,
drafting round-robin from the capability with fewest test cases, keeping one test case from each pair that differs only in the order of actions, skipping near-duplicates (maximum remaining TF-IDF similarity 0.44), and dropping 58 test cases that every model passed or every model failed.

\paragraph{Scoring.}

Unlike benchmarks~\citep{vbench,evalcrafter} that are common for video generation, we do not score visual quality directly.
We judge whether the simulation proceeds correctly, not whether the video has appealing aesthetics or lacks visual artifacts.
The judge answers whether the main action was completed, and any questions for tagged capabilities (except for viewpoint), in a single pass.
Questions on camera motion are resolved by comparing the trajectories estimated by VGGT-$\Omega$ with the viewpoint requirements for each specific test case, following the thresholds in Supplementary Table~\ref{tab:camera-thresholds}.
Pans and tilts require rates of at least $2.2^\circ\,\mathrm{s}^{-1}$ for at least 0.6 seconds.
A full turn requires at least $270^\circ$ net rotation; the single-action turn to something behind the camera requires at least $150^\circ$ net rotation or $300^\circ$ cumulative rotation.
Turn-and-return requirements vary by case.


\paragraph{Confidence intervals.}
Confidence intervals use 10,000 percentile bootstrap draws over the 204 base cases, retaining all questions, seeds and paired modalities of a drawn case.
Modality-specific tables resample test cases within that modality.
Multi-step intervals resample the 100 plans with both modalities and all seeds.
Paired differences use the same draws for both conditions.
These intervals quantify test-case variation for the observed seeds, not variation over new generations.

\paragraph{Results by modality.}
\label{app:capability-modality-results}

Supplementary Table~\ref{tab:cap-full-modality-methods} shows scores broken down by each starting-state modality, I2V compared to T2V.

\begin{table*}[t]
\centering
\small
\setlength{\tabcolsep}{4pt}
\caption{\textbf{Single-action test cases by starting-state modality.} Image and text (I2V): 201 test cases; text only (T2V): 99. Scores (\%) are means over seeds 123--125, with 95\% percentile bootstrap confidence intervals in brackets. $^\dagger$Generated once through the fal API; intervals condition on that generation. Seedance~2.5 did not return 12 I2V simulations, which are excluded.}
\label{tab:cap-full-modality-methods}
\resizebox{\textwidth}{!}{
\begin{tabular}{llccccccc}
\toprule
Modality & Model & Overall & \shortstack{Physical\\consistency} & \shortstack{Causal\\coherence} & \shortstack{Entity\\persistence} & \shortstack{Constraint\\satisfaction} & \shortstack{Viewpoint\\control} & \shortstack{External\\dynamism} \\
\midrule
\multirow{6}{*}{I2V} & Cosmos 3 & 66.91\,{\scriptsize[63.08, 70.62]} & 62.00\,{\scriptsize[52.28, 71.19]} & 63.85\,{\scriptsize[54.76, 72.76]} & 74.16\,{\scriptsize[66.57, 81.25]} & 64.83\,{\scriptsize[55.32, 73.73]} & 40.00\,{\scriptsize[32.00, 47.97]} & 87.76\,{\scriptsize[82.71, 92.35]} \\
 & LTX-2.5 & 60.87\,{\scriptsize[56.25, 65.37]} & 47.37\,{\scriptsize[37.44, 57.01]} & 51.69\,{\scriptsize[42.05, 61.20]} & 67.90\,{\scriptsize[58.76, 76.42]} & 51.79\,{\scriptsize[41.86, 61.63]} & 59.17\,{\scriptsize[49.25, 69.05]} & 77.75\,{\scriptsize[70.86, 84.31]} \\
 & H3 & 86.82\,{\scriptsize[84.18, 89.34]} & 79.22\,{\scriptsize[70.89, 86.76]} & 86.79\,{\scriptsize[80.48, 92.58]} & 96.44\,{\scriptsize[92.84, 99.24]} & 83.33\,{\scriptsize[75.98, 90.22]} & 76.67\,{\scriptsize[68.98, 83.77]} & 94.38\,{\scriptsize[90.14, 97.96]} \\
 & PAN~2 & 83.55\,{\scriptsize[80.31, 86.67]} & 76.19\,{\scriptsize[67.12, 84.62]} & 81.13\,{\scriptsize[73.01, 88.82]} & 89.92\,{\scriptsize[84.39, 94.62]} & 80.08\,{\scriptsize[72.35, 87.20]} & 75.42\,{\scriptsize[67.13, 83.12]} & 93.49\,{\scriptsize[89.94, 96.62]} \\
\cmidrule{2-9}
 & FLUX~3$^\dagger$ & 78.51\,{\scriptsize[74.65, 82.20]} & 77.92\,{\scriptsize[68.23, 86.90]} & 78.87\,{\scriptsize[69.01, 88.33]} & 94.05\,{\scriptsize[88.46, 98.75]} & 70.27\,{\scriptsize[59.21, 80.30]} & 43.75\,{\scriptsize[32.91, 54.67]} & 97.32\,{\scriptsize[93.97, 100.00]} \\
 & Seedance~2.5$^\dagger$ & 77.61\,{\scriptsize[73.32, 81.67]} & 71.23\,{\scriptsize[60.38, 81.43]} & 82.09\,{\scriptsize[72.06, 90.77]} & 85.71\,{\scriptsize[77.27, 93.06]} & 67.12\,{\scriptsize[56.10, 77.92]} & 56.94\,{\scriptsize[45.45, 68.25]} & 94.39\,{\scriptsize[89.91, 98.20]} \\
\midrule
\multirow{6}{*}{T2V} & Cosmos 3 & 71.32\,{\scriptsize[65.68, 76.64]} & 65.00\,{\scriptsize[51.61, 77.48]} & 72.42\,{\scriptsize[58.33, 85.06]} & 82.06\,{\scriptsize[71.43, 91.18]} & 72.47\,{\scriptsize[60.18, 83.61]} & 48.78\,{\scriptsize[35.66, 62.04]} & 86.27\,{\scriptsize[78.00, 93.33]} \\
 & LTX-2.5 & 62.25\,{\scriptsize[55.80, 68.55]} & 44.33\,{\scriptsize[31.29, 57.29]} & 64.17\,{\scriptsize[49.96, 77.88]} & 60.46\,{\scriptsize[46.74, 73.54]} & 63.99\,{\scriptsize[51.17, 76.14]} & 61.79\,{\scriptsize[48.33, 74.80]} & 74.90\,{\scriptsize[64.72, 84.40]} \\
 & H3 & 88.01\,{\scriptsize[84.53, 91.27]} & 78.16\,{\scriptsize[67.50, 88.06]} & 83.91\,{\scriptsize[73.61, 92.60]} & 97.88\,{\scriptsize[92.85, 100.00]} & 90.92\,{\scriptsize[83.21, 97.16]} & 85.37\,{\scriptsize[75.93, 93.50]} & 92.00\,{\scriptsize[84.80, 97.87]} \\
 & PAN~2 & 86.48\,{\scriptsize[82.83, 89.85]} & 78.33\,{\scriptsize[66.67, 88.89]} & 82.76\,{\scriptsize[73.74, 91.23]} & 91.66\,{\scriptsize[85.42, 96.97]} & 89.19\,{\scriptsize[80.21, 96.58]} & 82.93\,{\scriptsize[72.81, 91.67]} & 92.76\,{\scriptsize[86.55, 97.78]} \\
\cmidrule{2-9}
 & FLUX~3$^\dagger$ & 75.00\,{\scriptsize[68.16, 81.39]} & 66.67\,{\scriptsize[51.28, 80.95]} & 78.57\,{\scriptsize[61.76, 92.59]} & 90.62\,{\scriptsize[78.95, 100.00]} & 75.68\,{\scriptsize[60.98, 88.89]} & 53.66\,{\scriptsize[38.30, 68.75]} & 86.27\,{\scriptsize[76.36, 95.24]} \\
 & Seedance~2.5$^\dagger$ & 80.62\,{\scriptsize[75.12, 85.65]} & 67.50\,{\scriptsize[52.50, 81.58]} & 75.86\,{\scriptsize[59.26, 90.91]} & 90.32\,{\scriptsize[78.57, 100.00]} & 86.11\,{\scriptsize[73.81, 96.77]} & 73.17\,{\scriptsize[58.97, 86.27]} & 90.00\,{\scriptsize[80.49, 97.92]} \\
\bottomrule

\end{tabular}}
\end{table*}

\subsection{Multi-step plans}
\label{app:tsr-details}

\paragraph{Plans.}
Plans include refilling a bird feeder or taking a dog out (fetching the leash from behind, clipping it on, opening the back door and going out).
Models tend to fall back on shortcuts (e.g., a partial turn, a small pan that suggests motion, a hard cut or transition, or the target simply appearing in front of the camera).
Every plan is designed to be feasible from its starting state; plans that should fail, and how a model resolves them, are left to future work (see Discussion).

\paragraph{Scoring.}
Camera motion is checked from the latest time of an earlier completed step onward.

Every camera step except the 11 pure-movement steps also requires the judge to confirm the step; for a turn towards a target, this includes confirming that the target is faced.

\paragraph{Dimensions used.}
\label{app:tsr-characterization}
Each plan draws on several dimensions together.
Of the 456 steps, 202 are tagged with constraint satisfaction and 171 with
viewpoint invariance
(the person turns to find objects), and fewer with physical consistency (40), causal coherence (33),
objective dynamism
(6) and entity persistence (4; objects the person looked away from must still be there).
Each plan uses two to four dimensions
(2.63 on average);
viewpoint invariance
appears in 99 plans and constraint satisfaction in 95.

\paragraph{Relation to other benchmarks.}
Supplementary Table~\ref{tab:tsr-coverage} compares the multi-step plans with public video-generation and world-model benchmarks on six properties.
Several recent procedural and egocentric benchmarks~\citep{ego2act,spiral} check goal-directed multi-step tasks step by step, but none requires movement to targets that start out of view verified by camera geometry, and none has all six properties.
The closest is the concurrent Ego2Act~\citep{ego2act}, which also poses goal-directed egocentric tasks and checks each subgoal separately.
Our plans differ in requiring movement to targets that start out of view, verified by camera geometry, and in posing every plan both with an image and as text alone (Ego2Act conditions every case on a starting frame); our simulations are also longer by design (20~s; Ego2Act caps generation at 15~s).

\begin{table}[t]
\centering
\scriptsize
\setlength{\tabcolsep}{3pt}
\caption{\textbf{Properties of the multi-step plans in public video-generation and world-model benchmarks.}
Multi-step: at least three ordered actions in one simulation.
Goal task: the actions together complete one meaningful, goal-directed task.
Ego.: first-person viewpoint.
Per-step: each step is checked separately rather than scored once per video.
Long clip: $\geq$15\,s in one clip by design (\pmark{} = 10--15\,s or a subset; \xmark{} = model default, typically $<$10\,s).
Geo.\ nav.: movement to a target that starts out of view, verified by camera-pose estimation (\pmark{} = pose-estimated trajectory accuracy without targets).
\cmark{} = yes, \pmark{} = partially or for a subset, \xmark{} = no. Every cell was checked against the benchmark's own paper.}
\label{tab:tsr-coverage}
\begin{tabular}{lcccccc}
\toprule
Benchmark & Multi-step & Goal task & Ego. & Per-step & Long clip & Geo.\ nav. \\
\midrule
\multicolumn{7}{l}{\emph{Video generation: quality, physics, composition}} \\
VBench-2.0~\citep{vbench2}                & \pmark & \pmark & \xmark & \pmark & \xmark & \xmark \\
T2V-CompBench~\citep{t2vcompbench}        & \xmark & \xmark & \xmark & \pmark & \xmark & \xmark \\
Physics-IQ~\citep{physicsiq}              & \xmark & \xmark & \xmark & \xmark & \xmark & \xmark \\
PhyGenBench~\citep{phygenbench}           & \xmark & \xmark & \xmark & \pmark & \xmark & \xmark \\
\multicolumn{7}{l}{\emph{Multi-event, temporal and procedural generation}} \\
StoryEval~\citep{temporal}                & \pmark & \pmark & \xmark & \cmark & \pmark & \xmark \\
TC-Bench~\citep{tcbench}                  & \xmark & \pmark & \xmark & \pmark & \xmark & \xmark \\
SeqBench~\citep{seqbench}                 & \pmark & \xmark & \xmark & \cmark & \xmark & \xmark \\
LabInstruct~\citep{labinstruct}           & \pmark & \cmark & \pmark & \cmark & \pmark & \xmark \\
ActVideoGen-Bench~\citep{spiral}          & \cmark & \cmark & \pmark & \cmark & \pmark & \xmark \\
Ego2Act~\citep{ego2act}                   & \cmark & \cmark & \cmark & \cmark & \cmark & \xmark \\
\multicolumn{7}{l}{\emph{World models: general, embodied and interactive}} \\
WorldModelBench~\citep{worldmodelbench}   & \xmark & \pmark & \pmark & \xmark & \xmark & \xmark \\
WorldScore~\citep{worldscore}             & \xmark & \xmark & \xmark & \pmark & \xmark & \pmark \\
WorldSimBench~\citep{worldsimbench}       & \xmark & \pmark & \pmark & \pmark & \xmark & \xmark \\
PAI-Bench-G~\citep{paibench}              & \pmark & \pmark & \pmark & \pmark & \xmark & \xmark \\
WorldArena~\citep{worldarena}             & \pmark & \pmark & \xmark & \pmark & \xmark & \xmark \\
EWMBench~\citep{ewmbench}                 & \xmark & \pmark & \cmark & \pmark & \xmark & \xmark \\
WR-Arena~\citep{wrarena}                  & \pmark & \pmark & \pmark & \cmark & \pmark & \xmark \\
WBench~\citep{wbench}                     & \pmark & \pmark & \cmark & \cmark & \pmark & \pmark \\
MIND~\citep{mindbench}                    & \pmark & \xmark & \pmark & \xmark & \cmark & \pmark \\
\midrule
Multi-step plans (this study)                   & \cmark & \cmark & \cmark & \cmark & \cmark & \cmark \\
\multicolumn{7}{l}{\quad 100 plans, 200 simulations (image and text, text only), 456 steps (3--6 per plan), 20-s simulations} \\
\bottomrule
\end{tabular}
\end{table}

\begin{table}[!t]
\centering
\small
\caption{\textbf{Camera-motion requirements by move type.} Thresholds are set per case and stored with each test case; the table groups the stored values. Camera motion is estimated with VGGT-$\Omega$ at 2 frames per second. A pan or tilt is detected when the yaw or pitch rate is at least 2.2$^\circ$\,s$^{-1}$; translations (truck, pedestal, dolly) are detected from the camera's translation speed. Zoom is not tested. A hard scene cut fails single-action camera questions only. }
\label{tab:camera-thresholds}
\begin{tabularx}{\linewidth}{@{}>{\raggedright\arraybackslash}p{0.24\linewidth}>{\raggedright\arraybackslash}X>{\raggedright\arraybackslash}p{0.25\linewidth}@{}}
\toprule
Move & Requirement & Cases \\
\midrule
\multicolumn{3}{@{}l}{\textit{Single-action test cases (121 camera questions: 80 image + text, 41 text)}} \\
\addlinespace[2pt]
Pan, tilt, truck, pedestal, dolly, tracking & The stated motion for at least 0.6\,s, in the stated order (with little opposite motion); where the prompt states times, the motion must cover at least half of the stated window. Some cases set a longer minimum duration (0.5--4\,s). & Most questions \\
Orbit / circle around & Orbit in either direction; cam\_v2\_02/04 also need a net rotation $\geq$ 270$^\circ$, camp\_06 and camp\_12 a cumulative rotation $\geq$ 90$^\circ$ and $\geq$ 270$^\circ$ & cam\_v2\_02/04/06/21, camo\_01, camc\_10, camh\_05/07, camp\_06/12, camr\_01/09 \\
Full circle / all the way around & Net rotation $\geq$ 270$^\circ$ & cam\_v2\_13/18/19/20, camo\_06, camp\_09 \\
Turn to face what is behind & Net rotation $\geq$ 150$^\circ$, or cumulative $\geq$ 300$^\circ$ & cam\_v2\_07 \\
Turn around and back & Cumulative rotation $\geq$ 300$^\circ$ (camp\_02), 270$^\circ$ (op\_3, op\_9$^{a}$), 200$^\circ$ (op\_19), 150$^\circ$ (op\_20) & camp\_02, op\_3/9/19/20 \\
Turn away and back (small turns) & The two pans in order; net rotation $\leq$ 45$^\circ$ (cam\_v2\_15 also cumulative $\geq$ 90$^\circ$ for image + text) & cam\_v2\_15/16 \\
Follow a subject there and back & The follow movements in order$^{a}$ & op\_1/4/11/14/16/18 \\
Turn to face a target & The motion above \emph{and} the video judge confirms the camera ends facing the target & 82 of 121 \\
\midrule
\multicolumn{3}{@{}l}{\textit{Multi-step plans (171 camera steps)}} \\
\addlinespace[2pt]
Pan left/right, pan back & Peak rotation $\geq$ 10$^\circ$ in the stated direction, measured from the end of the previous step, or the matching truck motion for 0.6\,s & 76 \\
Tilt up/down & Peak rotation $\geq$ 8$^\circ$, or the matching pedestal motion for 0.6\,s & 29 \\
Turn around / turn back & Peak rotation $\geq$ 150$^\circ$ in either direction & 43 \\
Climb, descend, walk forward & The matching translation for 0.6\,s & 16 \\
Combined or look back & Each component above & 7 \\
\addlinespace[2pt]
\multicolumn{3}{@{}p{\linewidth}@{}}{In multi-step plans every camera step except pure movement steps (11) must also be confirmed by the video judge.} \\
\bottomrule
\end{tabularx}
\par\smallskip
\raggedright\footnotesize $^{a}$The end direction is not checked for these cases: their stored requirements state a limit on net rotation ($\leq$ 90$^\circ$ for op\_9, $\leq$ 45$^\circ$ otherwise), but they are scored on the order of the movements only.
\end{table}

\paragraph{Results by modality and conceptual reasoning backbone.}
\label{app:tsr-modality-results}
Supplementary Table~\ref{tab:tsr-full-methods} gives step completion for each modality and full-plan completion, including PAN~2 with two other LLMs as its conceptual reasoning backbone.

\begin{table}[t]
\centering
\small
\setlength{\tabcolsep}{3pt}
\caption{\textbf{Multi-step plans by starting-state modality.} 100 plans, each posed with an image and text (I2V) and as text alone (T2V). Step completion: mean fraction of steps completed per simulation (\%). Full-plan completion: share of simulations in which every step is completed, both modalities pooled (\%). Center [95\% CI]: center = mean over seeds 123--125; CI = test-case bootstrap over the 100 plans (10{,}000 resamples; a drawn plan brings its I2V and T2V simulations and all seeds; seeds averaged within a plan). Rewriters as in Supplementary Information~\ref{app:shared-protocol}; PAN~2 uses Kimi-K3 unless another LLM is named. $^\dagger$Generated once through the fal API without a rewriter, excluding simulations the API did not return (2 and 4).}
\label{tab:tsr-full-methods}
\begin{tabular}{lcccc}
\toprule
 & \multicolumn{3}{c}{Step completion} & \\
\cmidrule(lr){2-4}
Model & All & I2V & T2V & Full-plan completion \\
\midrule
Cosmos~3 & 47.74\,{\scriptsize[43.54, 52.08]} & 44.53\,{\scriptsize[39.82, 49.34]} & 50.96\,{\scriptsize[46.26, 55.66]} & 9.50\,{\scriptsize[5.33, 14.17]} \\
LTX-2.5 & 38.46\,{\scriptsize[34.45, 42.55]} & 39.90\,{\scriptsize[35.14, 44.70]} & 37.02\,{\scriptsize[32.76, 41.38]} & 4.00\,{\scriptsize[1.50, 7.00]} \\
MiniMax-H3 & 75.12\,{\scriptsize[71.81, 78.31]} & 76.21\,{\scriptsize[72.04, 80.13]} & 74.03\,{\scriptsize[70.40, 77.58]} & 32.50\,{\scriptsize[26.00, 39.17]} \\
PAN~2 & 75.10\,{\scriptsize[71.65, 78.51]} & 76.66\,{\scriptsize[72.67, 80.57]} & 73.53\,{\scriptsize[69.84, 77.20]} & 34.00\,{\scriptsize[27.17, 41.17]} \\
\midrule
FLUX~3$^\dagger$ & 72.44\,{\scriptsize[68.63, 76.18]} & 74.28\,{\scriptsize[69.85, 78.59]} & 70.61\,{\scriptsize[66.25, 74.95]} & 26.77\,{\scriptsize[19.70, 34.18]} \\
Seedance~2.5$^\dagger$ & 68.95\,{\scriptsize[64.88, 72.89]} & 72.53\,{\scriptsize[67.55, 77.27]} & 65.50\,{\scriptsize[60.73, 70.13]} & 24.49\,{\scriptsize[17.59, 31.63]} \\
\midrule
\multicolumn{5}{l}{\emph{PAN~2 with other LLMs as its conceptual reasoning backbone}} \\
GPT-6 Astra & 78.35\,{\scriptsize[75.06, 81.59]} & 79.47\,{\scriptsize[76.07, 82.75]} & 77.22\,{\scriptsize[73.45, 80.92]} & 38.33\,{\scriptsize[31.33, 45.50]} \\
Claude Opus 5.5 & 80.93\,{\scriptsize[77.79, 84.02]} & 82.29\,{\scriptsize[79.05, 85.43]} & 79.56\,{\scriptsize[76.10, 82.97]} & 43.17\,{\scriptsize[36.17, 50.50]} \\
\bottomrule
\end{tabular}
\end{table}

\subsubsection{Compounding of step failures}
\label{app:compounding}

A simulation counts towards full-plan completion $F$ only if every one of its steps is completed, so $F$ compounds on step completion $S$.
If the $N$ steps of a plan succeeded independently with probability $S$, then $F=\mathbb E[S^N]$, which falls exponentially with the rate of step failure, $\ln F\approx-\mathbb E[N](1-S)$, near $S=1$.
We fit the corresponding form
\begin{equation}
F=S^{k}\,e^{-c(1-S)},
\label{eq:compounding}
\end{equation}
which gives $F=1$ at $S=1$ by definition, to the six models of Supplementary Table~\ref{tab:tsr-full-methods} by least squares on $\ln F$.
With the current six model results, the fit gives $k=1.67$, $c=2.49$ and $R^2=0.986$ on $\ln F$. Its pointwise 95\% band resamples the 100 plans 10,000 times, retaining all modalities and seeds and using the same draw across models before refitting; the band is shown over the observed range of model scores.

The fit is tight partly because $S$ and $F$ summarize the same step verdicts from the same judge, so the curve largely reflects the benchmark, its plan lengths and how failures cluster, rather than the models.
What it shows is that all six models distribute failures across simulations in about the same way, so one curve predicts full-plan completion from step completion for any of them.
Because the judge is shared, some of the clustering could reflect correlated judge errors rather than failures of whole plans.
On a validation set of 120 human-labelled multi-step simulations from four earlier model configurations, full-plan completion is 20.8\% for both annotators and the judge (paired difference 0.0 points, 95\% CI $[-8.3,8.3]$), at step completion of 62.9\% and 65.4\%, respectively. Agreement on which simulations complete every step is moderate (Cohen's $\kappa=0.34$), and this set is too small to resolve clustering within individual models.
In a separate judge comparison on 200 multi-step simulations of each of PAN~2, MiniMax-H3 and Cosmos~3 (seed 123), using an earlier Kimi-rewritten Cosmos~3 run rather than the Cosmos Reasoner run of Fig.~\ref{fig:main-results}, re-judging with Kimi-K3 raises step completion by 3.7 points (95\% CI $[2.4,5.0]$) but full-plan completion by 2.0 ($[-0.5,4.7]$). Excess full-plan completion over independent-step predictions persists, at 2.9 points ($[0.6,5.1]$), compared with 5.2 ($[2.7,7.3]$) under the production judge.
Part of the clustering therefore depends on the judge; the cross-family check also shares the production questions, negative checks and camera measurements, so it is not fully independent.

Independently of any curve, step completion bounds full-plan completion.
Full-plan completion can be no larger than $S$, and it is smallest when every simulation that fails misses exactly one step and the failures fall on the longest plans, since a failed simulation of $N_i$ steps completes at most $(N_i-1)/N_i$ of them.
From the step counts of the 100 plans, this minimum is about 48\% at $S=90\%$ and 73\% at $S=95\%$ (the dots in Fig.~\ref{fig:main-results}c).

\subsection{Public Benchmarks}
\label{sec:public-benchmarks}

We also study PAN~2's physical and embodied simulation capabilities on three public benchmarks (Extended Data Table~\ref{tab:public-benchmarks}).
Unless stated otherwise, we generate at 720p and 24 frames per second with three seeds (123--125) per prompt and report the mean; for other models we follow the inference settings, negative prompts and schedulers recommended in each model's documentation.
PAN~2 receives each prompt through Kimi-K3 as its conceptual reasoning backbone, and where a benchmark provides an upsampled variant of its prompts, we use that variant, as published leaderboards do.

Physics-IQ~\citep{physicsiq} tests whether a model can continue real recordings of physical experiments, comparing each generated continuation with real recordings of the same scenario through motion metrics (spatial IoU, spatiotemporal IoU, weighted spatial IoU and mean squared error) rather than a VLM judge.
We report its Verified split~\citep{radsch2026verified}, so that the numbers can be compared with the public leaderboard,\footnote{\url{https://github.com/google-deepmind/physics-IQ-benchmark}} in two settings.
In the system-default setting, every model runs its own default pipeline.
In the public-leaderboard setting, models are scored as submitted to the leaderboard.
For PAN~2's leaderboard-setting entry, we follow the practice of FLUX~3's leaderboard submission: Claude Opus 5.5 (medium effort) rewrites each of the benchmark's scene descriptions under a single instruction, adapted from FLUX~3's prompt-rewriting instruction and fixed before any videos were generated, and FLUX~3's scene-specific negative prompts, which list physical failures likely in each scene (for example, objects passing through each other or bouncing higher than they were released), are combined with our standard negative prompt without its generic terms against stillness; we report the mean over four seeds with a 95\% scenario-bootstrap confidence interval.
Scores for Cosmos~3 Super and MiniMax-H3 in the public-leaderboard setting are their leaderboard submissions. LTX-2.5 has no leaderboard entry; its run with the official Gemma-4 E2B prompt enhancer and the proprietary models generated once through the fal API are reported in the system-default column and scored with the official evaluator.

PhyGround~\citep{phyground} consists of 250 curated prompts, each paired with an expected physical outcome, covering 13 physical laws in rigid-body mechanics, fluid dynamics and optics.
Following the benchmark, a VLM judge (here Qwen3.8-27B) scores each video from 1 to 5 from its frames, the augmented prompt and a scoring rubric, with law-specific questions that help it assess whether the expected behavior appears; each criterion is scored separately, in 5 to 7 queries per video (three general dimensions and two to four applicable laws), so that the judge attends to one criterion at a time and each aspect of the video receives its own score.

RBench~\citep{rbench-deng26r} evaluates physical plausibility and instruction following in robotic image-to-video generation, on 650 image--text pairs: 250 task-oriented cases in five categories (common manipulation, long-horizon planning, multi-entity collaboration, spatial relationships and visual reasoning; 50 each) and 400 embodiment-oriented cases in four (single-arm, dual-arm, humanoid and quadruped robots; 100 each).
Task-oriented cases are judged with category-specific multimodal rubrics; embodiment-oriented cases are assessed for physical and semantic plausibility, task adherence, stability of the robot, motion amplitude and smoothness, which are combined into task-completion and visual-quality scores.
We report the unweighted mean of the nine category scores, using Kimi-K3 as PAN~2's conceptual reasoning backbone.


\subsection{Human preference}
\label{sec:human-eval}

The single-action test cases and multi-step plans are scored question by question, which makes them scalable and reproducible and attributes each failure to
a specific dimension or step.
A person viewing a simulation, in contrast, judges it as a whole, attending at once to the action, the physics and the coherence of the world, including conditions that must hold in every frame rather than at a discrete event.
We therefore complement the automatic studies with blind pairwise comparisons by human reviewers.

\paragraph{Prompts.}
The 140 prompts, 90 given as text alone and 50 with an image, are drawn from public benchmarks, from our test cases and from prompts newly written for this study; the reused prompts tie the human comparisons to the automatic studies, and the new ones extend them to what those studies cannot score.
The new prompts cover what the automatic studies score less well.
They emphasize sustained interaction and dexterous manipulation, such as a 20-second badminton rally, a judo throw, pleating a dumpling, folding a shirt, placing puzzle pieces in sequence and a humanoid robot unloading a dishwasher, in which contacts must be maintained, objects must keep their identity and agents must keep moving and reacting for the whole clip, whose length follows the natural duration of the action (6 to 20~s).
Their specificity ranges from a single under-specified sentence (``a blue ball rolls off the table'') to a paragraph describing the starting frame and what follows, so the set also tests what a model supplies on its own and whether it follows every stated detail.

\paragraph{Protocol.}
We perform blind pairwise comparisons covering three open-weight models, Cosmos~3 Super, MiniMax-H3 and LTX-2.5, and PAN~2, with one simulation per model and prompt.
No model receives the prompt verbatim: an LLM rewrites each prompt into each model's recommended format, and into the structured text-based state for PAN~2.
In this study, PAN~2's conceptual reasoning backbone is GPT-6 Astra at medium effort, and a second configuration with Claude Opus 5.5 is included in the same comparisons.
Cosmos~3 Super uses its recommended Cosmos Reasoner and LTX-2.5 its recommended Gemma-4 E2B prompt enhancer; MiniMax-H3 uses Kimi-K3 with its official prompt skill.
Reviewers see two simulations side by side, labeled only A and B, beneath the original prompt, and give a single overall vote for A, a tie, or B, considering which best satisfies three criteria together.
\emph{Action adherence}, covering the prompted action and
viewpoint invariance:
the video shows the subjects, actions and camera movement the prompt specifies, and the scene stays spatially coherent as the camera moves.
\emph{Physics}, covering physical consistency and constraint satisfaction: materials, fluids, heat, light and gravity behave as they would in reality, and obstacles, support, reach and clearance constrain what can happen.
\emph{World consistency}, covering entity persistence, causal coherence and
objective dynamism:
objects keep their identity and state even when hidden or out of view, outcomes follow from the order of earlier actions, and entities that move on their own keep moving naturally and respond to events rather than freezing.
A pool of 13 reviewers contributes 1,400 votes on 925 distinct configuration-pair/prompt comparisons across 140 prompts, with A and B positions randomized. Of these comparisons, 375 are rated by multiple reviewers, giving 594 shared reviewer-pair judgments; 298 votes are ties.

\paragraph{Results.}
We fit Elo ratings to all comparisons with a Bradley--Terry model (Supplementary Table~\ref{tab:human-eval-elo}) and report each configuration's preference rate, with a win contributing one, a tie one-half and a loss zero.
Mean pairwise inter-reviewer agreement is 59.3\% (95\% CI [52.9, 66.9]), the unweighted mean exact-label agreement over 71 reviewer pairs with shared comparisons; two tie votes count as agreement. The pooled agreement across all 594 shared reviewer-pair judgments is 58.4\% (95\% CI [53.2, 63.4]).
We use 10,000 percentile bootstrap resamples of whole prompts (seed 20261029), keeping all configuration comparisons and reviewer votes for each prompt together and the reviewer panel fixed. Elo ratings are refitted on each draw; preference and agreement intervals use the same draws. These pointwise intervals measure variation across test prompts conditional on the observed reviewers, generated clips and assignment mixture.

MiniMax-H3 and the two PAN~2 configurations are rated within about 60 Elo points of each other (1559 to 1620), and Cosmos~3 Super and LTX-2.5 about 200 points lower (1346 to 1380).

\begin{table}[t]
\centering
\small
\caption{\textbf{Blind pairwise human preference.} Bradley--Terry maximum-likelihood ratings (base 1500, scale 400, regularization 0.5), rounded to whole numbers, and preference rates (\%), counting a win as one, a tie as one-half and a loss as zero. Brackets give pointwise 95\% percentile intervals from 10,000 whole-prompt bootstrap resamples (seed 20261029), with the 13 reviewers held fixed and Elo refitted on each draw. PAN~2 uses GPT-6 Astra at medium effort unless Claude Opus 5.5 is named. No Kimi-K3 configuration is included.
}
\label{tab:human-eval-elo}
\begin{tabular}{lcc}
\toprule
Model & Elo [95\% CI] & Preference rate [95\% CI] \\
\midrule
MiniMax-H3 & \shortstack{1620\\{\scriptsize[1593, 1650]}} & \shortstack{68.8\\{\scriptsize[64.7, 72.9]}} \\
PAN~2 (GPT-6 Astra) & \shortstack{1595\\{\scriptsize[1568, 1623]}} & \shortstack{63.7\\{\scriptsize[59.3, 67.9]}} \\
PAN~2 (Claude Opus 5.5) & \shortstack{1559\\{\scriptsize[1537, 1582]}} & \shortstack{58.9\\{\scriptsize[55.2, 62.6]}} \\
Cosmos~3 Super & \shortstack{1380\\{\scriptsize[1350, 1408]}} & \shortstack{31.8\\{\scriptsize[27.7, 35.9]}} \\
LTX-2.5 & \shortstack{1346\\{\scriptsize[1313, 1376]}} & \shortstack{25.6\\{\scriptsize[21.4, 30.0]}} \\
\bottomrule
\end{tabular}
\end{table}

\section{Further studies of how the parts work together}
\label{app:ablations}


\subsection{Conceptual reasoning on further benchmarks}
\label{app:additional-reasoner-results}

The main text shows the impact of removing or varying the conceptual reasoning backbone on single-action and multi-step simulation (Fig.~\ref{fig:reasoner-caption}a,b).
Here, we also provide results for three public benchmarks in Supplementary Table~\ref{tab:reasoner-ablation}.
Using the conceptual reasoner backbone raises scores on PhyGround and RBench substantially compared to no LLM (by 0.29 to 0.45 points out of 5 on PhyGround and 15.7 to 18.5 points on RBench).
Physics-IQ Verified, which compares continuations of real recordings by motion metrics, changes less ($-1.6$ to $+4.0$ points).
Full reasoning improves on format conversion in every combination with data except Qwen 3.8 Flash on Physics-IQ Verified, although many differences are within the confidence intervals.


\begin{table}[t]
\centering
\small
\setlength{\tabcolsep}{4pt}
\caption{\textbf{Conceptual reasoning on public benchmarks.} A fixed PAN~2 LDM receives the action as a single event without an LLM (action only), a text-based state written by an LLM that only converts the input's format (format conversion), or one written by an LLM instructed to reason about how the world will evolve (full reasoning). $\Delta$ rows give full reasoning minus format conversion for the same LLM; $^{\ast}$ indicates that the corresponding confidence interval excludes 0. The Kimi-K3 full-reasoning row is the main PAN~2 run (Extended Data Table~\ref{tab:public-benchmarks}). GPT-6 Astra uses medium reasoning effort, as do the Claude Opus 5.5 full-reasoning runs. Results on single-action test cases and multi-step plans are in the main text (Fig.~\ref{fig:reasoner-caption}a,b).}
\label{tab:reasoner-ablation}
\resizebox{\linewidth}{!}{%
\begin{tabular}{llccc}
\toprule
LLM & Condition & Physics-IQ Verified & PhyGround & RBench \\
\midrule
None & Action only & 37.51\,{\scriptsize[33.14, 42.11]} & 4.30\,{\scriptsize[4.23, 4.36]} & 37.55\,{\scriptsize[35.22, 39.89]} \\
\midrule
Kimi-K3 & Format conversion & 35.88\,{\scriptsize[31.60, 40.36]} & 4.66\,{\scriptsize[4.60, 4.71]} & 53.27\,{\scriptsize[51.02, 55.52]} \\
 & Full reasoning & 37.48\,{\scriptsize[32.97, 42.08]} & 4.73\,{\scriptsize[4.69, 4.78]} & 54.54\,{\scriptsize[52.30, 56.74]} \\
 & $\Delta$ & $+1.61^{\ast}$\,{\scriptsize$[+0.04, +3.20]$} & $+0.08^{\ast}$\,{\scriptsize$[+0.04, +0.11]$} & $+1.26$\,{\scriptsize$[-0.46, +3.01]$} \\
\midrule
Claude Opus 5.5 & Format conversion & 41.30\,{\scriptsize[36.29, 46.55]} & 4.64\,{\scriptsize[4.59, 4.69]} & 54.63\,{\scriptsize[52.38, 56.80]} \\
 & Full reasoning & 41.49\,{\scriptsize[36.74, 46.38]} & 4.74\,{\scriptsize[4.69, 4.78]} & 54.77\,{\scriptsize[52.59, 56.90]} \\
 & $\Delta$ & $+0.19$\,{\scriptsize$[-1.46, +1.81]$} & $+0.09^{\ast}$\,{\scriptsize$[+0.06, +0.13]$} & $+0.14$\,{\scriptsize$[-1.45, +1.71]$} \\
\midrule
GPT-6 Astra & Format conversion & 39.03\,{\scriptsize[34.12, 44.20]} & 4.69\,{\scriptsize[4.64, 4.73]} & 55.32\,{\scriptsize[53.11, 57.50]} \\
 & Full reasoning & 40.02\,{\scriptsize[34.95, 45.27]} & 4.74\,{\scriptsize[4.70, 4.79]} & 56.09\,{\scriptsize[53.93, 58.25]} \\
 & $\Delta$ & $+0.98$\,{\scriptsize$[-1.20, +3.11]$} & $+0.06^{\ast}$\,{\scriptsize$[+0.02, +0.09]$} & $+0.77$\,{\scriptsize$[-0.60, +2.16]$} \\
\midrule
Qwen 3.8 Flash & Format conversion & 37.40\,{\scriptsize[32.63, 42.37]} & 4.59\,{\scriptsize[4.53, 4.64]} & 53.93\,{\scriptsize[51.71, 56.09]} \\
 & Full reasoning & 36.75\,{\scriptsize[31.89, 41.82]} & 4.72\,{\scriptsize[4.67, 4.76]} & 54.90\,{\scriptsize[52.64, 57.14]} \\
 & $\Delta$ & $-0.66$\,{\scriptsize$[-2.96, +1.59]$} & $+0.13^{\ast}$\,{\scriptsize$[+0.09, +0.17]$} & $+0.97$\,{\scriptsize$[-0.75, +2.69]$} \\
\bottomrule
\end{tabular}%
}
\end{table}

\subsection{Ablation of the perceptual backbone}
\label{app:perceptual-compute}

%

The perceptual-backbone ablation uses
the final PAN~2 checkpoint and fixed Kimi-K3 positive and negative captions on all 300 single-action test cases and 200 multi-step simulations, with seeds 123--125.
All 33,000 videos across 22 conditions were produced and scored.
The full-schedule reference was generated and evaluated alongside the reduced-computation conditions;
it is a separate run from the main model comparison.
Table~\ref{tab:perceptual-compute} gives the numerical results underlying Fig.~\ref{fig:reasoner-caption}c--e.
6D-PI uses decided questions under the standard scoring policy, and multi-step plan simulation uses all clips.
Bootstrap resampling uses seed 0.

\emph{Respacing} uses $N$ Euler steps over the full denoising interval,
with the same flow shifts (9 for I2V, 7 for T2V) and guidance scale 3.
Guidance-start and zero-initialization step counts are scaled by $N/N_{\mathrm{full}}$ and rounded to the nearest integer, with half-integers rounded upward.
\emph{Early termination} instead takes
the first $N-1$ steps of the original schedule, then uses the $N$th velocity prediction for one Euler jump to the clean endpoint before VAE decoding.
It retains the original guidance schedule; for a one-step T2V output, the final jump uses the conditional velocity rather than the zero-initialization velocity.
Both guidance branches are evaluated at every step in both conditions.

The two resolutions are $1248\times704$ and $864\times480$, with the same source image resized to cover and center-cropped for I2V.
Initial noise is shared across step budgets and schedules within each resolution, but drawn independently at the two latent shapes.
Clip durations, frame rate, text encoding, VAE decoding and evaluation protocol are otherwise unchanged.

\begin{table*}[t]
\centering
\footnotesize
\setlength{\tabcolsep}{4pt}
\caption{\textbf{Ablation of the perceptual backbone.} Mean scores (\%) with 95\% case-bootstrap intervals, using the same resampling as Fig.~\ref{fig:reasoner-caption}c--e. Steps are I2V/T2V denoiser evaluations. Early termination and respacing are defined in the text. $^{\dagger}$For zero full-plan completion, the interval is replaced by the one-sided 95\% upper bound over 100 plans. }
\label{tab:perceptual-compute}
\begin{adjustbox}{max width=\textwidth}
\begin{tabular}{lllccc}
\toprule
Resolution & Schedule & Steps & 6D-PI & Step completion & Full-plan completion \\
\midrule
720p & Full & 50/75 & 84.95\,{\scriptsize[82.20, 87.56]} & 75.48\,{\scriptsize[72.19, 78.69]} & 33.00\,{\scriptsize[26.50, 39.67]} \\
 & Respaced & 25/38 & 83.73\,{\scriptsize[80.97, 86.34]} & 74.54\,{\scriptsize[71.23, 77.78]} & 32.50\,{\scriptsize[26.00, 39.17]} \\
 & Respaced & 12/18 & 82.29\,{\scriptsize[79.38, 85.04]} & 71.99\,{\scriptsize[68.39, 75.52]} & 28.83\,{\scriptsize[22.50, 35.67]} \\
 & Respaced & 4/6 & 70.33\,{\scriptsize[66.65, 73.80]} & 46.79\,{\scriptsize[42.61, 50.93]} & 8.50\,{\scriptsize[5.00, 12.33]} \\
 & Respaced & 2/3 & 45.91\,{\scriptsize[40.93, 50.80]} & 9.05\,{\scriptsize[7.11, 11.14]} & 0.00\,{\scriptsize$[0,2.95]^{\dagger}$} \\
 & Respaced & 1/1 & 25.88\,{\scriptsize[21.57, 30.38]} & 0.39\,{\scriptsize[0.13, 0.69]} & 0.00\,{\scriptsize$[0,2.95]^{\dagger}$} \\
 & Early termination & 25/38 & 84.84\,{\scriptsize[82.22, 87.29]} & 73.77\,{\scriptsize[70.43, 76.99]} & 30.33\,{\scriptsize[24.17, 36.83]} \\
 & Early termination & 12/18 & 82.50\,{\scriptsize[79.78, 85.11]} & 65.50\,{\scriptsize[61.68, 69.16]} & 20.83\,{\scriptsize[15.33, 26.83]} \\
 & Early termination & 4/6 & 64.81\,{\scriptsize[60.76, 68.71]} & 16.68\,{\scriptsize[13.67, 19.84]} & 0.83\,{\scriptsize[0.17, 1.83]} \\
 & Early termination & 2/3 & 31.12\,{\scriptsize[26.34, 35.95]} & 1.77\,{\scriptsize[0.96, 2.72]} & 0.00\,{\scriptsize$[0,2.95]^{\dagger}$} \\
 & Early termination & 1/1 & 26.60\,{\scriptsize[22.02, 31.25]} & 1.67\,{\scriptsize[0.80, 2.69]} & 0.00\,{\scriptsize$[0,2.95]^{\dagger}$} \\
\midrule
480p & Full & 50/75 & 84.64\,{\scriptsize[82.01, 87.14]} & 72.57\,{\scriptsize[69.13, 75.93]} & 30.67\,{\scriptsize[24.67, 37.00]} \\
 & Respaced & 25/38 & 82.41\,{\scriptsize[79.82, 84.91]} & 72.18\,{\scriptsize[68.84, 75.43]} & 30.00\,{\scriptsize[24.17, 36.17]} \\
 & Respaced & 12/18 & 83.18\,{\scriptsize[80.49, 85.81]} & 72.61\,{\scriptsize[69.12, 75.98]} & 29.83\,{\scriptsize[23.67, 36.33]} \\
 & Respaced & 4/6 & 69.83\,{\scriptsize[66.21, 73.33]} & 50.08\,{\scriptsize[46.03, 54.11]} & 10.00\,{\scriptsize[6.17, 14.17]} \\
 & Respaced & 2/3 & 42.57\,{\scriptsize[37.40, 47.58]} & 7.26\,{\scriptsize[5.44, 9.09]} & 0.00\,{\scriptsize$[0,2.95]^{\dagger}$} \\
 & Respaced & 1/1 & 21.69\,{\scriptsize[17.62, 25.93]} & 0.48\,{\scriptsize[0.10, 0.96]} & 0.00\,{\scriptsize$[0,2.95]^{\dagger}$} \\
 & Early termination & 25/38 & 84.20\,{\scriptsize[81.62, 86.64]} & 70.24\,{\scriptsize[66.74, 73.64]} & 26.50\,{\scriptsize[21.00, 32.50]} \\
 & Early termination & 12/18 & 80.41\,{\scriptsize[77.54, 83.15]} & 55.73\,{\scriptsize[51.59, 59.87]} & 14.17\,{\scriptsize[10.00, 18.67]} \\
 & Early termination & 4/6 & 52.18\,{\scriptsize[47.66, 56.55]} & 5.88\,{\scriptsize[4.31, 7.58]} & 0.00\,{\scriptsize$[0,2.95]^{\dagger}$} \\
 & Early termination & 2/3 & 27.52\,{\scriptsize[22.83, 32.30]} & 0.80\,{\scriptsize[0.33, 1.33]} & 0.00\,{\scriptsize$[0,2.95]^{\dagger}$} \\
 & Early termination & 1/1 & 24.33\,{\scriptsize[19.73, 29.10]} & 1.34\,{\scriptsize[0.64, 2.13]} & 0.00\,{\scriptsize$[0,2.95]^{\dagger}$} \\
\bottomrule
\end{tabular}
\end{adjustbox}
\end{table*}

\subsection{Modality-specific and shared weights}
\label{app:architecture-ablation}

Here we consider the impact of a key choice in the design of the LDM itself.
Every block of PAN~2's transformer has separate weights for video and text tokens, sharing only the joint attention computation itself (Methods).
We compare this with a popular alternative: transformers that share one set of weights for all modalities.

\begin{figure}[t]
\centering
\includegraphics[width=0.5\linewidth]{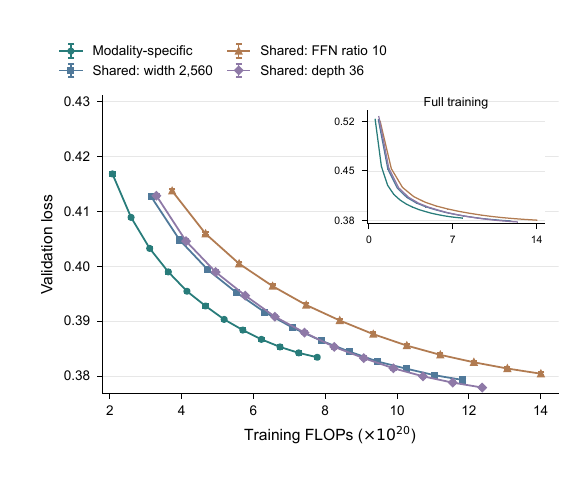}
\caption{\textbf{Modality-specific versus shared weights: validation loss.} Validation loss against estimated cumulative training FLOPs during joint image and video training, for modality-specific weights (3.061B parameters) and three approximately parameter-matched shared-weight variants: width 2,560 (3.048B), feed-forward expansion 10, or depth 36, compared with defaults of 2,048, 4 and 22. Main: checkpoints 4k--15k; inset: all 15 checkpoints. Error bars show 95\% Student-$t$ intervals over 100 matched evaluation passes on 9,998 fixed clips, conditional on a single training run per architecture. They measure evaluation Monte Carlo uncertainty. Curve endpoints have different compute budgets.
}
\label{fig:architecture-quality}
\end{figure}

For this expeirment, we target approximately 3 billion parameters, matching parameter counts by making the shared-weight transformer have a larger token dimension, larger feed-forward ratio, or more blocks.
We train on 25 million images at $256{\times}256$, and 25 million videos at $448{\times}256$ with 12 frames per second for 15,000 steps.
We evaluate validation loss on 9,998 held-out videos every 1,000 steps (Supplementary Fig.~\ref{fig:architecture-quality}).
Evaluation uses 100 passes with randomly-sampled timestep, Gaussian-noise, and caption-dropout kept identical across architectures and checkpoints (evaluation seed 0; uniform timesteps with shift 3, and caption-dropout probability 0.05).
The confidence intervals are 95\% Student-$t$ intervals over pass means, with 99 degrees of freedom, that measure evaluation Monte Carlo uncertainty conditional on these clips and a single training run per architecture.

At equal training compute, our transformer design has lower validation loss than all three shared-weight variants.
It reaches a final loss of 0.3834 with $7.78\times10^{20}$ FLOPs, whereas linear interpolation between neighboring checkpoints of the shared-weight variants at matched compute gives 0.3868 (wider), 0.3920 (wider feed-forward networks) and 0.3868 (deeper).

Separately, we time a training step and a denoiser call at 3 and 29 billion parameters in Supplementary Table~\ref{tab:architecture-speed}.
For training, we median time from 100 steps after warmup with kernel compilation and autotuning.
For inference, we take the median from 20 denoiser calls after three warm-up calls.
The tests are repeated five times, such that both architectures are benchmarked on the same five GPU nodes under matched settings.
We compute the speedup ratio within each node and report the mean $\pm 1.96$ standard errors across the ratios from the five different nodes, giving approximate 95\% confidence intervals under a normal approximation.
Modality-specific weights are 1.16 to 1.29 times faster per training step (which includes forward and backward) and 1.21 to 1.29 times faster per denoiser call than the wider shared-weight transformer.

\begin{table}[t]
\centering
\small
\setlength{\tabcolsep}{5pt}
\caption{\textbf{Modality-specific versus shared weights: speed.} Speed-up of the transformer with modality-specific weights over the approximately parameter-matched, wider transformer with shared weights, $t_{\mathrm{shared}}/t_{\mathrm{specific}}$, at two model sizes. Timings on synthetic videos measure one denoiser call on a single GPU (inference) and one training step on eight GPUs (training), with settings matched within each comparison. Values are the mean $\pm 1.96$ standard errors across five paired speedup ratios measured on five different nodes, giving approximate 95\% confidence intervals under a normal approximation.
}
\begin{tabular}{@{}lrrrr@{}}
\toprule
& \multicolumn{2}{c}{3B} & \multicolumn{2}{c}{29B} \\
\cmidrule(lr){2-3}\cmidrule(lr){4-5}
Video & Inference & Training & Inference & Training \\
\midrule
480p, 10 s & $1.287 \pm 0.008$ & $1.161 \pm 0.005$ & $1.266 \pm 0.014$ & $1.290 \pm 0.002$ \\
480p, 20 s & $1.268 \pm 0.013$ & $1.210 \pm 0.007$ & $1.234 \pm 0.012$ & $1.270 \pm 0.003$ \\
720p, 10 s & $1.267 \pm 0.017$ & $1.228 \pm 0.010$ & $1.233 \pm 0.012$ & $1.262 \pm 0.001$ \\
720p, 20 s & $1.256 \pm 0.017$ & $1.267 \pm 0.002$ & $1.209 \pm 0.011$ & $1.238 \pm 0.001$ \\
\bottomrule
\end{tabular}
\label{tab:architecture-speed}
\end{table}

\subsection{Text-based state}

\subsubsection{Caption recipes}
\label{app:caption-prompts}

\begin{figure*}[t]
\centering
\includegraphics[width=\textwidth]{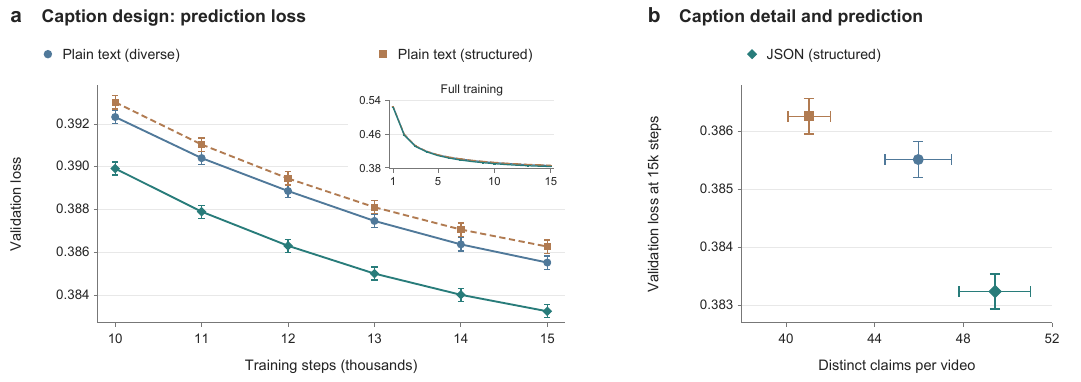}
\caption{\textbf{Text-based next-state description.} \textbf{a,} Caption-recipe validation loss on 9,998 held-out clips, with 100 evaluation passes per checkpoint. Main: 10--15k steps; inset: all 15 checkpoints. \textbf{b,} Loss at 15k steps versus mean distinct claims per video on 200 TemporalBench videos. Loss error bars in a and b show 95\% Student-$t$ intervals over 100 shared Monte Carlo evaluation passes (99 degrees of freedom), conditional on the fixed 9,998 clips and one trained model per recipe; they exclude variation across clips and training runs. Horizontal error bars in b show percentile 95\% bootstrap intervals for mean claim counts over 200 videos (10,000 shared whole-video resamples, seed 123), conditional on the extracted counts. Counts measure asserted detail, not video-verified correctness. Caption recipes differ in content and annotation procedure as well as serialization.}
\label{fig:caption-recipes}
\end{figure*}

Since the text-based state guides the LDM's prediction, its content and structure shape how well the LDM learns how the world evolves.
To study the impact of this design, we compare our captioning recipe, JSON (structured), with two earlier candidate recipes: Plain text (diverse), which emphasizes diversity in caption content, and Plain text (structured), which requires a fixed structure but remains in plain text.
The JSON (structured) recipe has the lowest held-out loss at all 15 measured checkpoints (Supplementary Fig.~\ref{fig:caption-recipes}a).

To understand how each recipe differs in information content, we measure caption information content as the number of distinct atomic claims per video, excluding formatting, schema keys, IDs, and standalone timestamps.
Supplementary Fig.~\ref{fig:caption-recipes}b shows that for 200 randomly selected TemporalBench~\citep{cai2024temporalbenchbenchmarkingfinegrainedtemporal} videos, JSON (structured) yields 49.43 assertions/video, compared with 45.97 for Plain text (diverse) and 41.04 for Plain text (structured).
This measures asserted detail per video, rather than information per token or video-verified correctness.
The results suggest that richer text-based states help the LDM learn to realize the states the LLM produces, although they do not isolate the effects of caption content and JSON serialization.

For the comparison of held-out loss, we train the 3B transformer with modality-specific weights on each recipe, on images and on videos at $448\times256$ as in the architecture study, and evaluate it on 9,998 held-out clips with 100 evaluation passes per checkpoint. The held-out clips are sampled from the matched 25-million-video set (seed 0) and excluded from training; two of the 10,000 sampled clips are excluded before evaluation by the fixed minimum-display-width rule, and none are skipped during evaluation. Each pass evaluates all clips with new timestep, noise and caption-dropout draws; corresponding passes share the same draws across recipes and checkpoints (evaluation seed 0). Loss is weighted by latent-element count. Confidence intervals are mean $\pm t_{0.975,99}$ times the standard error over the 100 pass means; paired intervals use the 100 corresponding pass differences. They describe evaluation Monte Carlo uncertainty conditional on the clips and one training run per recipe. At 15k steps, the paired JSON-minus-diverse-prose loss difference is $-0.002278$ (95\% CI $[-0.002308,-0.002247]$), and JSON minus structured prose is $-0.003024$ ($[-0.003057,-0.002991]$).
For the comparison of caption content, we select 200 of the 2,140 TemporalBench~\citep{cai2024temporalbenchbenchmarkingfinegrainedtemporal} videos with aligned captions in a fixed random order (seed 42); all 200 complete without exclusions, and every recipe describes the same videos and segment intervals.
Qwen3.6-27B, with thinking disabled and temperature 0, extracts the atomic claims of each caption, and a separate verification pass, run on recipe labels that are anonymized and shuffled, checks every non-identical pair within proposed groups of equivalent claims before repetitions and paraphrases are merged.
All three recipes use Qwen3.6-27B. The two plain-text recipes use a single captioning pass, whereas the JSON recipe uses event and action segmentation, camera evidence derived from geometry, a separate initial-frame description and a final structured caption, so the comparison is between complete captioning recipes, not serialization alone.
We estimate percentile 95\% confidence intervals using 10,000 whole-video bootstrap resamples (seed 123), with identical sampled video indices across recipes. Mean distinct claims per video are 45.965 [44.460, 47.475] for diverse prose, 41.040 [40.095, 42.000] for structured prose and 49.430 [47.805, 51.030] for structured JSON. Paired differences are 3.465 [1.410, 5.495] for JSON minus diverse prose, 8.390 [6.675, 10.090] for JSON minus structured prose and 4.925 [3.275, 6.565] for diverse minus structured prose. All 600 counts are present, with no missing, duplicate or excluded records. These intervals condition on the existing extracted counts and do not include extraction-model uncertainty.

\subsubsection{Camera annotation from geometry}
\label{subsubsec:camera_ablation}

Our pipeline derives camera labels from estimated geometry (VGGT-$\Omega$; Supplementary Information~\ref{app:world-caption-details}) rather than from the captioning VLM.
To test this choice, we replace VGGT-$\Omega$ with Qwen3.8-Flash-Next, which predicts camera labels from the same frames, event boundaries and vocabulary of labels.

Each method divides every event into camera windows, contiguous intervals each assigned a camera-motion label, and the two methods may place the boundaries of windows differently: in a 10-second event, VGGT-$\Omega$ may predict \texttt{tilt\_up} over 0--4~s and \texttt{static} over 4--10~s, while the VLM predicts \texttt{static} over 0--5~s and \texttt{pan\_left} over 5--10~s.
Because segmentation is part of the prediction, we evaluate each method on its own windows, grouped into three settings.
Window disagreement covers windows predicted by only one method, as all four are in this example, and tests segmentation and labelling together.
Label disagreement covers windows predicted by both methods with different labels; because both labels are judged on the same footage, it isolates the accuracy of the labels.
Label agreement covers windows predicted by both with the same label, scored once for both; it cannot separate the methods but shows how reliable the labels are when the methods agree.
We sample 125 label-disagreement and 100 window-disagreement windows per method and 50 label-agreement windows, 500 in total, oversampling disagreements because they distinguish the methods.
Three annotators rate every window, without knowing the method, as correct, partly correct or wrong, and we report 95\% confidence intervals from a bootstrap that resamples whole clips (10,000 replicates, stratified by setting), which keeps each clip's ratings together and the comparison paired.

Annotators rate the geometry-derived labels correct more often than the VLM's under both window and label disagreement (Supplementary Table~\ref{tab:camera_ablation}).
On the same footage, across label disagreements, annotators prefer the VGGT-$\Omega$ label on nearly twice as many windows as the VLM label (exact sign test, $p<0.01$).
Most of the VLM's errors are missed camera motion: in most label disagreements, the VLM labels the window \texttt{static}.

\begin{table}[h]
\centering
\caption{\textbf{Camera labels from geometry (VGGT-$\Omega$) and from a VLM (Qwen3.8-Flash-Next)}, rated blind by three annotators. Acc., correct (\%); Acc.$_\text{P}$, correct or partly correct (\%); $\alpha$, Krippendorff's $\alpha$ (ordinal); $\kappa$, Cohen's $\kappa$ (quadratic weights); $\pm$, 95\% confidence interval from a bootstrap over clips. Disagreements are oversampled, so the overall row is not an estimate of accuracy on the corpus.}
\label{tab:camera_ablation}
\small
\setlength{\tabcolsep}{3pt}
\begin{tabular}{l r cc cc cc}
\toprule
 & & \multicolumn{2}{c}{VGGT-$\Omega$ (ours)} & \multicolumn{2}{c}{VLM} & \multicolumn{2}{c}{Agreement} \\
\cmidrule(lr){3-4}\cmidrule(lr){5-6}\cmidrule(lr){7-8}
Setting & $n$ & Acc. & Acc.$_\text{P}$ & Acc. & Acc.$_\text{P}$ & $\alpha$ & $\kappa$ \\
\midrule
Window disagreement & 200 & 57.3{\scriptsize$\pm$8.3} & 70.3{\scriptsize$\pm$7.9} & 44.3{\scriptsize$\pm$8.1} & 52.0{\scriptsize$\pm$8.3} & 0.60{\scriptsize$\pm$0.08} & 0.61{\scriptsize$\pm$0.08} \\
Label disagreement  & 250 & 58.7{\scriptsize$\pm$7.0} & 74.1{\scriptsize$\pm$6.8} & 43.7{\scriptsize$\pm$7.5} & 48.8{\scriptsize$\pm$7.7} & 0.64{\scriptsize$\pm$0.08} & 0.65{\scriptsize$\pm$0.08} \\
Label agreement     &  50 & 64.0{\scriptsize$\pm$11.8} & 67.3{\scriptsize$\pm$11.9} & 64.0{\scriptsize$\pm$11.8} & 67.3{\scriptsize$\pm$11.9} & 0.72{\scriptsize$\pm$0.17} & 0.73{\scriptsize$\pm$0.17} \\
\midrule
Overall             & 500 & 59.2{\scriptsize$\pm$4.8} & 71.5{\scriptsize$\pm$4.7} & 47.6{\scriptsize$\pm$4.9} & 53.3{\scriptsize$\pm$5.1} & 0.63{\scriptsize$\pm$0.05} & 0.64{\scriptsize$\pm$0.05} \\
\bottomrule
\end{tabular}
\end{table}

\subsection{Training and optimization}
\subsubsection{Merging checkpoints}
\label{sec:model-soup-human-preference}

We compare the released model, which merges the Stage~III checkpoint (70\%) with the Stage~IV checkpoint after 575 steps (30\%),
with two equal-weight merges of Stage~II, III and IV checkpoints (Supplementary Table~\ref{tab:model-soup-human-preference}).
Reviewers were asked to give an overall preference for each pair.
The released model has the highest Bradley--Terry estimate (1015 Elo) and ranks first in 68.82\% of reviewer-bootstrap replicates, against 13.73\% and 17.45\% for the two alternatives.
Its paired Elo differences from the three- and four-checkpoint merges are $17.7\;[-34.1,90.8]$ and $28.3\;[-43.0,128.7]$, respectively (95\% reviewer-cluster bootstrap intervals); both include zero, so the study does not resolve a preference between the retained models.
A sensitivity analysis restricted to matched reviewers and cases (291 judgments, ten reviewers, 72 cases) also leaves all paired intervals spanning zero and changes the point ordering.
Rank-1 frequencies summarize conditional resampling, rather than calibrated probabilities of a true population ranking; their Monte Carlo precision describes finite simulation error, not uncertainty from sampling reviewers or cases.

\begin{table*}[t]
\centering
\small
\setlength{\tabcolsep}{3.5pt}
\caption{\textbf{Human preference between merged checkpoints.} A regularized Bradley--Terry fit uses 571 stored pairwise judgments from 15 reviewers over 173 observed cases, with ties counting as half a win and mean rating centered at 1000. The 95\% Elo intervals are reviewer-cluster percentile bootstrap intervals conditional on the benchmark, not win-rate intervals. Rank-1 is the frequency of the highest rating in 10,000 reviewer-resampling fits (seed 123). Its Monte Carlo standard error (MCSE) and Wilson 95\% Monte Carlo interval describe finite-simulation precision only, not study-sampling uncertainty. Comparison participation counts sum to twice the number of votes. The released model merges Stage~III (70\%) with Stage~IV after 575 steps (30\%); the other rows are equal-weight merges.
}
\label{tab:model-soup-human-preference}
\resizebox{\linewidth}{!}{%
\begin{tabular}{p{0.34\textwidth}rrrrrr}
\toprule
Model & Comparisons & Elo & 95\% Elo CI & Rank-1 (\%) & MCSE (pp) & 95\% MC interval (\%) \\
\midrule
Released: 70\% Stage~III + 30\% Stage~IV (575 steps)
& 373 & 1015.3 & $[975.2,\,1072.4]$ & 68.82 & 0.46 & $[67.91,\,69.72]$ \\
Stage~IV (575 steps) + Stage~III + Stage~II
& 375 & 997.6 & $[978.7,\,1013.8]$ & 13.73 & 0.34 & $[13.07,\,14.42]$ \\
Stage~IV (325 steps) + Stage~II + Stage~IV (575 steps) + Stage~III
& 394 & 987.0 & $[943.2,\,1018.8]$ & 17.45 & 0.38 & $[16.72,\,18.21]$ \\
\bottomrule
\end{tabular}
}
\end{table*}

\subsubsection{Optimizer}
\label{sec:optimizer-ablation}

\begin{figure}[t]
\centering
\includegraphics[width=0.85\linewidth]{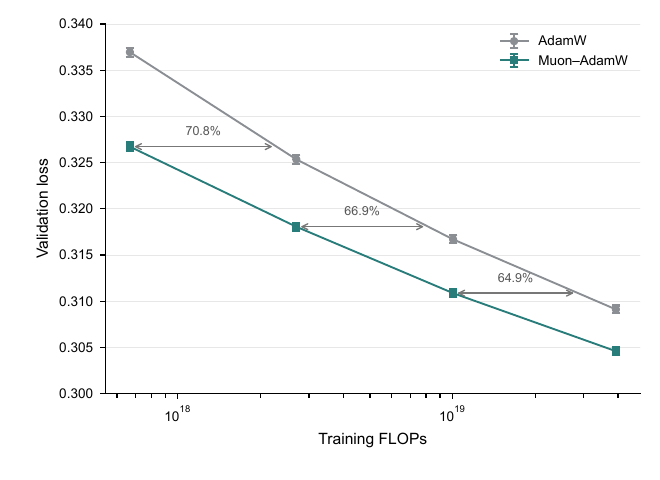}
\caption{\textbf{Optimizer scaling on images.} Final validation loss against training compute at the selected batch size and learning rate, with 95\% Student-$t$ intervals over 100 matched evaluation passes on 5,000 fixed images. Intervals measure evaluation Monte Carlo uncertainty conditional on these images, selected settings and single training runs. Horizontal annotations give FLOP reductions at matched loss and their 95\% shared-pass bootstrap intervals (10,000 resamples, seed 0), using interpolation in log FLOPs. They exclude interpolation uncertainty and optimizer-step computation, and are not wall-clock speed-ups. No reduction is extrapolated beyond the largest AdamW point.
}
\label{fig:optimizer-scaling}
\end{figure}

We compare AdamW with our hybrid of Muon and AdamW in a scaling study that trains image models of 185 million to 1.52 billion parameters on 100 million images, selecting the batch size and learning rate with the lowest validation loss for each point.
The hybrid optimizer reaches lower validation loss at all four model sizes (Supplementary Table~\ref{tab:optimizer-scaling} and Supplementary Fig.~\ref{fig:optimizer-scaling}).
The advantage shrinks with model size, from 0.010 at 185 million parameters to 0.005 at 1.52 billion, but persists across the sizes tested; interpolation along the curves gives 65--71\% fewer training FLOPs at matched loss. Each point is selected from a grid of batch sizes 512, 1,024 and 2,048 and learning rates $2\times10^{-4}$, $4\times10^{-4}$ and $8\times10^{-4}$. We evaluate each selected checkpoint in 100 passes over 5,000 fixed held-out images, with identical timestep, noise and caption-dropout draws across settings (evaluation seed 0). Loss intervals use Student-$t$ intervals over pass means (99 degrees of freedom); difference intervals use paired pass differences. These intervals condition on the selected hyperparameters, validation images and one training run per setting (training seed 42), excluding uncertainty from hyperparameter selection, validation-image sampling and repeated training. Matched-loss FLOP-reduction intervals use 10,000 shared-pass bootstrap resamples (seed 0) and interpolation in log FLOPs, excluding interpolation uncertainty. FLOPs count forward and backward computation and exclude the optimizer step.

\begin{table}[t]
\centering
\small
\setlength{\tabcolsep}{4pt}
\caption{\textbf{Optimizer scaling on images.} Validation loss at the selected batch size and learning rate for each optimizer and model size; lower is better. $\Delta$, hybrid Muon--AdamW minus AdamW. Brackets give 95\% Student-$t$ intervals over 100 matched evaluation passes on 5,000 fixed images (99 degrees of freedom); difference intervals use paired pass differences. Intervals quantify evaluation Monte Carlo uncertainty conditional on the selected settings and one trained model per setting.
}
\label{tab:optimizer-scaling}
\begin{tabular}{lcccc}
\toprule
Model & Depth $\times$ width & AdamW & Muon--AdamW & $\Delta$ loss \\
\midrule
185M & $8\times768$ & \shortstack{0.33696\\{\scriptsize[0.33648, 0.33743]}} & \shortstack{0.32673\\{\scriptsize[0.32627, 0.32719]}} & \shortstack{$-0.01023$\\{\scriptsize$[-0.01026, -0.01020]$}} \\
390M & $10\times1024$ & \shortstack{0.32535\\{\scriptsize[0.32490, 0.32581]}} & \shortstack{0.31806\\{\scriptsize[0.31761, 0.31850]}} & \shortstack{$-0.00730$\\{\scriptsize$[-0.00732, -0.00727]$}} \\
756M & $13\times1280$ & \shortstack{0.31669\\{\scriptsize[0.31625, 0.31713]}} & \shortstack{0.31087\\{\scriptsize[0.31044, 0.31130]}} & \shortstack{$-0.00582$\\{\scriptsize$[-0.00584, -0.00580]$}} \\
1.52B & $16\times1664$ & \shortstack{0.30911\\{\scriptsize[0.30868, 0.30954]}} & \shortstack{0.30459\\{\scriptsize[0.30417, 0.30501]}} & \shortstack{$-0.00452$\\{\scriptsize$[-0.00454, -0.00450]$}} \\
\bottomrule
\end{tabular}
\end{table}

\section{Related Work}
\label{sec:related-work}

\subsection{Generative World Modeling}

\cite{ha2018world} propose a latent representation of the world as an environment for training agents.
\cite{hafner2019learning} learn the latent dynamics of an environment from pixel observations by predicting future states.
Subsequent works~\citep{hafner2019dream,hafner2020mastering,hafner2023mastering} use reinforcement learning in ``latent imagination'' to train agents with such reconstruction-based world models.
Other methods use next-state prediction objectives focused on not only latent states but also rewards and actions to learn a model of only the task-relevant parts of the world~\citep{hansen2022temporal,ICLR2024_cf73d57b}.
Generative latent prediction (GLP)~\citep{xing2025critique}, first instantiated by PAN~\citep{panteam2026panworldmodelgeneral}, formalizes an architecture of an encoder, a backbone and a decoder that learns by reconstructing the next observation from the predicted next state.

Some works instantiate world models as interactive game environments, learning to simulate the consequences of user actions from gameplay videos~\citep{bruce2024genie,parkerholder2024genie2,genie3,valevski2025diffusion,alonso2024diffusion,yang2023learning}.
These models generate visual observations conditioned on user controls, typically using autoregressive diffusion-based prediction to simulate the evolution of the environment.

PAN~2's perceptual backbone builds on video generation.
Video synthesis is accomplished using diffusion-based architectures~\citep{ho2022video,ho2022imagen,blattmann2023stable,bar2024lumiere}, masked token prediction~\citep{villegas2022phenaki}, or autoregressive generation~\citep{kondratyuk2023videopoet}.
Recent large-scale video generation models~\citep{yang2025cogvideox,wan2025,zheng2024opensorademocratizingefficientvideo,nvidia2025cosmosworldfoundationmodel,lin2025stiv} demonstrate increasing capabilities with advancements in parameter and data scaling and efficiency.

Some systems focus on autonomous driving~\citep{hu2023gaia,russell2025gaia,wang2024drivedreamer,zhao2025drivedreamer}.
They generate future driving scenarios conditioned on vehicle actions, trajectories, road layouts, or other structured scene information.
These domain-specific world models show the value of action-conditioned video generation for simulation, while relying on representations and action spaces specialized to driving.

\subsection{Predictive World Modeling}

An alternative approach to world modeling learns to predict future states directly in a latent representation space, without reconstructing the corresponding observations.
Joint-Embedding Predictive Architectures (JEPA)~\citep{lecun2022path,assran2023self} learn representations by predicting the embeddings of unobserved inputs from observed context.
V-JEPA~\citep{bardes2024revisiting} extends this framework to video, learning representations that capture temporal structure through masked feature prediction.
Subsequent works~\citep{assran2025v,mur2026v} scale these predictive representations and demonstrate their utility for video understanding and physical-world planning~\citep{zhang2026hjepaendtoendlearninghierarchical}.
Other approaches, such as DINO-WM~\citep{zhou2024dino}, predict future states using pretrained visual representations for model-based planning.

\subsection{Other Approaches}

World-action models (WAMs) couple predictions of future world states with action generation, integrating environmental dynamics and embodied control~\citep{lu2026world,wang2026world}.
DreamZero~\citep{ye2026world} jointly generates future video observations and robot actions using an autoregressive video diffusion model, leveraging video prediction to learn transferable manipulation policies.
ViPRA~\citep{routray2026vipra} instead learns motion-centric latent actions alongside future visual observations from videos, subsequently mapping these representations to continuous robot controls.
Other approaches integrate predictive simulation more directly into action selection.
$\tau_0$-WM~\citep{zhou2026tau0wmunifiedvideoactionworld} unifies video prediction, action generation, and action evaluation through a shared diffusion backbone, using imagined futures to evaluate and refine candidate actions.
OpenWAM~\citep{yu2026openwam} explores modular architectures for coupling world prediction with action generation, while JEPA-WAM~\citep{lin2026jepa} incorporates latent predictive world modeling into vision-language-action policies.
OSCAR~\citep{wu2026oscar} conditions video prediction on embodiment-independent skeletal representations to simulate the consequences of robotic actions and support virtual policy evaluation.

Other approaches represent environmental dynamics through executable programs or structured state representations, explicitly separating the mechanisms governing world evolution from their visual observations.
Code World Model~\citep{chen2026code} employs a coding agent to maintain persistent world state and execute rule-consistent transitions, using a video generator to render the resulting observations.
Code World Models for General Game Playing~\citep{lehrach2026code} similarly synthesizes executable transition functions from natural language game rules, enabling search-based planning over explicitly simulated states.
Related work models computational environments by learning from program execution and interaction trajectories~\citep{copet2025cwm}, allowing language models to predict the consequences of actions within executable environments.
A longer line of research represents physical dynamics through structured object interactions.
Interaction Networks~\citep{battaglia2016interaction} and Graph Network-based Simulators~\citep{sanchez2020learning} model state evolution through learned interactions between objects or particles, while DreMa~\citep{barcellona2025dream} combines explicit physics simulation with neural rendering to construct compositional robotic environments.

\section{Case studies}
\label{app:case-studies}

\paragraph{Selection and training-data search.}
The examples in Fig.~\ref{fig:simulation_branches} are selected by qualitative inspection.
For the training-data search, we embed captions with SentenceTransformer and search the entire training corpus.
We visually inspect the videos corresponding to the top 30 caption matches for each query.
We do not find training data that substantively matches the examples we show, except for where specified in the main text.

\paragraph{Paper-plane simulation.}
A human supplies a sequence of folding instructions, and the last frame of each generated clip is used as the starting image for the next instruction (Fig.~\ref{fig:agent-interaction}a).
Starting frames are repaired with an image restoration model if their quality has degraded.

\paragraph{Tube simulation.}
Kimi-K3 proposes a plan to remove a ball from a tube with a removable blue cap and a glued red cap (Fig.~\ref{fig:agent-interaction}b).
PAN~2 is given the hidden glue constraint, but the acting Kimi-K3 agent is not.
Each of the six attempts generates a five-second simulation from the same starting image.
Kimi-K3 reviews the clip its previous plans to formulate its plan for the next attempt.

\paragraph{Persistent kitchen simulation.}
We take inspiration from FastVideo DreamVerse~\citep{zhang2025vsa} to enable persistent world simulation within PAN~2.
In an isolated setup conversation, Kimi-K3 acts as a setup utility external to both the agent model and the world model.
The setup utility receives the goal and starting image and writes a world description naming the available objects (e.g., a pot with a lid, a package of spaghetti, salt, a colander, and a long-handled spoon).
Kimi-K3 chooses the locations of unseen objects, placing the pot in the cabinet under the stove and the colander under the sink, for example.
The acting Kimi-K3 agent starts in a fresh conversation with the goal and starting image, without the setup description or any of the location secrets.
PAN~2 maintains their location secrets and object states, and the position of the camera, in private text memory; it never receives the goal.
For each action, PAN~2's LDM is conditioned on its text memory and the previous clip's last frame.
PAN~2's conceptual reasoner checks the generated clip against the requested action and remembered state, to ensure instruction-following and world consistency, and allows up to two regenerations before updating the memory.
The conceptual reasoner additionally chooses clip lengths from 8 to 20 seconds in two-second increments; generation uses $704\times1248$ resolution, 24 frames per second, 50 sampling steps and guidance 3.0.
The displayed episode comprises 19 actions and 294 seconds of simulated video; 14 of 30 generated clips were flagged by the automated checks from the conceptual reasoner, and retried to ensure that the simulated video output aligns with the plans proposed by the agent.

\end{document}

%% file: math_commands.tex
\usepackage{amsmath,amsfonts,bm}

\newcommand{\quality}{\operatorname{quality}}

\def\eqref#1{equation~\ref{#1}}

\def\1{\bm{1}}

\DeclareMathAlphabet{\mathsfit}{\encodingdefault}{\sfdefault}{m}{sl}
\SetMathAlphabet{\mathsfit}{bold}{\encodingdefault}{\sfdefault}{bx}{n}

\usepackage{amsmath}
\usepackage{amssymb}
\usepackage{booktabs}
\usepackage{tabularx}
\usepackage{array}
\usepackage{xcolor}
\usepackage{xurl}
\usepackage{listings}
\usepackage{tcolorbox}
\tcbuselibrary{breakable,listings,skins}
\usepackage{tikz}
\usepackage{adjustbox}
\usetikzlibrary{arrows.meta,positioning}

\definecolor{TGCInk}{HTML}{17212B}
\definecolor{TGCMuted}{HTML}{596774}
\definecolor{TGCRule}{HTML}{D6DEE5}
\definecolor{TGCStageOne}{HTML}{2F6B9A}
\definecolor{TGCStageTwo}{HTML}{16827A}
\definecolor{TGCStageThree}{HTML}{7253A3}
\definecolor{TGCLightOne}{HTML}{F2F7FB}
\definecolor{TGCLightTwo}{HTML}{EFF9F6}
\definecolor{TGCLightThree}{HTML}{F6F2FA}
\definecolor{TGCCodeBg}{HTML}{F7F8FA}
\definecolor{TGCJsonString}{HTML}{0B756A}
\definecolor{TGCJsonLiteral}{HTML}{9A4D13}

\newtcolorbox{tgcstagebox}[1][]{
  colback=TGCLightThree,
  colframe=TGCStageThree,
  coltext=TGCInk,
  colbacktitle=TGCLightThree,
  coltitle=TGCStageThree,
  fontupper=\small,
  fonttitle=\bfseries\small,
  boxrule=0.55pt,
  arc=1.0mm,
  left=1.1mm,
  right=1.1mm,
  top=0.8mm,
  bottom=0.8mm,
  before skip=3pt,
  after skip=4pt,
  #1
}

\lstdefinelanguage{tgcjson}{
  basicstyle=\ttfamily\footnotesize,
  showstringspaces=false,
  breaklines=true,
  breakatwhitespace=false,
  columns=fullflexible,
  keepspaces=true,
  morestring=[b]",
  stringstyle=\color{TGCJsonString},
  morekeywords={true,false,null},
  keywordstyle=\bfseries\color{TGCJsonLiteral},
  comment=[l]{//},
  commentstyle=\itshape\color{TGCMuted}
}

\newtcblisting{tgcjsonbox}[1]{
  enhanced,
  breakable,
  listing only,
  listing engine=listings,
  title={#1},
  fonttitle=\bfseries\small,
  coltitle=TGCInk,
  colback=TGCCodeBg,
  colframe=TGCRule,
  boxrule=0.6pt,
  arc=1.2mm,
  left=1.2mm,right=1.2mm,top=1.0mm,bottom=1.0mm,
  listing options={
    language=tgcjson,
    basicstyle=\ttfamily\scriptsize,
    showstringspaces=false,
    breaklines=true,
    columns=fullflexible,
    keepspaces=true
  }
}

\newtcblisting{tgccontextbox}[1]{
  enhanced,
  breakable,
  listing only,
  listing engine=listings,
  title={#1},
  fonttitle=\bfseries\small,
  coltitle=TGCInk,
  colback=TGCLightThree,
  colframe=TGCRule,
  boxrule=0.6pt,
  arc=1.2mm,
  left=1.2mm,right=1.2mm,top=1.0mm,bottom=1.0mm,
  listing options={
    basicstyle=\ttfamily\scriptsize,
    showstringspaces=false,
    breaklines=true,
    breakatwhitespace=false,
    columns=fullflexible,
    keepspaces=true
  }
}